\documentclass{article}
\usepackage{jfrExamplee}
\usepackage{graphicx}
\usepackage{natbib}
\usepackage{apalike}
\usepackage[doublespacing]{setspace}
\usepackage{url}
\usepackage{amsmath,amssymb}
\usepackage{textcomp}
\usepackage{bm}
\usepackage{epstopdf}
\usepackage{booktabs}
\usepackage{tabularx}
\usepackage{paralist}
\usepackage{pgfplots}
\usepackage[binary-units=true,product-units=single,per-mode=symbol]{siunitx}
\usepackage{tikz}
\usetikzlibrary{fit,shapes.callouts,shapes.geometric,backgrounds,positioning,arrows.meta}
\usepackage{xspace}
\usepackage[english,capitalise]{cleveref}

\graphicspath{{./figures/}}

\newcommand{\ie}{i.e.,\ }
\newcommand{\eg}{e.g.,\ }

\newcommand{\reffig}[1]{Fig.~\ref{#1}}
\newcommand{\reftab}[1]{Tab.~\ref{#1}}
\newcommand{\refsec}[1]{Sec.~\ref{#1}}

\DeclareMathOperator{\atantwo}{atan2}

\newlength{\figureheight}

\definecolor{MBZIRCgreen}{rgb}{0.0,0.5,0.0}
\definecolor{MBZIRCred}{rgb}{1.0,0.0,0.0}
\definecolor{MBZIRCblue}{rgb}{0.0,0.0,1.0}
\definecolor{MBZIRCyellow}{rgb}{0.6,0.6,0.0}

\title{Team NimbRo at {MBZIRC} 2017: Fast Landing on a Moving Target and Treasure Hunting with a Team of MAVs}

\author{
Marius Beul, Matthias Nieuwenhuisen, Jan Quenzel, Radu Alexandru Rosu, \\
\textbf{Jannis Horn, Dmytro Pavlichenko, Sebastian Houben, and Sven Behnke} \\
Autonomous Intelligent Systems Group \\
University of Bonn \\
Bonn, Germany \\
\texttt{mbeul@ais.uni-bonn.de}\\
}

\begin{document}

\begin{center}
  \vspace*{-3cm}
  \fbox{Journal of Field Robotics (JFR) 36(1):204-229, Wiley, January 2019.}
  \vspace*{2cm}
\end{center}

\maketitle

\begin{abstract}
The Mohamed Bin Zayed International Robotics Challenge (MBZIRC) 2017 has defined ambitious new benchmarks to advance the state-of-the-art in autonomous operation of ground-based and flying robots. This article covers our approaches to solve the two challenges that involved micro aerial vehicles (MAV). Challenge~1 required reliable target perception, fast trajectory planning, and stable control of an MAV in order to land on a moving vehicle. Challenge~3 demanded a team of MAVs to perform a search and transportation task, coined ``Treasure Hunt'', which required mission planning and multi-robot coordination as well as adaptive control to account for the additional object weight. We describe our base MAV setup and the challenge-specific extensions, cover the camera-based perception, explain control and trajectory-planning in detail, and elaborate on mission planning and team coordination. We evaluated our systems in simulation as well as with real-robot experiments during the competition in Abu Dhabi. With our system, we---as part of the larger team NimbRo---won the MBZIRC Grand Challenge and achieved a third place in both subchallenges involving flying robots.
\end{abstract}

\section{Introduction}
Operating complex robotic systems without human interaction in only partially known environments is a demanding task.
In particular, these systems have to be robust against failures, environmental changes, \eg varying lighting conditions, and should not rely on central infrastructure.
Robotic competitions like the Mohamed Bin Zayed International Robotics Challenge (MBZIRC) expedite the development of systems that can be quickly adapted to new situations and work robustly outside of a controlled lab environment \citep{ram_competitions}. The MBZIRC competition took place in Abu Dhabi March 16th--18th, 2017.

Even though the individual subtasks at MBZIRC were of moderate complexity, their combination, time constraints, and fully autonomous operation posed high demands on the participating teams.
One of the major challenges was the very limited test and setup time before competition runs, denying, \eg color calibration for the current lighting conditions before a run. Challengers were allowed to enter the arena only two times for \SI{35}{\minute} before the competition days, while flying outside of the arena was completely prohibited.
The development of the required skills for these tasks complement the advancement of individual components beyond the current state-of-the-art by employing and robustifying simple but easy-to-handle components.

In this article, we describe our approach to the Micro Aerial Vehicle (MAV) challenges at MBZIRC, namely Challenge 1 "Landing on a Moving Platform" and Challenge 3 "Treasure Hunt".

\begin{figure}[tb]
  \centering
  \setlength{\figureheight}{0.21\linewidth}
  \includegraphics[height=\figureheight]{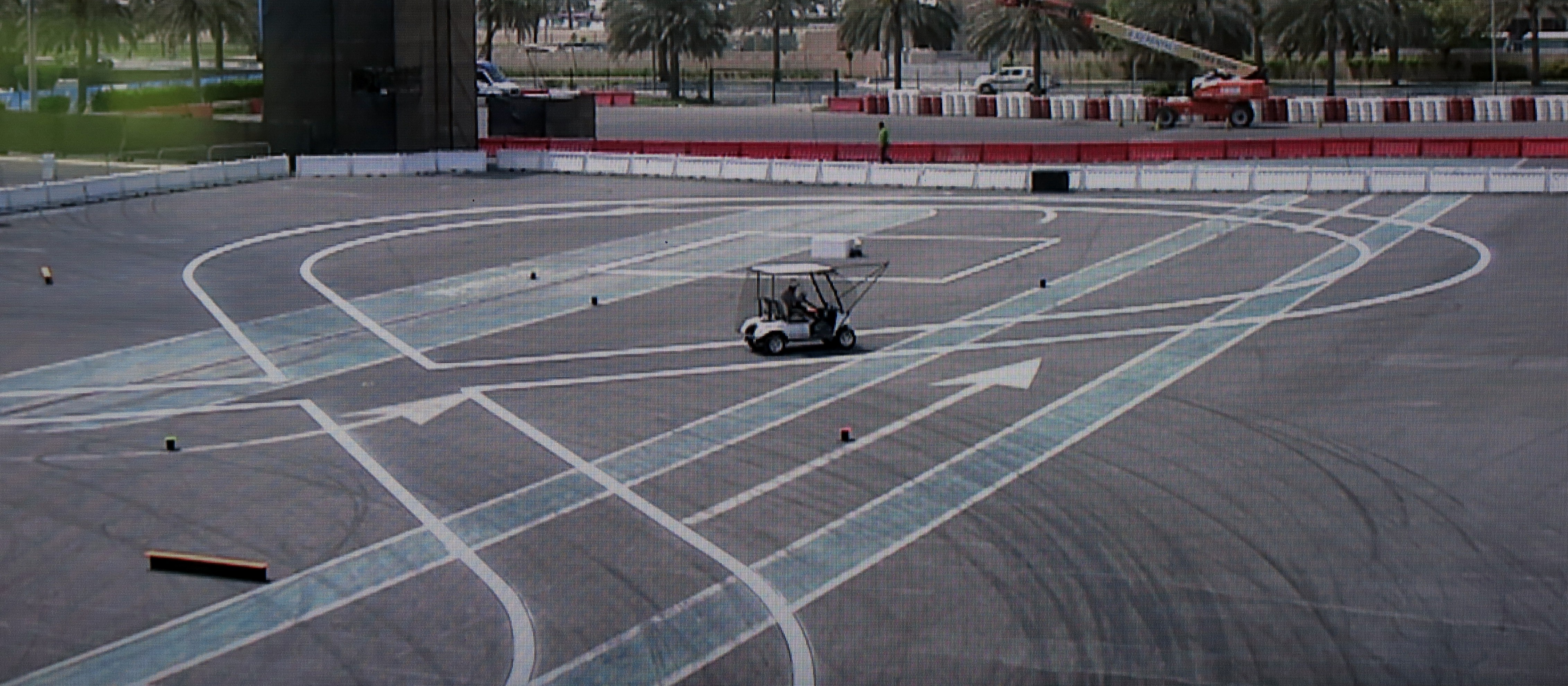}~
  \begin{tikzpicture}
    \definecolor{red}{rgb}{0.7,0.0,0.0}
    \node[inner sep = 0,anchor=north west] at (0,0) {\includegraphics[trim=00mm 30mm 00mm 10mm,clip,height=\figureheight]{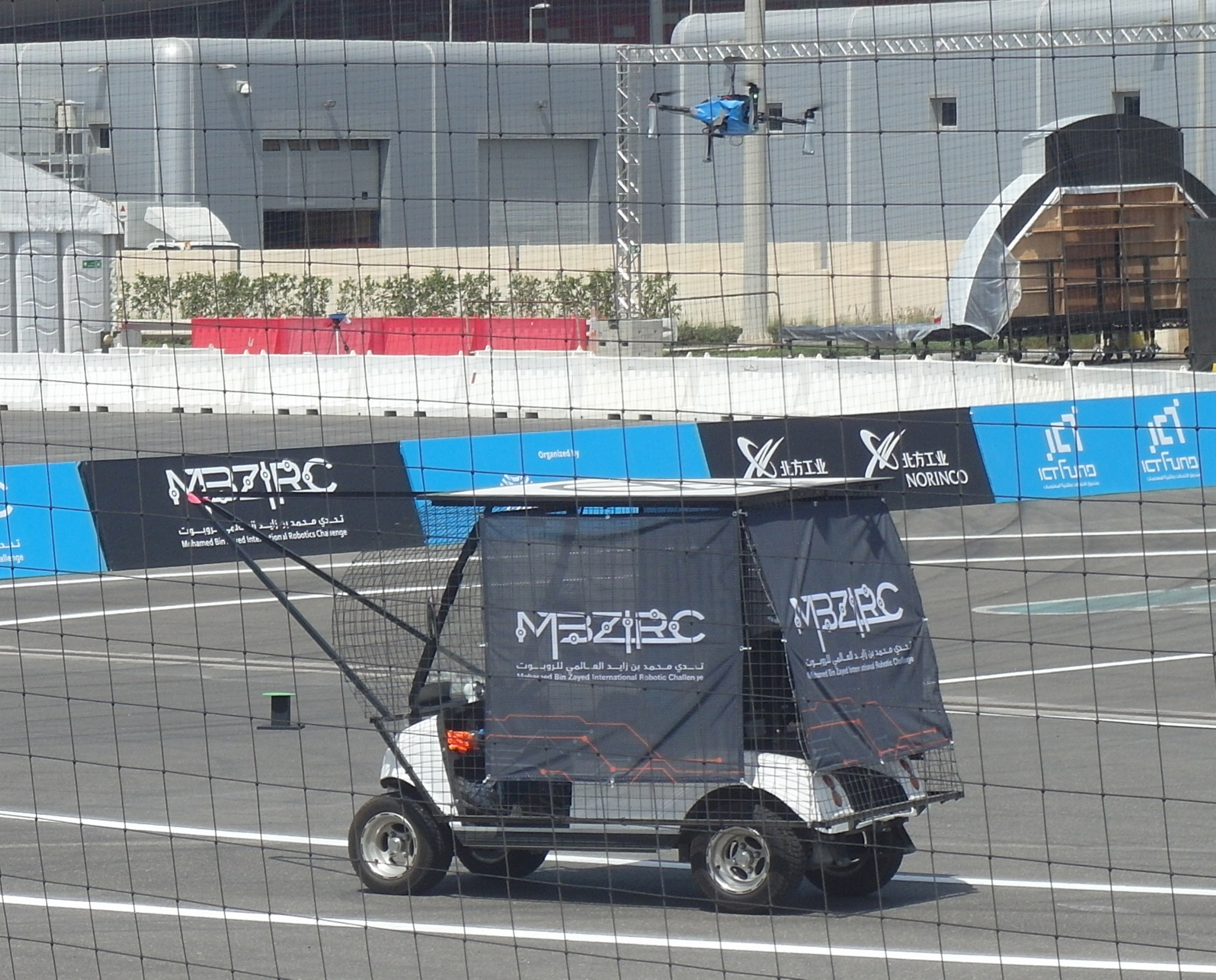}};
    \draw[line width=0.4mm, red, opacity=1.0] (2.82, -0.39) circle (0.5cm and 0.35cm);
  \end{tikzpicture}~
  \includegraphics[height=\figureheight]{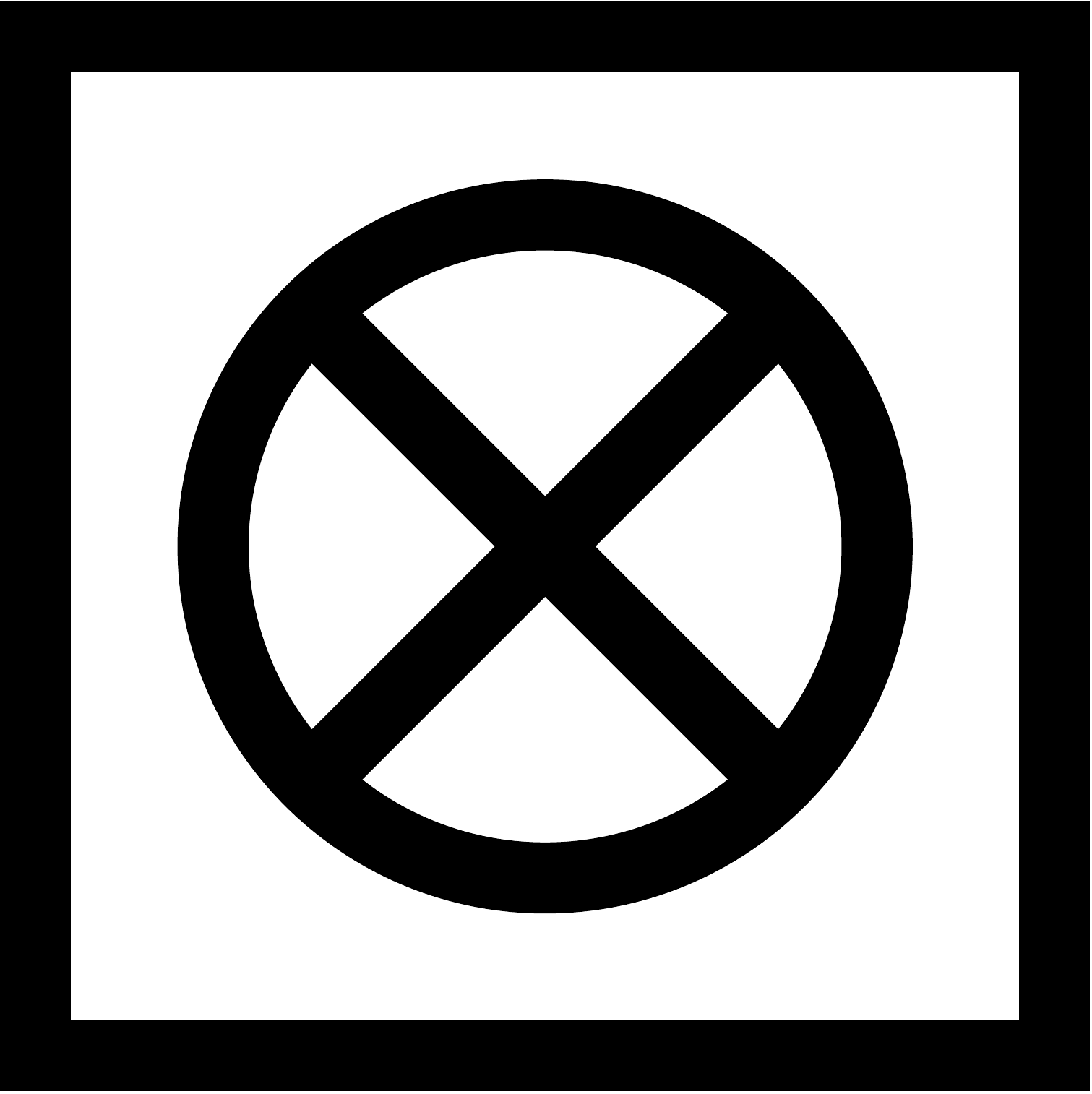}
  \caption{Landing on a moving platform. Left: The vehicle with the platform is driving on a figure eight track with \SI{15}{\kilo\meter\per\hour}. Center: Final approach only moments before a successful landing in the Grand Challenge. Right: Illustration of the landing pattern on top of the vehicle.}
  \label{fig:golf_cart}
\end{figure}

In Challenge 1, an MAV had to search for and land on a marked platform mounted on a vehicle, driving with \SI{15}{\kilo\meter\per\hour} on a figure eight track in the \SI{90 x 60}{\meter} arena.
The main metric in this challenge was the time that it took the MAV to land on the platform after the challenge start signal.
Although the vehicle would slow down after eight minutes, a team had to land the robot in the first minutes---autonomously and without any damage---in order to receive a competitive score for its run.
In fact, the teams ranked highest were able to complete the task in less than \SI{1}{\minute} after takeoff, including the time needed to search for the moving target. \Cref{fig:golf_cart} shows the arena setup and the final approach of our MAV before landing.

\begin{figure}[t]
  \setlength{\figureheight}{0.22\linewidth}
  \centering
  \begin{tikzpicture}
    \definecolor{red}{rgb}{0.7,0.0,0.0}
    \node[inner sep = 0,anchor=north west] at (0,0) {\includegraphics[trim=00mm 00mm 00mm 00mm,clip,height=\figureheight]{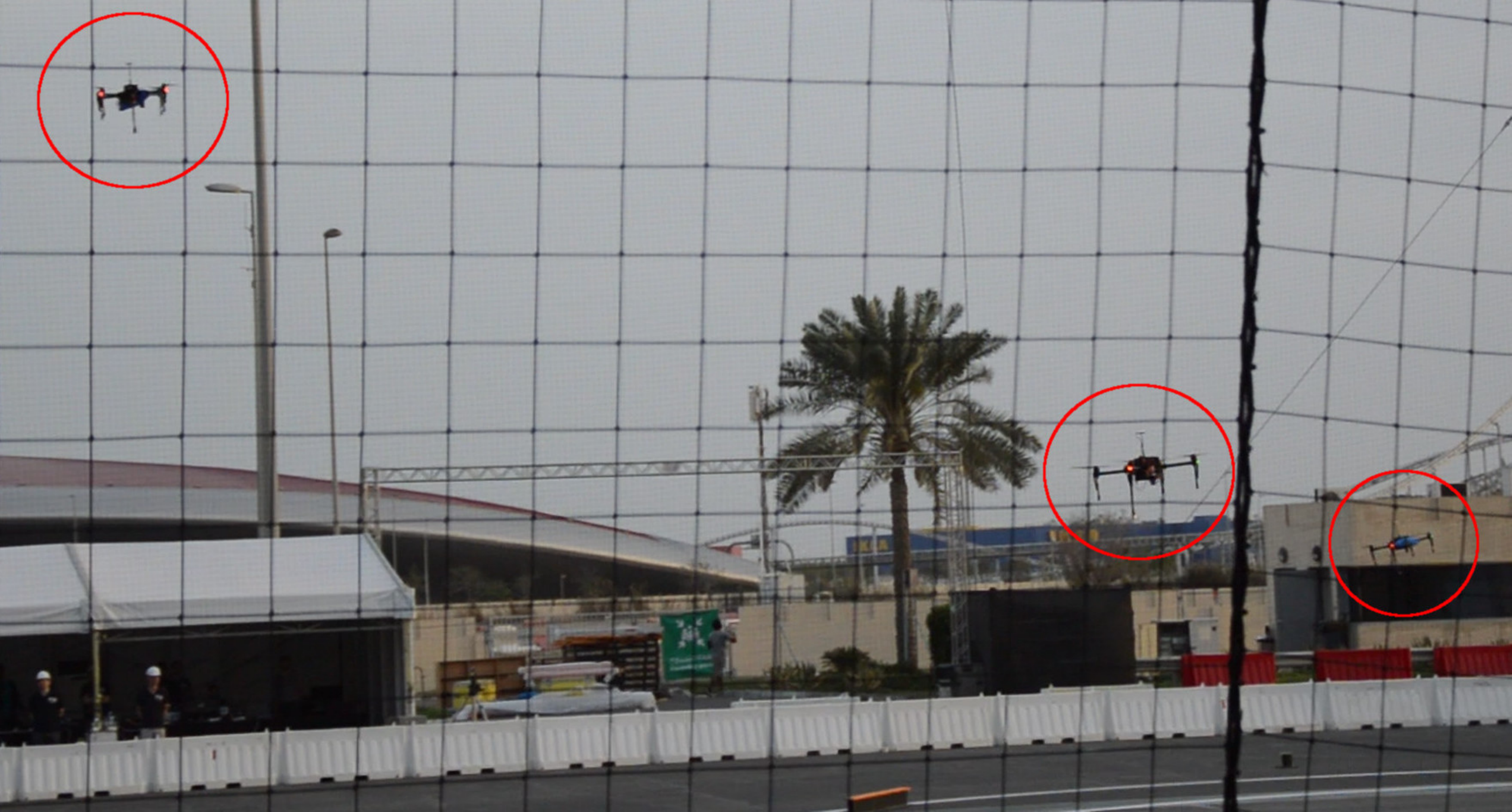}};
    \draw[line width=0.4mm, red, opacity=1.0] (0.6, -0.45) circle (0.42cm and 0.39cm);
    \draw[line width=0.4mm, red, opacity=1.0] (5.11,-2.12) circle (0.42cm and 0.39cm);
    \draw[line width=0.4mm, red, opacity=1.0] (6.29,-2.44) circle (0.33cm and 0.32cm);
  \end{tikzpicture}~
  \includegraphics[height=\figureheight]{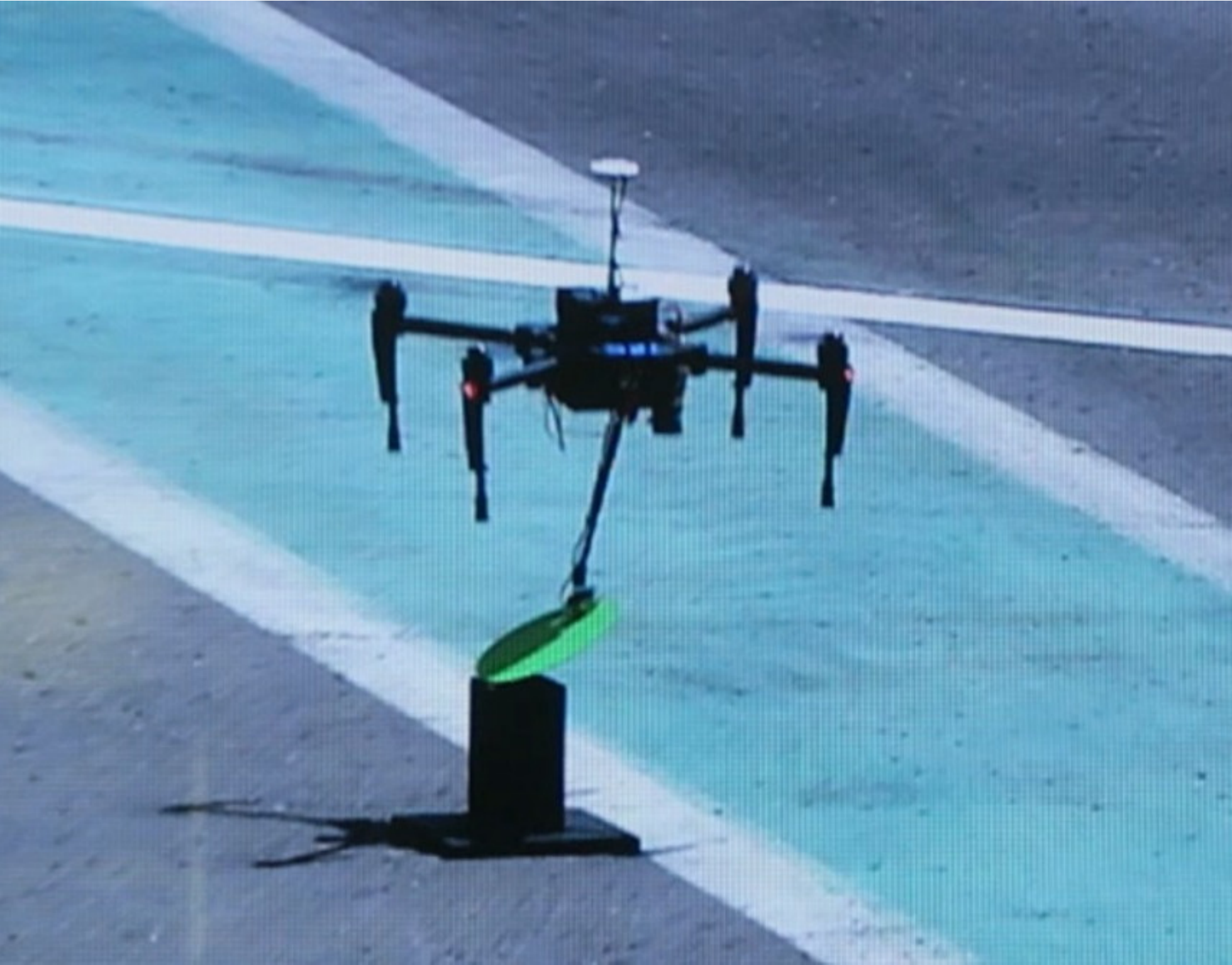}~
  \includegraphics[height=\figureheight]{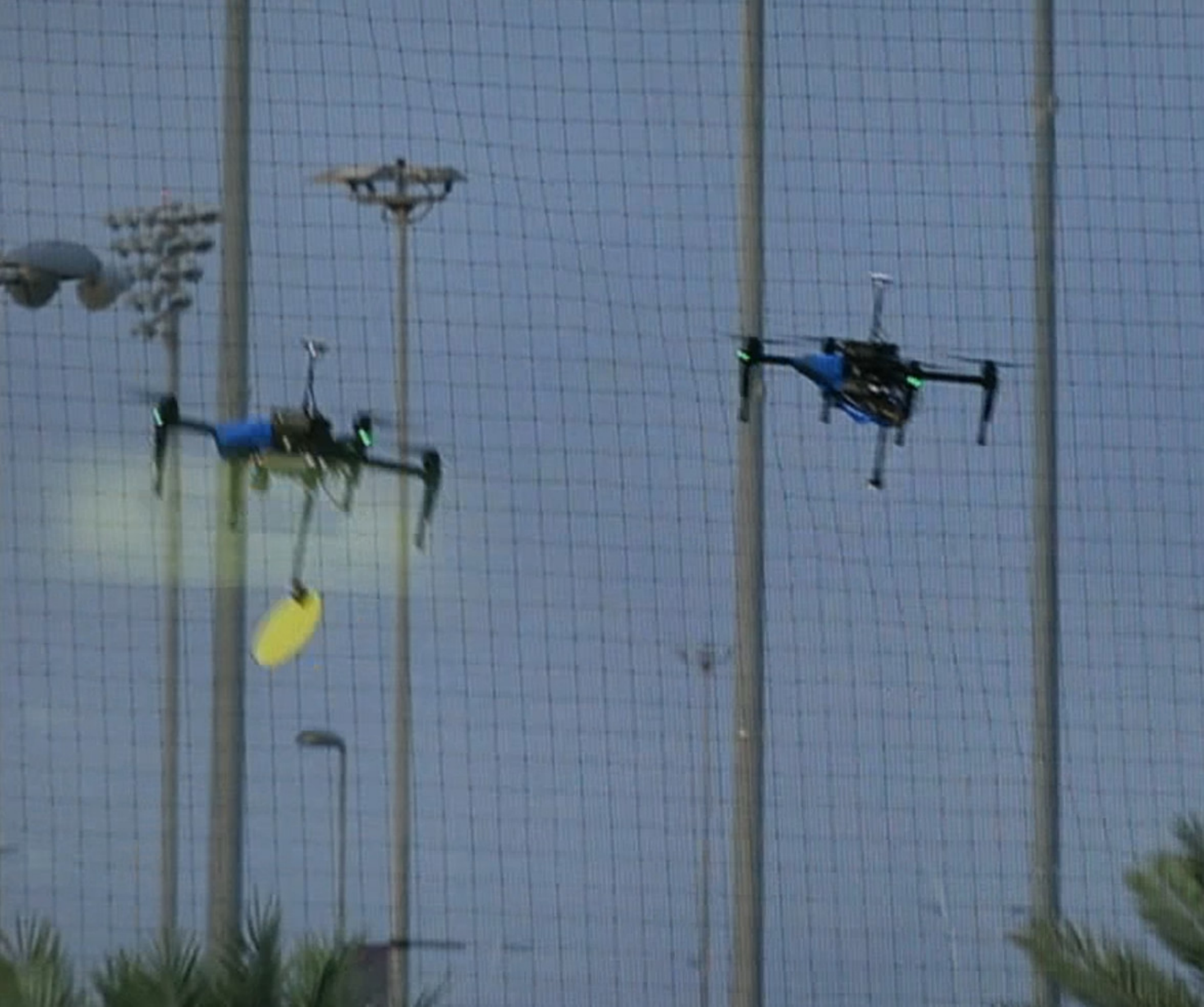}
  \caption{Treasure hunt at MBZIRC. Left: Three MAVs explore the arena to detect objects. Middle: Detected objects are picked with a flexible gripper. Right: The objects are delivered to a shared drop box, which requires team coordination.}
  \label{fig:teaser-new}
\end{figure}
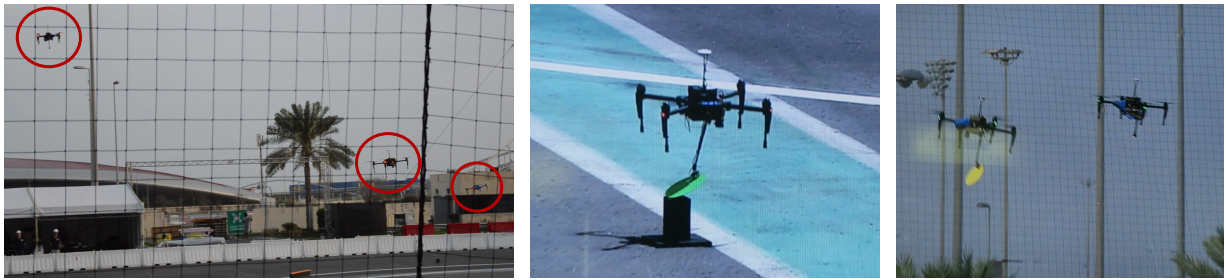

In Challenge 3, a team of up to three MAVs had to collect colored objects---some of them moving---distributed over the whole arena and deliver them to a drop box. Larger, heavier objects that had to be transported with a larger MAV or with two MAVs at the same time, were also placed within the arena. Although the score for heavy objects was the highest among all objects in the challenge (and was even doubled when using two MAVs for transport), none of the teams tried to actively move them. The task required coordinated exploration of the arena, aerial picking and transport of the objects, and detecting the drop zone to deliver the objects. Teams were provided with rough specifications of the objects in advance, \ie diameter, height above ground, maximum weight, maximum speed, and the possible colors. The drop box was specified by its approximate dimensions and coarse position in the arena. Nevertheless, the exact arena setup---including colored markings on the ground making color-based perception challenging---was not known in advance. \Cref{fig:teaser-new} shows the coordinated exploration of the arena, picking of an object, and transport to the drop box, shared by all MAVs.

In the Grand Challenge, the two MAV challenges and Challenge 2 "Operating a Steam Valve" with a ground robot were combined, \ie the tasks had to be performed in parallel in the same arena.

The entire competition was held under challenging outdoor conditions with changing overcast, temperatures of up to \SI{38}{\celsius}, and strong gusts of wind.
This article is an extended version of our own previous works \citep{ecmr2017_c1,ecmr2017_c3}.

\section{Related Work}
Autonomous landing of MAVs on moving vehicles is an active research topic.
For reasons of brevity, in this overview, we cover lines of research performing autonomous landing without cooperation between the robot and the ground vehicle.
\citet{Lee2012landing} demonstrated the viability of the task using visual servoing in order to maneuver above the moving pattern, and relying on a motion-capture system for external state estimation.
A similar system was demonstrated by \citet{Serra2016landing} who also use visual servoing but do not rely on a vision-based distance estimation to the target.
In comparison to our system, both approaches are evaluated with a slow or even static target.
\citet{Borowczyk2017landing} use a system of two cameras and filter the detections together with an IMU and GPS receiver mounted at the target.
They report landing velocities of up to \SI{50}{\kilo\metre\per\hour}.
Landing indoors on an inclined plane was achieved by \citet{Vlantis2015landing} who designed an adapted model-predictive controller to optimize the local trajectory in real-time.
However, due to the computational demand, optimization was done on an external base station and, thus, required a stable network connection.

Fast real-time trajectory planning and control is an active area of research.
\citet{Ezair2014} compare polynomial trajectory generation algorithms regarding the order, state constraints, and constraints on initial and final conditions.
\citet{Mueller2013_2} present a trajectory generator similar to our work.
It is also capable of attaining the full target state (position, velocity, and acceleration) and is real-time capable.
Analogue to our approach, they use jerk (respectively the rotational velocity $\omega$) as system input, but the convex optimization problem is solved numerically and generated trajectories are not time-optimal.

From other MBZIRC participants, we want to cite the early work by the team of the Korea Advanced Institute of Science and Technology \citep{KAIST2017landing} where landing on a larger platform at a velocity of \SI{0.75}{\metre\per\second} was demonstrated, but the visual detection was still simplified by a marker reflecting infrared light.
Another contribution from a challenger team is \citet{falanga2017}. Their system lands successfully on a platform moving with \SI{1.5}{\meter\per\second}. Nevertheless, no results from the actual competition are reported yet.
The team from Czech Technical University Prague also reports their approaches to landing during the MBZIRC 2017~\citep{baca2017}. Like us, they focus on target perception and model predictive control. In contrast to our work, the authors incorporate assumptions about the future movement of the car by modeling the a-priori known path of the target. They placed second in Challenge 1---landing on a moving target---and third in the Grand Challenge---a combination of the subchallenges---together with the University of Pennsylvania, the British University of Lincoln and the University of Padua.
Also \citet{Cantelli2017} and \citet{Battiato2017} from the University of Catania have presented their systems. They placed fourth in Challenge 1, in which it took them \SI{144}{\second} to land. In contrast to us, they employ a differential RTK-GPS for state estimation of the MAV, but similar to us a downward looking camera estimates the target position.
Like our approach, \citet{Acevedo2017} from the Center for Advanced Aerospace Technologies Seville in collaboration with the Robotics, Vision and Control Research Group make no assumptions about the target trajectory. Their successful landing, however, was on a slowed-down target driving at only \SI{5}{\kilo\meter\per\hour}.

\newcolumntype{P}[1]{>{\raggedright\arraybackslash}p{#1}}
\begin{table}[t]
\small
\caption{Comparison of the reported perception approaches for the landing challenge.}
\label{tab:Comparison_Perception}
\begin{center}
\begin{tabularx}{\columnwidth}{p{0.22\columnwidth}p{0.18\columnwidth}p{0.52\columnwidth}}
  \toprule
Team & Camera & Approach \\
  \midrule
\parbox[t]{0.22\columnwidth}{ETH Z\"{u}rich\\\citep{bahnemann2017eth_mbzirc}}
& 752$\times$480 @ \SI{50}{\hertz}
& One blob detector for platform, another detector for cross of pattern, tracker uses allocentric positional distribution over figure eight track\\[1.em]

\parbox[t]{0.22\columnwidth}{University of Prague\\\citep{baca2017}}
& \parbox[t]{0.15\columnwidth}{752$\times$480 @ \SI{50}{\hertz}}
& Full image adaptive threshold using altitude, circle detection, morphological operations to recognize four equally-sized quadrants around the cross\\[1.em]

\parbox[t]{0.22\columnwidth}{University of Z\"{u}rich\\\citep{falanga2017}}
& \parbox[t]{0.18\columnwidth}{Two cameras\\@ \SI{40}{\hertz}}\ 
& Thresholding and searching for largest polygon (platform), circle detection, cross detection\\[1.em]

\parbox[t]{0.22\columnwidth}{University of Bonn\\(Ours)}
& \parbox[t]{0.18\columnwidth}{Two cameras\\1920$\times$1080 @ \SI{40}{\hertz}}
& Inverse mapping via MAV attitude, symmetry segmentation, circle and line detection\\[0.5em]
\bottomrule
\end{tabularx}
\end{center}
\end{table}

In comparison to what other participants reported until the time of writing (see \reftab{tab:Comparison_Perception}), we can point out several key differences in our perception pipeline for both challenges. First, while all teams rely on a rather high detection rate of \SI{40}-\SI{50}{\hertz}, the reported camera resolution is at about a fourth of ours.

\begin{table}[t]
\small
\caption{Comparison of the reported target state estimation and MAV control methods.}
\label{tab:Comparison_Action}
\begin{center}
\begin{tabularx}{\columnwidth}{p{0.22\columnwidth}P{0.15\columnwidth}P{0.15\columnwidth}P{0.15\columnwidth}P{0.21\columnwidth}}
  \toprule
Team & \parbox[t]{0.15\columnwidth}{Target state\\estimation} & Assumptions on target motion & \parbox[t]{0.15\columnwidth}{Trajectory\\generation} & Trajectory tracking\\
\midrule
\parbox[t]{0.22\columnwidth}{ETH Z\"{u}rich\\\citep{bahnemann2017eth_mbzirc}}
& Holonomic motion model with steering and velocity noise
& Particle filter on figure eight
& Sampled minimum jerk motion primitives
& Nonlinear MPC with disturbance observer, reactive collision avoidance\\[1.em]

\parbox[t]{0.22\columnwidth}{University of Prague\\\citep{baca2017}}
& UKF with nonholonomic motion model
& Path curvature prediction with known map of figure eight
& MPC with \SI{8}{\second} prediction horizon
& SO(3) state feedback controller\\[3.em]

\parbox[t]{0.22\columnwidth}{University of Z\"{u}rich\\\citep{falanga2017}}
& EKF with nonholonomic motion model
& Constant velocity
& Sampled minimum jerk motion primitives
& Nonlinear controller\\[4.em]

\parbox[t]{0.22\columnwidth}{University of Bonn\\(Ours)}
& Complementary filter
& Constant velocity
& Nonlinear MPC for time-optimal trajectory generation
& Trajectory generation is running in closed loop and also serves as real-time trajectory tracker\\[0.5em]
\bottomrule
\end{tabularx}
\end{center}
\end{table}

We compare our approaches for target state estimation and MAV control with other participants in \reftab{tab:Comparison_Action}. All participants use cascaded position and attitude control loops, but in particular the approaches for position control vary.

Aerial manipulation has been investigated by multiple research groups as well.
\citet{morton2016}, for example, developed an MAV with manipulation capabilities for outdoor use. The MAV is equipped with a 3-DoF arm, which is operated during hovering, without object perception or autonomous flight.
An MAV with a 2-DoF robot arm that can lift relatively heavy weights was presented by \citet{kim2017}. The controller explicitly models the changes in the vehicle dynamics by attaching a heavy object. In contrast, we employ a trajectory generator that uses a very simple dynamics model with frequent replanning on top of a model-free attitude controller to achieve robustness against changes in flight dynamics.
\citet{ghadiok2011} built a lightweight quadrotor for grasping objects in indoor environments. Similar to our work, they employ a lightweight and compliant gripper to cope with uncertainties during grasping. Target objects are equipped with infrared beacons; we detect objects based on coarse color specifications.

The ETH Z\"urich MBZIRC team ``Electronic Treasure Hunters'' also describe their approach to solving the challenge \citep{baehneman2017}. They employ an electro-permanent magnetic gripper and color blob detection for visual servoing, placing second in both Challenge 3 and the Grand Challenge.
Also the joint team from CTU Prague, UPenn, and UoL \citep{Loianno2018} use an electro-permanent gripper and color blob detection for picking. Like the team from Z\"urich, they employ RTK-GNSS but omit additional visual-inertial odometry. They won the picking challenge with 56.154 points, followed by Z\"urich with 55.385 points, and our team with 53.846 points.

Groups of MAVs navigate relative to each other in the work of \citet{saska2017system} using video-based detections of the other MAVs. In our work, we rely solely on GPS readings---which are considered to have sufficient relative accuracy between the MAVs in an outdoor scenario---and separation of the MAV working areas.

Cooperative transport is a rather new field of research which has so far been covered only sporadically.
\citet{Michael2011} describe cooperative control laws in order to manipulate a given object with exactly three MAVs while keeping the object pose in a static equilibrium. However, this work assumes the object to be permanently connected to all MAVs with tows, \ie no picking and placing.
\citet{tagliabue2017} present a master-slave approach for transporting objects with two MAVs. The slave deploys an estimator for external force and torque that is based on a visual-inertial navigation system. A reference state (including pose, velocity and angular velocity) is generated which the MAV will try to reach in order to equilibrate the external influence.
Similarly, \citet{gassner2017dynamic} worked on a leader-follower control to move a bulky object with two MAVs, but the follower used visual perception in order to track the relative position of the leader and the relative position of a tow that was anchored to the object. Also, in contrast to the previous approach, the anchor point has to be known.

\begin{table}[t]
\small
\caption{Comparison of the approaches for the Treasure Hunt task reported by three of four teams that solved the task autonomously. It can be seen that all these teams selected relatively simple but robust components over complex ones.}
\label{tab:review_c3}
\begin{center}
\begin{tabularx}{\columnwidth}{p{0.20\columnwidth}P{.17\columnwidth}P{.11\columnwidth}P{.2\columnwidth}P{.2\columnwidth}}
  \toprule
  Team & Hardware & State estimation & Object perception & Coordination\\
  \midrule
  \parbox[t]{0.2\columnwidth}{CTU Prague,\\UPenn, UoL\\\citep{Loianno2018}}
&
  DJI F550 with Pixhawk FCU, flexible gripper with passive joints (base--linear--ball--end effector) and electro-permanent magnet, Hall effect sensors
&
  Differential RTK-GNSS, TeraRanger lidar for altitude
&
  \parbox[t]{0.2\columnwidth}{\raggedright
  Rolling shutter camera\\(1920$\times$1080),\\color blob detection, RGB lookup table for HSV Gaussian mixture model, shape constraints}
&
  Area decomposition (equally-sized) for exploration, altitude separation, broadcast of odometry, avoidance of received descent corridors, time slotting at drop box as fallback
  \\[2.em]
  \parbox[t]{0.2\columnwidth}{ETH Z\"urich\\\citep{bahnemann2017eth_mbzirc}} 
&
  AscTec Neo, flexible gripper with passive joints (base--linear--ball--end effector) and electro-permanent magnet, Hall effect sensors
&
  RTK-GPS, visual-inertial odometry with VI-Sensor
&
  \parbox[t]{0.2\columnwidth}{\raggedright
  Global shutter camera\\(2048$\times$1536),\\blob detection by thresholding in CIE L*a*b color space, shape constraints}
&
  Area decomposition (drop box focused) for exploration and picking, altitude separation, broadcast of odometry, avoidance of received MAV/UGV positions, no explicit fallback at drop box
  \\[1.em]
  \parbox[t]{0.22\columnwidth}{University of Bonn\\(Ours)}
&
  DJI Matrice 100, flexible gripper with passive joints (base--ball--linear--ball--end effector) and electromagnet, gripper contact detection by push button
&
  GPS with corrections from ground station, Lidar Lite v3 for altitude
&
  \parbox[t]{0.2\columnwidth}{\raggedright
  Global shutter camera\\(1920$\times$1200),\\ground plane alignment, color blob detection, lookup table for HSV Gaussian mixture model, shape and color/background constraints}
&
  Area decomposition (drop box focused) for exploration and picking, altitude separation, broadcast of MAV states and objects, mutual exclusive access to drop zone, time slotting as fallback
  \\[.5em]
\bottomrule
\end{tabularx}
\end{center}
\end{table}

In \reftab{tab:review_c3}, we summarize the selected approaches for picking from MBZIRC teams found in the literature so far.
These include the teams that reached the first three places in the Treasure Hunt subchallenge: The joint team from CTU Prague, UPenn, and the UoL \citep{Loianno2018}, the Electronic Treasure Hunters from ETH Z\"urich \citep{bahnemann2017eth_mbzirc}, and our team NimbRo.
It can be seen that many approaches and design principles of these successful teams follow similar ideas: The favored hardware and software components are relatively simple to reduce the system complexity.
The similarity in the general design decisions of the individually developed systems---from perception algorithms that are robust to noise and do not require much pretraining to MAV separation instead of complex multi agent planning---backs our assumption that simplicity is key for the operation of still complex systems under competition conditions.

\section{System Setup}
\label{sec:System_Setup}
\begin{figure}[t]
  \centering
  \resizebox{1.0\linewidth}{!}{
    \begin{tikzpicture}[auto]
      \definecolor{red}{rgb}{0.7,0.0,0.0}
      \node[anchor=south west,inner sep=0] (image) at (0,0) {\includegraphics[width=0.58\linewidth]{ROB-17-0130_fig3a-eps-converted-to.pdf} \includegraphics[width=0.51\linewidth]{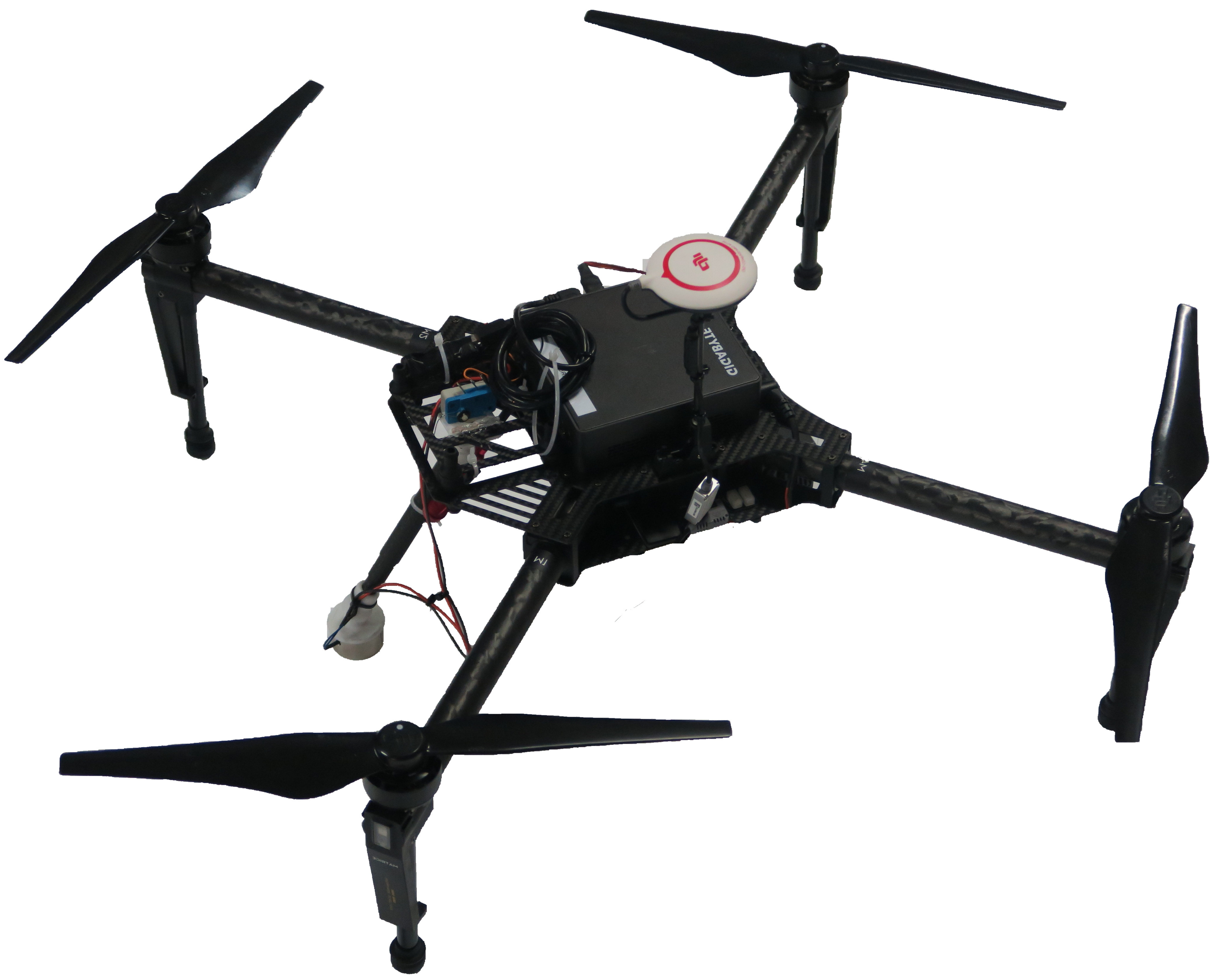}};
      \begin{scope}[x={(image.south east)},y={(image.north west)},font=\sffamily,every node/.style={align=center},line/.style={red, line width=1pt},box/.style={rectangle,draw=red,inner sep=0.3333em,thick},every node/.style={align=center,text height=1.5ex,text depth=.25ex,text centered},]
	\node[box,anchor=south west,rectangle callout,callout relative pointer={(-0.05,-0.02)}] (cam01) at (0.30, 0.05) {Magnetic feet};
	\node[box,anchor=south west,rectangle callout,callout relative pointer={( 0.07,-0.04)}] (cam03) at (0.05, 0.05) {Switches};
	\node[box,anchor=south west,rectangle callout,callout relative pointer={(-0.03, 0.01)}] (cam02) at (0.30, 0.90) {GNSS};
	\node[box,anchor=south west,rectangle callout,callout relative pointer={( 0.01, 0.01)}] (cam04) at (0.08, 0.38) {Front camera};
	\node[box,anchor=south west,rectangle callout,callout relative pointer={(-0.07, 0.05)}] (cam05) at (0.35, 0.32) {Bottom camera};
	\node[box,anchor=south west,rectangle callout,callout relative pointer={( 0.04,-0.04)}] (cam06) at (0.08, 0.75) {Computer};
	\node[box,anchor=south west,rectangle callout,callout relative pointer={( 0.03,-0.06)}] (cam07) at (0.70, 0.80) {GNSS};
	\node[box,anchor=south west,rectangle callout,callout relative pointer={( 0.02,-0.03)}] (cam08) at (0.64, 0.70) {Computer};
	\node[box,anchor=south west,rectangle callout,callout relative pointer={(-0.03, 0.04)}] (cam09) at (0.80, 0.30) {Color camera};
	\node[box,anchor=south west,rectangle callout,callout relative pointer={(-0.07,-0.06)}] (cam10) at (0.88, 0.70) {Laser sensor};
	\node[box,anchor=south west,rectangle callout,callout relative pointer={( 0.05, 0.04)}] (cam11) at (0.55, 0.42) {Ball joint};
	\node[box,anchor=south west,rectangle callout,callout relative pointer={( 0.07, 0.03)}] (cam11) at (0.55, 0.33) {Rod};
	\node[box,anchor=south west,rectangle callout,callout relative pointer={( 0.02, 0.02)}] (cam14) at (0.55, 0.24) {Magnet};
      \end{scope}
    \end{tikzpicture}
  }
  \caption{Closeup of our MAVs. Left: The MAV for the landing task is equipped with two cameras, magnetic feet and switches to detect landings. Right: The Treasure Hunt MAV is equipped with a down-pointing color camera for object and drop box detection. Objects are picked with an electromagnetic gripper on a telescopic rod and a ball joint. A small laser sensor measures the distance to the ground. All calculations are performed by a lightweight, but powerful onboard PC.}
  \label{fig:mav_closeup}
\end{figure}

Our MAVs, depicted in \cref{fig:mav_closeup}, are based on the DJI Matrice 100 quadcopter platform. This platform is designed for research and development---and consequently offers easy integration of custom hard- and software.
We equipped the platforms with small but powerful Gigabyte GB-BSi7T-6500 onboard PCs with an Intel Core i7-6500U CPU running at \SI{2.5/3.1}{\giga\hertz} and \SI{16}{\giga\byte} of RAM.
For allocentric localization and state estimation, we employ the filter onboard the DJI flight control that incorporates a global navigation satellite system (GNSS) as well as barometric and IMU data.
To avoid electromagnetic interference between components---in particular USB 3.0 and GPS---the core of our MAVs is wrapped in electromagnetic shielding material. This significantly increases the system stability.

\begin{figure}[ht]
  \centering
  \resizebox{1.0\linewidth}{!}{
    \begin{tikzpicture}[font=\sffamily,>={Stealth[inset=0pt,length=4pt,angle'=45]}]]
      \tikzset{content_node/.append style={minimum size=1.5em,minimum width=6em,draw,align=center,rounded corners,scale=0.65}}
      \tikzset{label_node/.append style={scale=0.5}}
      \tikzset{group_node/.append style={scale=0.65}}

      \definecolor{red}{rgb}     {0.5,0.0,0.0}
      \definecolor{green}{rgb}   {0.0,0.5,0.0}
      \definecolor{blue}{rgb}    {0.0,0.0,0.5}
      \definecolor{grey}{rgb}    {0.5,0.5,0.5}
      \definecolor{darkblue}{rgb}    {0,0,0.25}

      \draw[thick, rounded corners, densely dotted, grey!10!white,fill] (-7.0,1.9625) -- (-1.5,1.9625) -- (-1.5,4.4375) -- (-7.0,4.4375) -- cycle;
      \draw[thick, rounded corners, densely dotted, grey!10!white,fill] (7.0,1.9625) -- (1.5,1.9625) -- (1.5,4.4375) -- (7.0,4.4375) -- cycle;

      \node(Cameras)[content_node,darkblue,text=black,fill=green!15!white] at (-6.0,3.5) {2x Cameras};
      \node(Camera)[content_node,red,text=black,fill=green!15!white] at (6.0,3.5) {Camera};
      \node(Pattern_Detection)[content_node,darkblue,text=black,fill=blue!15!white] at (-3.0,3.5) {2x Pattern\\Detection};
      \node(Object_Detection)[content_node,red,text=black,fill=blue!15!white] at (3.0,3.5) {Object\\Detection};
      \node(Filter)[content_node,fill=blue!15!white] at (0.0,3.5) {Filter};
      \node(State_Machine)[content_node,minimum height=5em,fill=blue!15!white] at (0.0,1.9375) {State Machine};
      \node(Trajectory_Generation)[content_node,fill=blue!15!white] at (0.0,0.45) {Trajectory Generation};
      \node(Operator)[content_node,fill=green!15!white] at (-3.0,1.625) {Operator};
      \node(Foot_Switches)[content_node,darkblue,text=black,fill=green!15!white] at (-3.0,2.875) {Foot Switches};
      \node(Gripper_Switch)[content_node,red,text=black,fill=green!15!white] at (3.0,2.875) {Gripper Switch};
      \node(Magnet)[content_node,red,text=black,fill=red!15!white] at (3.0,2.275) {Magnet};
      \node(Lidar)[content_node,red,text=black,fill=green!15!white] at (3.0,4.125) {Lidar};

      \node(MAV)[content_node,fill=red!15!white] at (0.0,-1.25+0.5) {MAV};
      \node(GNSS)[content_node,fill=green!15!white] at (-3.0,-1.0+0.5) {GNSS};
      \node(IMU)[content_node,fill=green!15!white] at (-3.0,-1.5+0.5) {IMU};

      \draw[->, thick,densely dotted,darkblue] (Cameras) -- node[label_node,midway,below] {2x\SI{40}{\hertz}} node[label_node,midway,above] {Image} (Pattern_Detection);
      \draw[->, thick,densely dotted,red] (Camera) -- node[label_node,midway,below] {\SI{40}{\hertz}} node[label_node,midway,above] {Image} (Object_Detection);
      \draw[->, thick,densely dotted,darkblue] (Pattern_Detection) -- node[label_node,midway,below] {2x\SI{40}{\hertz}} node[label_node,midway,above] {3D Position} (Filter);
      \draw[->, thick,densely dotted,red] (Object_Detection) -- node[label_node,midway,below] {\SI{40}{\hertz}} node[label_node,midway,above] {3D Position} (Filter);
      \draw[->, thick] (Filter) -- node[label_node,midway,left,align=center] {2D Pattern/Object\\Position \& Velocity\\Height Correction} node[label_node,midway,right,align=center] {\SI{50}{\hertz}} (State_Machine);
      \draw[->, thick] (State_Machine) -- node[label_node,midway,left] {\SI{50}{\hertz}} node[label_node,midway,right,align=left] {3D~Target~Position\\3D~Target Velocity\\Yaw} (Trajectory_Generation);
      \draw[->, thick] (Trajectory_Generation) -- node[label_node,midway,left] {\SI{50}{\hertz}} node[label_node,midway,right,text width=1cm] {Roll Pitch Climbrate Yawrate} (MAV);
      \draw[->, thick] (MAV) -- (2.25,-1.25+0.5) -- node[label_node,midway,left] {\SI{100}{\hertz}} node[label_node,midway,right,text width=2.8cm] {3D~Position 3D~Velocity 3D~Acceleration Yaw} (2.25,0.45) -- (Trajectory_Generation);
      \draw[->,thick] (2.25,0.45) -- (2.25,1.625 -| 2.25,0.45) node[label_node,midway,left] {\SI{100}{\hertz}} node[label_node,midway,right,align=left] {3D~Position\\3D~Velocity\\Yaw} -- (State_Machine.east |- Operator);

      \draw[->, thick] (GNSS) -- (GNSS -| -1.5,-1.0)  -- (-1.5,-1.0 |- MAV.175) -- node[label_node,midway,left] {} node[label_node,midway,right] {} (MAV.175);
      \draw[->, thick] (IMU) --  (IMU -| -1.5,-1.0) --  (-1.5,-1.0 |- MAV.185)  -- node[label_node,midway,left] {} node[label_node,midway,right] {} (MAV.185);
      \draw[->, thick] (Operator) --  node[label_node,midway,below,align=center] {Start/Stop\\Command} node[label_node,midway,left] {} node[label_node,midway,right] {} (State_Machine.west |- Operator);
      \draw[->, thick,densely dotted,darkblue] -- (Foot_Switches) -- (Foot_Switches|-State_Machine.160) -- node[label_node,midway,above] {Pushed?}  node[label_node,midway,below] {\SI{50}{\hertz}}(State_Machine.160);
      \draw[->, thick,densely dotted,red] -- (Gripper_Switch) -- node[label_node,midway,above] {Pushed?} node[label_node,midway,below] {\SI{50}{\hertz}} (State_Machine.45 |- Gripper_Switch) --  (State_Machine.45);
      \draw[->, thick,densely dotted,red] (Lidar) -- node[label_node,midway,below] {\SI{50}{\hertz}} node[label_node,midway,above] {Distance} (0.0,4.125) -- (Filter);
      \draw[->, thick,densely dotted,red] (State_Machine.east|-Magnet) -- node[label_node,midway,below] {\SI{10}{\hertz}} node[label_node,midway,above] {On/Off} (Magnet);

      \node(Landing_Label)[group_node,anchor=south west] at (-6.0,4.4375) {\textbf{Landing}};
      \node(Picking_Label)[group_node,anchor=south west] at ( 1.5,4.4375) {\textbf{Picking}};
    \end{tikzpicture}
  }
  \caption{Structure of our method. Green boxes represent external inputs like sensors, blue boxes represent software modules, and the red boxes indicate the MAV hardware and the electromagnet. Blue lines depict components used for the Landing challenge and red lines depict components for the Treasure Hunt challenge. All software components use ROS as middleware. Position, velocity, acceleration, and yaw are allocentric.}
  \label{fig:Structure}
\end{figure}
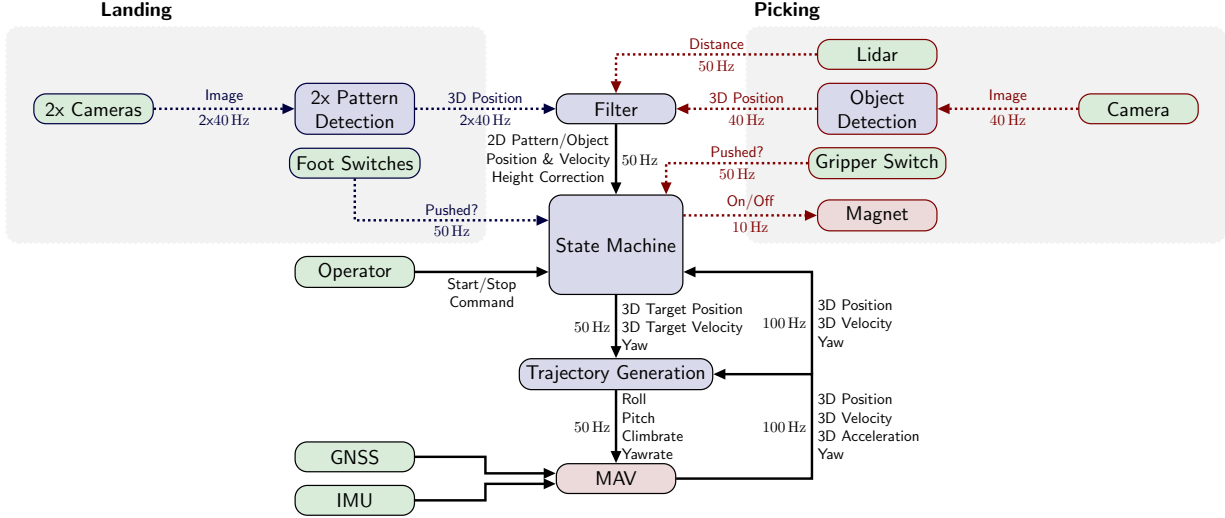

In addition to the basic MAV platform, we added task-specific equipment to the MAVs. \Cref{fig:Structure} gives an overview of the information flow in our system.

One MAV is designated to accomplishing the landing task.
The landing pattern is perceived by two Point Grey BFLY-U3-23S6M-C grayscale cameras with 2.3\,MP. The first camera---equipped with a Lensagon BF2M2020S23 wide-angle lens with an apex angle of \SI{195 x 195}{\degree}---is pointing downwards.
To facilitate the detection of a far-away pattern and to keep it in the field of view (FoV) during descent on a glide path, the second camera--- equipped with a Lensagon CY04818 lens with an apex angle of only \SI{69 x 85}{\degree}---is pointing \SI{30}{\degree} into forward direction.
Both cameras capture 40 frames per second, resulting in 80 frames per second in total.
We replaced the landing feet of the MAV with strong magnets with a total rated force of \SI{860}{\newton} and a high friction foam rubber coating to keep it in place after landing on the moving target with a ferromagnetic surface. A successful landing is detected by eight micro switches attached to the landing feet. The switches are individually connected to an Arduino Nano v3.0 that serves as a bridge to our onboard computer.

Four MAVs are equipped for the Treasure Hunt task---one of them as a backup.
For object and drop box detection, we employ a downward facing Point Grey BFLY-U3-23S6C-C color camera with a Lensagon CY04818 lens, providing an apex angle of \SI{69 x 85}{\degree}.
A small Garmin Lidar Lite v3 measures the distance to the ground to allow for exact, drift-free vertical navigation close to the ground.
Our gripper is an electromagnet on the end of a telescopic rod that is attached to a ball joint. The rod is passively extended to its full length of \SI{31}{\centi\meter} by gravity and can be shortened up to \SI{19}{\centi\meter} when in contact with an object.
A switch detects shortening of the rod.
Two dampers avoid fast oscillations of the ball joint, while still allowing the rod to align with the gravity vector.
The gripper weighs \SI{220}{\gram}, including mounting and electronics.
All MAVs for this task are similar in hardware and software---also most of the configuration is derived from a single MAV ID---to simplify the handling of multiple robots in stressful competition situations.

For both challenges, to make all components easily transferable between the test area at our lab and also different arenas on site, we define all coordinates $(x, y, z, \theta_{yaw})$ in a field-centric coordinate system.
The center and orientation of the current field were broadcasted by a base station PC to all active MAVs---and the ground robot in the Grand Challenge.
Furthermore, the base station PC, an Intel NUC equipped with a DJI N3 module, constantly measures its GNSS position and broadcasts position correction offsets to eliminate larger position deviations caused by atmospheric effects.
In contrast to other teams, we did not use advanced satellite-based localization methods like Real-Time Kinematic positioning (RTK-GPS) that need multiple GPS antennas on the MAVs.

Initially, we placed the onboard computer on top of our MAV as the battery compartment of the Matrice 100 is below the MAV. Thus, it was placed directly under the GPS antenna.
During our first experiments, we observed severe instability issues with the GPS signal once the computer was started---in some cases inhibiting to take off---due to electromagnetic interference.
Hence, we switched the battery compartment from below with the computer on top to reduce the interference.

The communication among the MAVs and the base station is conducted over a stationary WiFi infrastructure.
For robustness, we employ a UDP protocol that we developed for connections with low bandwidth and high latency~\citep{nimbro_networking}.\footnote{\url{github.com/AIS-Bonn/nimbro_network}}

\section{Visual Perception}
\label{sec:Visual_Perception}
In both challenges, we employed solely cameras to perceive and track the targets, \ie the landing pattern for the landing task, and pickable objects and the drop box for the Treasure Hunt.

Common to all further steps in our image processing pipeline is that our detectors operate on a bird's-eye representation of the field\footnote{Please note that a) the camera setup is not aligned to the ground plane and b) the camera may have an arbitrary orientation during rapid maneuvers, thus, a prior image transform is necessary.}, taking the MAV attitude into account.

For defining the bird's-eye transformation, let $\hat{r}_z$ be the IMU gravitational vector in camera coordinates. The rotation matrix
\begin{align}
\hat{R} &= \left(\hat{r}_x, \hat{r}_y, \hat{r}_z\right)\\
\mbox{with } \hat{r}_x &= \begin{pmatrix} 0 & 1 & 0\end{pmatrix}^T \times \hat{r}_z,\\
\hat{r}_y &:= \hat{r}_z \times \hat{r}_x,
\end{align}
describes the rotation from the camera frame into a frame where the image plane is aligned with the ground plane, \ie the matrix
\begin{equation}\label{eq:pixel_transform}
M = K_g \hat{R} K_c^{-1} 
\end{equation}
with accordingly chosen camera matrices $K_g, K_c$ describes a pixel coordinate transform into a bird's-eye representation via homogeneous coordinates. $K_c$ is given by the camera intrinsics. $K_g$ is a task specific matrix defined at description of the individual perception modules.
Finally, taking the lens distortion into account, we arrive at
\begin{equation}\label{eq:groundplane_transform}
(u, v) \mapsto M \begin{pmatrix} d(u, v) \\ 1 \end{pmatrix}
\end{equation}
with an invertible radial-tangential lens undistortion function $d$ operating on the image coordinates $(u, v)$.
The second part of the mapping in \cref{eq:groundplane_transform} is linear-projective and can be computed very efficiently, in particular on rectangular regions when not the entire image has to be traversed.

\subsection{Landing Pattern Detection}
\label{sec:Landing_Pattern_Detection}
\begin{figure}[t]
  \centering
  \includegraphics[trim=00mm 00mm 00mm 00mm,clip,width=0.8\columnwidth]{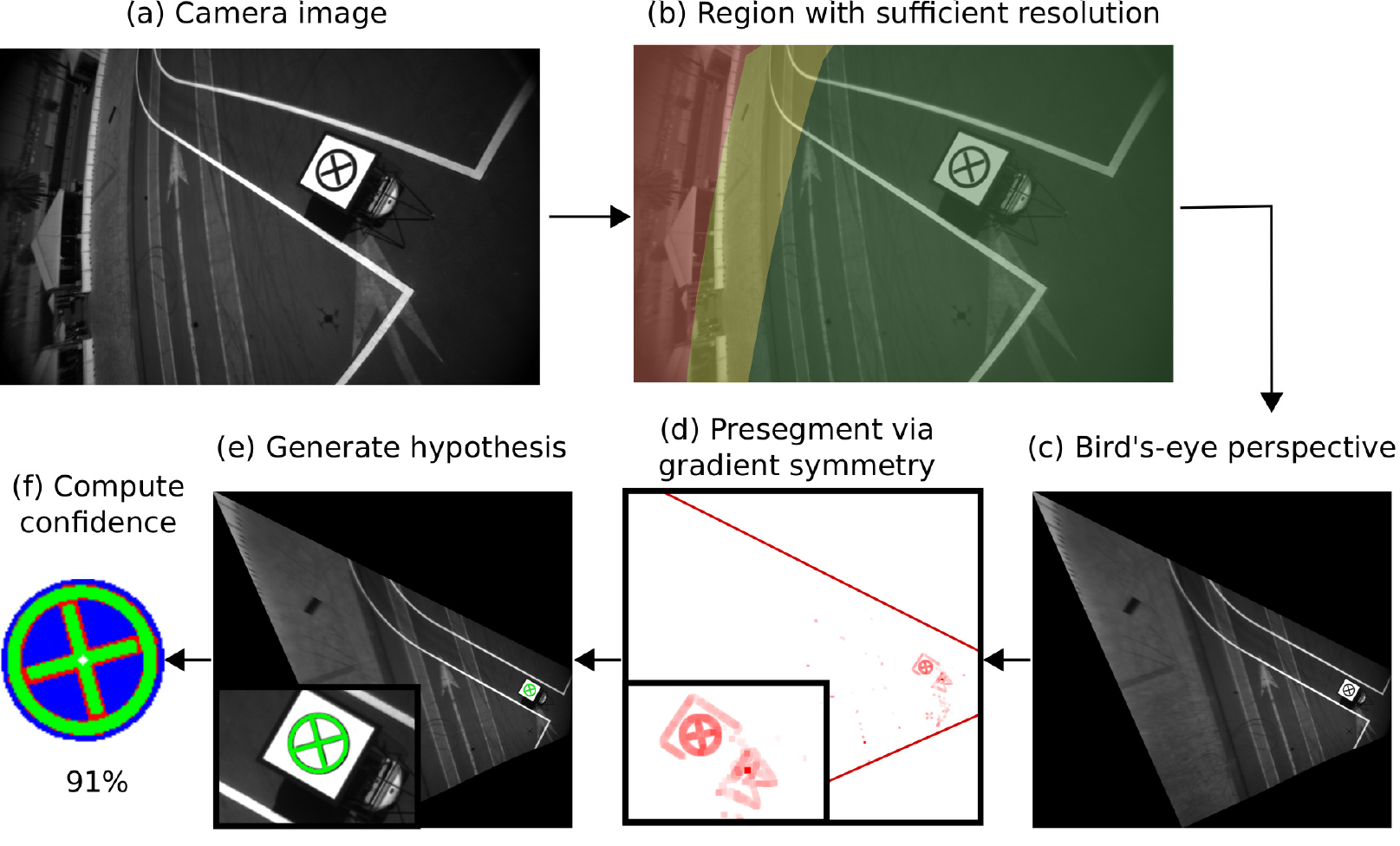}
  \caption{Landing pattern detection from the front camera during the competition: (a) original camera image, (b) image regions with sufficient resolution for pattern detection (green), insufficient resolution (yellow), and regions above the ground plane (red), (c) bird's-eye representation from a region with (mostly) sufficient resolution, (d) results from symmetry segmentation shown dilated for better visibility, (e) initial hypothesis in green, (f) confidence computation via pattern-detection overlay: green and blue denote correct pixels, red incorrect ones.}
  \label{fig:image_proc_pipeline_c1}
\end{figure}

When detecting the landing pattern with a camera, one must consider two main objectives:
\begin{compactitem}
\item the detection process itself should be low-latency and yield accurate results and
\item the detection range should be as wide as possible.
\end{compactitem}

We developed a multi-stage detection pipeline (see \cref{fig:image_proc_pipeline_c1}): The camera image is transformed to the bird's-eye representation (see \cref{fig:image_proc_pipeline_c1} (c)), a segmentation step detects line-like structures within the image (see \cref{fig:image_proc_pipeline_c1} (d)) that are processed via a circular Hough transform in order to generate a number of hypotheses and their respective confidence.

Let $r$ be the radius of the landing pattern in meters. In order to maximize the detection range the MAV's flying height $h$, obtained by relative barometric measurements, and its attitude, represented by the IMU gravitational vector $\hat{r}_z$ within camera coordinates, is taken into account. For the bird's-eye transform let $K_g$ in \cref{eq:pixel_transform} be
\begin{equation}
K_g = diag\left( \tfrac{2r \cdot \rho }{h}, \tfrac{2r \cdot \rho}{h}, 1\right) ,
\end{equation}
where $\rho$ is the desired resolution of the pattern (for $\rho$ we choose $60$ pixels or lower, depending on the maximum possible resolution of the pattern in the original camera image). It remains to compute those image regions that yield at least the resolution $\rho_{min}$ required for detection (chosen as $20$ pixels) when transformed by $M$. To this end, a point grid with a fixed stride of $k$ pixels in the original image is mapped via $M$ and the distance between neighboring points is computed. If this distance is below $\tfrac{2 r k}{\rho_{min}}$, the resolution of the corresponding image patch is too low for detection. The maximum rectangular region in the camera image containing as many grid points with sufficient resolution after mapping as possible (\ie the maximum rectangle enclosing as much of the green region and none of the red one in \cref{fig:image_proc_pipeline_c1} (b)) is computed via a heuristic approach and the resulting region of the camera image is subsequently transformed.

\begin{figure}[t]
  \centering
  \includegraphics[width=0.8\columnwidth]{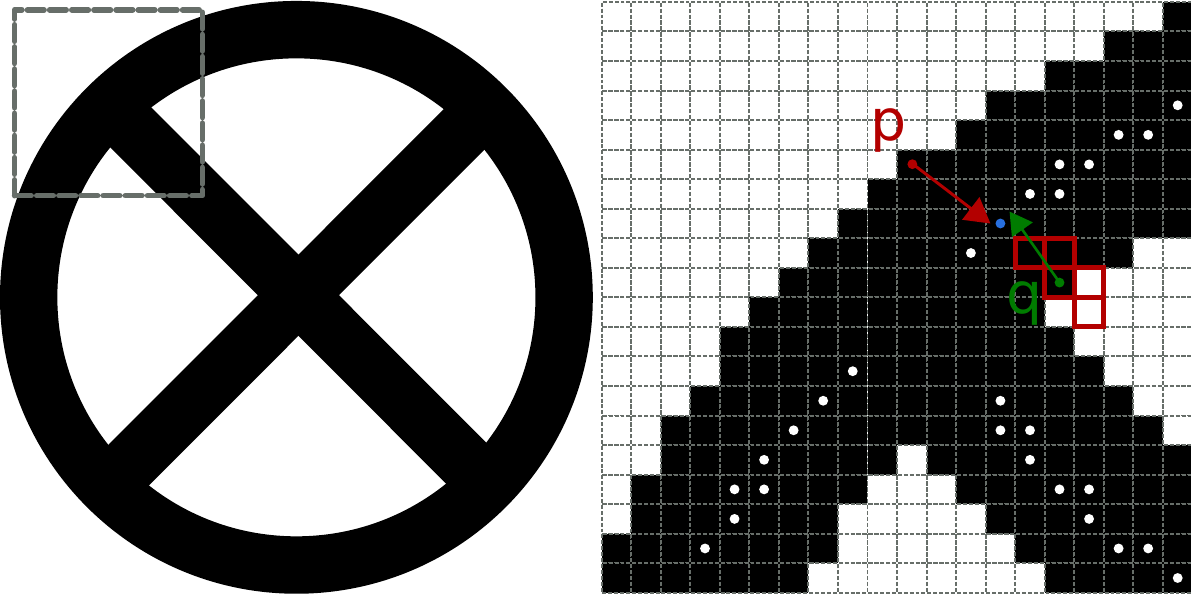}
  \caption{Fast symmetry detection scheme: Starting from pixel $p$ (red dot) a few pixels in direction of its negative gradient are traversed (red borders) in order to find a pixel with approximately converse gradient $q$ (green dot). Their center (blue dot) is incremented in the resulting symmetry image. Repeating this for every pixel with significant gradient magnitude reveals the pattern centerline (white dots).}
  \label{fig:landingPattern}
\end{figure}

In order to reach the desired latency, the now following detection pipeline is strongly tailored to the particular pattern used to identify the target. It is a crossed black circle on white ground (see \reffig{fig:landingPattern}). Its size and line width were known beforehand. Hence, the respective size in image dimensions is approximately known due to height measurements of the MAV. Regarding the high contrast, we first segment the line by means of a fast symmetry detection similarly to the method in \cite{houben2015rtip}. The detection yields a symmetry image of the same size as the transformed image from the bird's-eye camera and contains a pixelwise measure for the presence of the pattern line. A gradient image is computed with a Sobel filter and only a quantile of the image pixels with largest gradient magnitude are propagated to the next stage. For each remaining pixel position $p$, a line search in direction of the negative gradient is performed (the negative gradient should be oriented from bright to dark regions, thus, hopefully pointing to the centerline of the line pattern). Please refer to \reffig{fig:landingPattern} for a geometric illustration of this algorithm. The line search is efficiently implemented by a Bresenham-like traversion scheme and compares $p$ to a number of pixels approximately at the predicted line width. If these searched pixels contain a gradient pixel with nearly opposite orientation $q$, the pixel in between $\frac{1}{2}(p+q)$ is incremented in the resulting symmetry image. 

A circular Hough transform on this symmetry image provides a number of hypothesis. As the approximate diameter of the circle in image dimensions is known, the range of the circle radius is already highly restricted. These hypothesis are subsequently confirmed if two lines with a central perpendicular intersection are detected within. Again, these are detected via a fast Hough transform for lines. The intersection yields the target position with subpixel accuracy. Please also note that detecting the two crossing lines reveals the pattern's orientation and allows us to create a synthetic overlay over the image region. This is subsequently used to derive a confidence measure by thresholding the potential region of the bird's-eye camera image with the expected quantile of dark versus white pixels and computing the ratio of the thresholded pixels at the correct location (see \reffig{fig:image_proc_pipeline_c1} (f)).

After a sufficiently confident detection only a rectangular image region around the previous position is considered in the following iterations, reducing the algorithm to steps (c) -- (f)  from \cref{fig:image_proc_pipeline_c1}. In order to follow the pattern as long as possible, the requirement on the minimum image resolution of the pattern is ignored in this tracking stage.

\subsection{Pickable Object Detection}
\label{sec:Pickable_Object_Detection}
\begin{figure}[t]
  \centering
  \includegraphics[trim=00mm 00mm 00mm 00mm,clip,width=0.8\columnwidth]{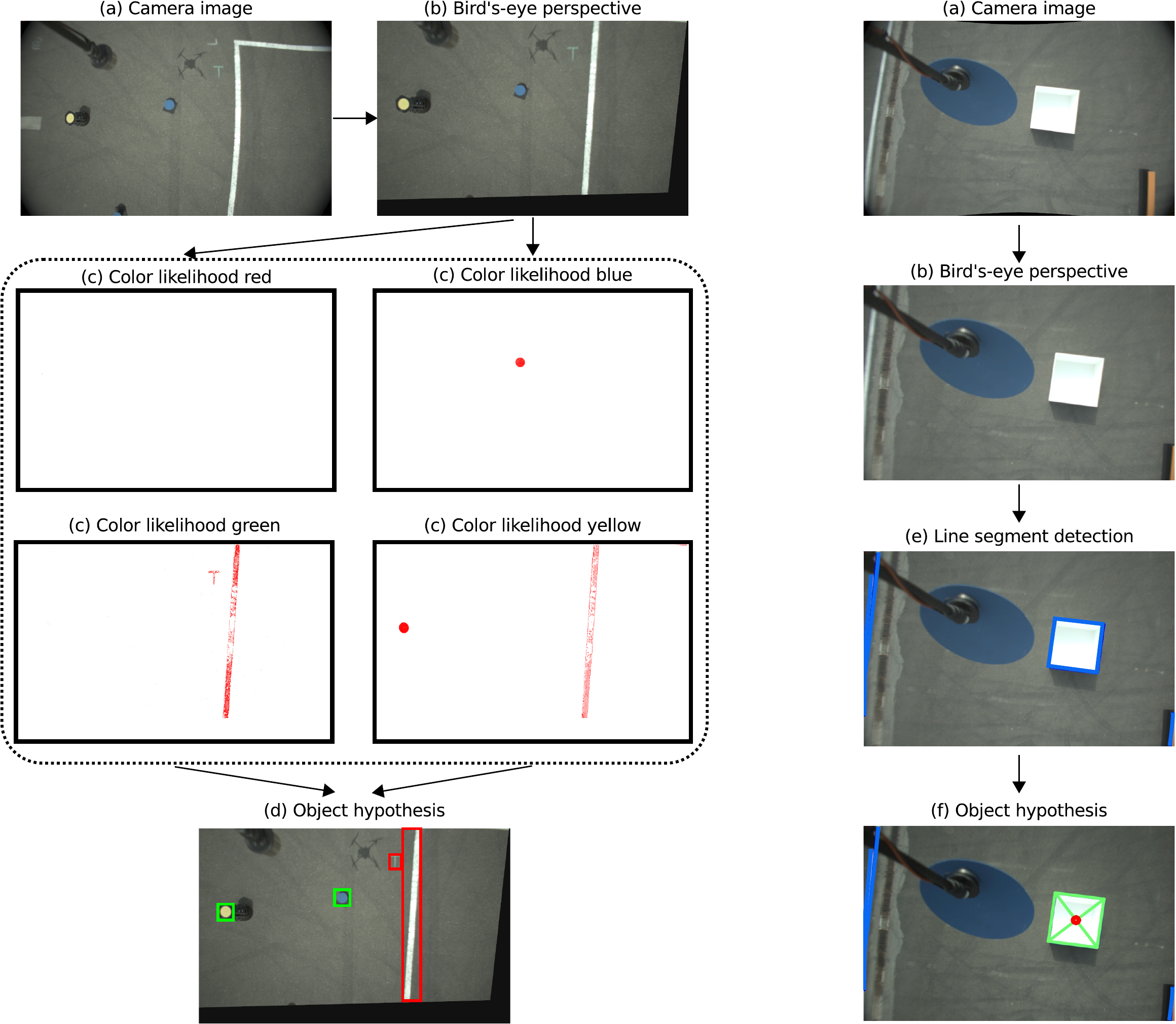}
  \caption{Overview of the image processing pipeline during the competition: \\
  Object detection (left): (a) original camera image, (b) undistorted bird's-eye representation, (c) color likelihood images, (d) detection hypothesis in green (accepted) and red (discarded).\\
  Drop box detection (right): (a) - (b) as before, (e) line segments, (f) final detection hypothesis.
  \label{fig:image_proc_pipeline_c3}}
\end{figure}

The detection of pickable objects demands for specific characteristics of the underlying algorithms as well.
The mission plan requires robust perception in two phases: i) when sweeping the field at a very high speed, the MAVs have to detect and track the pickable objects with low latency; ii) after arrival in the drop zone, the box has to be reliably detected and tracked during approach.

The challenge rules defined two kinds of pickable objects: thin ferromagnetic disks with a diameter of \SI{20}{\centi\meter} and a maximum weight of \SI{500}{\gram} and thin rectangular objects with a size of \SI{200 x 20}{\centi\meter} and a maximum weight of \SI{2}{\kilo\gram}. The former were colored in red, green, blue, and yellow while the latter were exclusively orange. Since little detail was given beforehand about the competition arena and, thus, possible distracting objects, the detection algorithm is based on both color and shape information where the specific color was supposed to be trained quickly on-site when the actual objects were available. The learned color distribution is also able to model the effect of different lighting conditions and reflective object surfaces.

The camera image is scaled down depending on the MAV altitude and the bird's-eye perspective transform is applied in order to account for its attitude (see \cref{fig:image_proc_pipeline_c3} (b)).
A pixelwise transform is computed assigning the likelihood of belonging to one of the relevant colors which results in one likelihood image per color (see \cref{fig:image_proc_pipeline_c3} (c)). A blob detection method identifies the connected regions which are then filtered by several shape (aspect ratio, convexity, size) and color (average likelihood, contrast to background) criteria (see \cref{fig:image_proc_pipeline_c3} (d)). 

The initial image transform serves to simplify the detection problem but also to limit the computational burden.
Let therefore be $r$ the magnitude of the shorter side of a detectable object in meters and $h$ the MAV altitude obtained by relative barometric and laser range measurements. The image is scaled by 
\begin{equation}
s = \mathop{max} \left( \mathop{min}\left(30 \cdot \tfrac{h}{rf}, 1.\right), 0.1 \right) \mbox{,}
\end{equation}
where $f$ is the camera focal length. After scaling with $s$, the object should have a size of about 30 pixels. However, the image is only scaled down (not scaled up) to this end. For later convenience, $s$ is rounded to the first position after the decimal point. Hence, $s$ may only obtain one of ten possible values. 

We transform the camera image into the bird's-eye perspective by inserting $K_g := s K_c$ into \cref{eq:pixel_transform}.
For efficiency, we precompute the pixelwise lookup table $d$ for each of the ten possible rounded scale factors $s$.

For detection processing, the color image is transformed into HSV space. As pixelwise color likelihood we use a max-mixture of Gaussians model:
\begin{equation}
\mathop{max}_i \left( exp\left\{ - \left(x - c_i\right)^T diag\left( \sigma_h^2, \sigma_s^2, \sigma_v^2 \right) \left(x - c_i\right) \right\} \right),
\label{eq:max_mixture}
\end{equation}
where $x = (x_h, x_s, x_v)^T$ denotes the three channel pixel value, $c_i = (c_{i,h}, c_{i,s}, c_{i,v})^T$ are a number of trained prototype pixel values, and $\sigma_h, \sigma_s, \sigma_v$ represent three hyperparameters. During training, the channel-wise mean of all pixels from a manually labeled object detection is computed and stored in a single prototype pixel value $c_i$. In order to efficiently calculate \cref{eq:max_mixture}, a lookup table is set up where the HSV color space is sampled with a grid of $20$ \texttimes{} $20$ \texttimes{} $20$ points.

The resulting images contain a point-wise likelihood for each of the detectable colors. For blob detection, we use the implementation by \citet{nister2008linear} of the maximally-stable extremal regions (MSER) algorithm which yields a number of initial hypotheses. In order to select the final detections, we regard 
\begin{compactitem}
\item the size: the number of pixels of the region,
\item the aspect ratio of its oriented bounding box,
\item the convexity: the ratio of the number of pixels over the area of their convex hull,
\item the color: the average likelihood in the region, and
\item the background: the average discrepancy between the likelihood of pixels inside the region and pixels sampled from a surrounding circle.
\end{compactitem}
A classifier can be trained on these quantities but for the scope of this venture, the selection criteria were manually and individually tuned, which allowed us to better follow the behavior of the detector and quickly adapt it in case of failure. 

Upon a significantly confident detection, the algorithm switches to a tracking mode where only a window around the last known object position is searched for only the identified color. This enables a much faster detection rate, in particular during the picking maneuver. In close proximity to the ground when the object is expected to be only partly visible in the image, the shape criteria are ignored when filtering the hypothesis.

It is further possible to use this algorithm to detect whether an object is attached to the gripper by filtering for very large detections in the specific color. 

\subsection{Drop Box Detection}

In contrast to the pickable objects, the visual appearance of the drop box was not specified by the challenge rules. Hence, we deployed a very general approach, only assuming that the box was rectangular and would provide some contrast to the surrounding ground. It is explicitly not assumed that the box would be uniformly colored.\footnote{As a matter of fact, it was uniformly colored.} Nevertheless, the dimensions of the box are parameterized.

As for the landing pattern and object detection, the camera image is transformed to a bird's-eye perspective to account for the MAV attitude. A Hough transform of the resulting gradient image yields line segments (see \cref{fig:image_proc_pipeline_c3} (e)) that are combined in a RANSAC-like procedure. In order to combine only promising pairs of line segments, a hash table with the line orientation as key is set up and only approximately perpendicular line segments are sampled. Testing the rectangularity, aspect ratio, and size of all RANSAC hypotheses provides the detection (see \cref{fig:image_proc_pipeline_c3} (f)).

\section{State Estimation}
We use two filters for state estimation running on the onboard computer. The first filter maintains a height offset between the measured height over ground and the barometer. The second filter estimates the position and velocity of (faster) moving objects, \ie the landing target and moving objects for picking.

\subsection{Landing Pattern and Object Tracking}
We modified our MAV state estimation filter from \citet{icuas2017chimneyspector} and estimate position and velocity of the target in an allocentric frame with a constant velocity assumption in the prediction step.
Our generic filter design does not make any model assumptions and all dimensions are treated independently. Thus, we can employ the same filter with different dimensionality for all use cases.

Furthermore, in contrast to our MAV state estimation filter, we only incorporate position measurements, letting the filter predict velocities without explicit correction.
In case of the landing pattern, we consider detections from both cameras as independent observations and the filter merges them to a coherent world view.
The pose of the MAV and the projection of the target perceptions into the allocentric frame are subject to the same localization error.
Thus, the allocentric estimate of the target is consistent with the egocentric control of the MAV.
Since we do not make any assumptions about the path of the landing target, e.g., moving in an eight pattern in case of the landing target, our method is applicable to arbitrary pattern motions and independent from exact absolute MAV localization.
Outputs of the filter are allocentric 2D position and velocity estimates, used to intercept the target objects.
Please note that for the very slow moving pickable objects used in the actual MBZIRC challenges (much slower than the allowed maximum of \SI{5}{\kilo\meter\per\hour}), estimating the object velocities was unnecessary such that we omitted their estimation in favor of system stability.

\subsection{Laser Height Correction}
Operating close to the ground during picking makes a good height over ground estimate obligatory. The Matrice 100 provides absolute GNSS altitude and a barometric height, relative to a starting height.
While the first is usually not very accurate, especially close to the ground, the second is prone to drift over time.
Hence, we employ a downward pointing laser distance sensor in order to correct the drift.
Laser measurements close to the ground are very noisy.
At greater heights they are assumed to not be reliable due to bright sunlight, but for intermediate heights these measurements yield an absolute, drift-free height above ground.
In contrast, the barometer is very reliable and locally consistent.
Thus, we maintain an offset between laser height and barometric measurements and use this offset to correct the barometer drift.
To acquire the correct heights, we first transform the laser measurements into an attitude-corrected frame.
If the resulting measurement is between \SIrange{0.1}{6}{\meter}, we use this value to correct the height offset.
The advantage of this approach is that even without laser measurements over longer periods of time, the MAV can safely navigate at higher altitudes, \eg to explore the arena or to deliver objects,
but the filter still converges quickly to the correct height over ground when picking.
Unfortunately, the employed laser sensor can report erroneous range measurements when distances are outside the valid range. We discuss this problem in \refsec{sec:evaluation}.

\subsection{Visual Height Correction}
Similar to the laser height correction on our Challenge 3 MAVs, we employ a visual height correction in Challenge 1.
When the landing pattern is sufficiently visible in one of the camera images---\ie the surrounding circle is at least partially visible to get a reliable distance estimate---we maintain an altitude offset based on the known height of the pattern above the ground and the estimated height of the MAV above the pattern.
This is necessary to facilitate precise landings in the last phase of the descent when the pattern is too close to the camera in order to estimate an accurate distance.

\section{Navigation and Control}
\label{sec:Navigation_and_Control}

For fast navigation and real-time onboard control during the challenges, we employ our time-optimal trajectory generation method described by~\citet{beul2017icuas}. We do not differentiate between the challenges since agile and precise flight characteristics were desirable in all cases.

\subsection{MAV Model}
We assume the MAV to follow rigid-body dynamics and simplify it as a point mass with jerk $j$ as system input. Following Newton's second law, the system is a triple integrator in each dimension $(x,y,z)$ with position $p$, velocity $v$, acceleration $a$, and jerk $j$:
\begin{align}
\label{eq:system}
\dot{p} = v, \qquad \dot{v} = a, \qquad \dot{a} = j.
\end{align}

Thus, the three-dimensional allocentric state of the MAV $X$ can be expressed by
\begin{equation}
\label{eq:statevector}
X = 
  \begin{pmatrix}
    p_x & p_y & p_z\\
    v_x & v_y & v_z\\
    a_x & a_y & a_z
  \end{pmatrix}.
\end{equation}

We assume jerk $j$ to be the direct control input to the linear system. Without loss of generality, we define the z-axis to be collinear to the gravity vector. Furthermore, we define the origin to be the middle of the arena and the xy-plane equal with the ground plane.
We do not model
\begin{compactitem}
  \item moment of inertia,
  \item drag,
  \item weight changes due to disk attachment,
  \item yaw dynamics, and
  \item coupling of the axes that occurs in non-hover conditions,
\end{compactitem}
but rely on fast replanning to account for model uncertainties and unmodeled effects instead. Since our model is parameterless, our approach generalizes to all multicopters and no cumbersome parameter tuning is required. After the challenge, we employed the method even on a hexacopter weighing \SI{11.2}{\kilo\gram} without any parameter changes to show the robustness of the approach.

\subsection{Time-optimal Control}
\begin{figure}[t]
  \centering
  \includegraphics[trim=00mm 00mm 00mm 00mm,clip,width=0.8\linewidth]{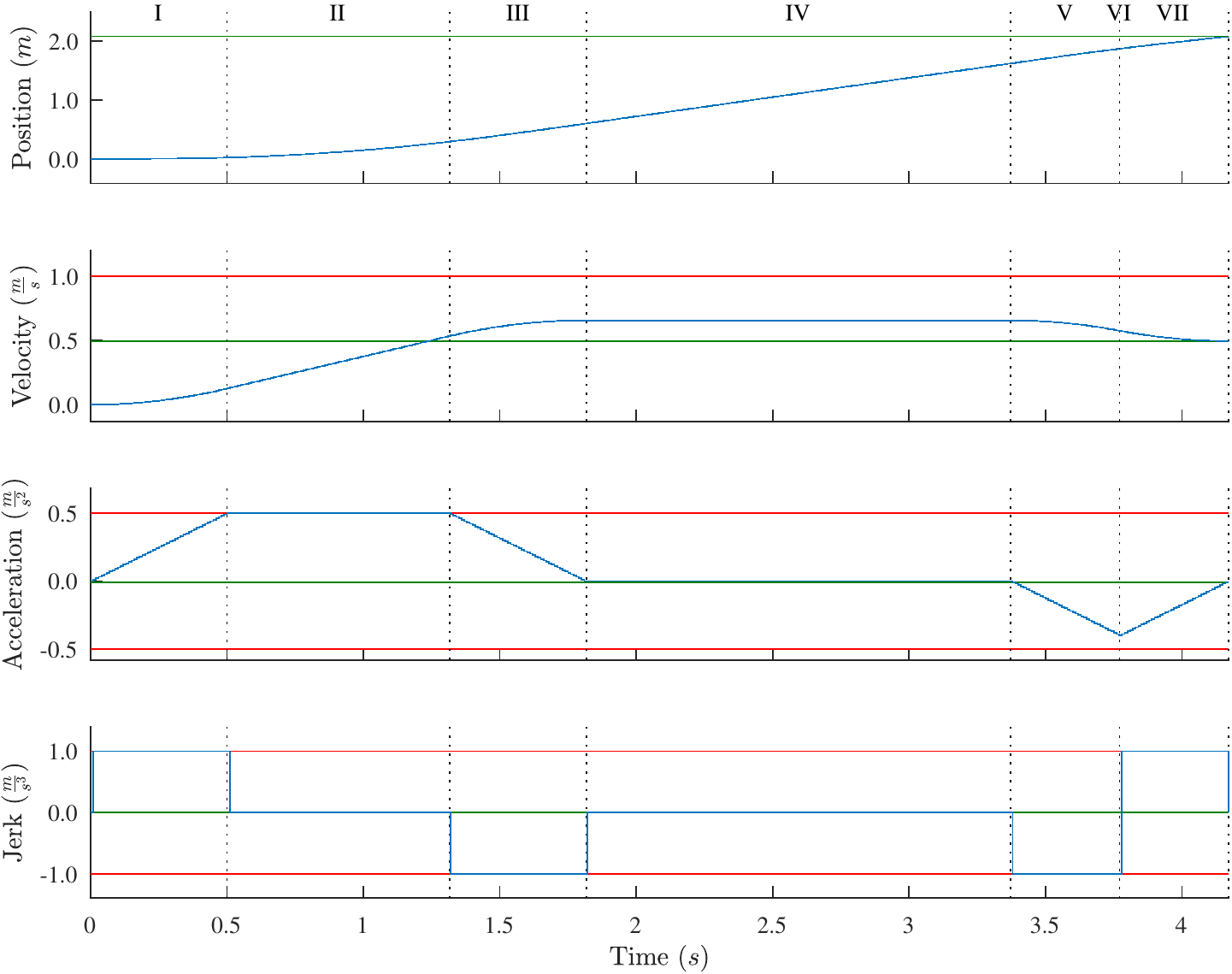}
  \caption{This time-optimal trajectory was generated with our method. Starting from state $(p_x = \SI{0}{\meter}, v_x = \SI{0}{\meter\per\second}, a_x = \SI{0}{\meter\per\second\squared})$, it brings the MAV to the target state $(p_x = \SI{2.08}{\meter}, v_x = \SI{0.5}{\meter\per\second}, a_x = \SI{0}{\meter\per\second\squared})$. The trajectory satisfies constraints $v_{max} = \SI{1}{\meter\per\second}$, $a_{max} = \SI{0.5}{\meter\per\second\squared}$, and $j_{max} = \SI{1}{\meter\per\second\cubed}$.
  The calculated switching times are $t_{1} = \SI{0.5}{\second}$, $t_{2} = \SI{0.82}{\second}$, $t_{3} = \SI{0.5}{\second}$, $t_{4} = \SI{1.55}{\second}$, $t_{5} = \SI{0.4}{\second}$, $t_{6} = \SI{0.0}{\second}$, and $t_{7} = \SI{0.4}{\second}$. The trajectory corresponds to the x-axis in \cref{fig:Simulation}. It is suboptimal (maximum velocity not reached) since this axis is slowed down to synchronize with the slower y-axis.}
  \label{fig:trajectory}
\end{figure}

Based on the simple triple integrator model, our method analytically generates third-order time-optimal trajectories that satisfy input ($j_{min} \leq j \leq j_{max}$) and state constraints ($a_{min} \leq a \leq a_{max}$, $v_{min} \leq v \leq v_{max}$). The planned trajectory consists of up to seven phases of constant jerk ($j = j_{max} \lor j = 0 \lor j = j_{min}$), resulting in a third-order bang-zero-bang trajectory.

\Cref{fig:trajectory} shows a 1-dimensional trajectory.
We synchronize all axes to arrive at the target at the same time. In doing so, the MAV flies on a relatively straight path towards the desired position.

For Challenge 1, we also use the ability of our trajectory generation method to calculate an optimal interception point, based on the current velocity of the target. We predict the target motion and do not fly to the current target location, but to the position, the MAV can intercept the target assuming a constant velocity motion and respecting the MAV constraints. Although the assumption of constant velocity may not be justified in the curved parts of the figure eight since the acceleration is relatively large ($\vert a_{target} \vert \approx \SI{1}{\meter\per\second\squared}$), we found the error to be compensable by fast replanning. We also intended to use this ability to pick up moving objects in Challenge 3, but since the objects were moving very slowly, this was unnecessary.

\subsection{MPC Application}
\label{sec:MPC_Application}
We use the above mentioned trajectory generation method as an MPC, running in a closed loop with $\SI{50}{\hertz}$. The Matrice 100 does not support directly sending jerk (or pitchrate) commands. Hence, we assume pitch and roll to directly relate to $\theta = \atantwo(a_x,g)$ and $\phi = \atantwo(a_y,g)$. We send smooth pitch $\theta$ and roll $\phi$ commands for horizontal movement and smooth climb rates $v_z$ in z direction instead. Due to the linearization, the acceleration constraint relates to an attitude constraint with $\theta_{max} = \atantwo(a_{max},g)$.

Our method plans the whole trajectory to the target instead of relying on a small constant lookahead, commonly found in MPCs. Since the whole future trajectory of the complete state (position, velocity and acceleration) is known, the acceleration setpoint for the underlying attitude control loop can be sampled at an arbitrary future time $\Delta t_{setp}$.
If the value is small (\eg $\Delta t_{setp} = \SI{0}{\second}$), the attitude control loop will react slowly, since the attitude setpoint differs little from the current attitude. If the lookahead value is too large, the overall system can become unstable or perform suboptimally. Also, communication delay has a negative impact on the system and is compensated by choosing an appropriate lookahead. We experimentally determined that the values found in \reftab{tab:Parameters_used_at_MBZIRC} offer good performance.

\begin{table}[t]
\small
\caption{Parameters used at MBZIRC}
\label{tab:Parameters_used_at_MBZIRC}
\begin{center}
\begin{tabularx}{0.71\columnwidth}{p{0.1\columnwidth}p{0.05\columnwidth}p{0.1\columnwidth}p{0.05\columnwidth}p{0.1\columnwidth}p{0.21\columnwidth}}
  \toprule
Parameter         & Axis & Value                                  & Axis  & Value                                  & Operation Mode \\
  \midrule
$v_{max}$         & x,y  & $\SI{8.33}{\meter\per\second}$         & z     & $\SI{1.00}{\meter\per\second}$         & normal \\
$v_{max}$         & x,y  & $\SI{6.00}{\meter\per\second}$         & z     & $\SI{1.00}{\meter\per\second}$         & during exploration \\
$v_{max}$         & x,y  & $\SI{8.33}{\meter\per\second}$         & z     & $\SI{0.50}{\meter\per\second}$         & during picking \\
\midrule
$a_{max}$         & x,y  & $\SI{4.73}{\meter\per\second\squared}$ & z     & $\SI{10.0}{\meter\per\second\squared}$ & always \\
$j_{max}$         & x,y  & $\SI{5.00}{\meter\per\second\cubed}$   & z     & $\SI{50.0}{\meter\per\second\cubed}$   & always \\
$\Delta t_{setp}$ & x,y  & $\SI{0.15}{\second}$                   & z     & $\SI{0.50}{\second}$                   & always \\
\bottomrule
\end{tabularx}
\end{center}
\end{table}

In the previous section, we report that we model the MAV in three orthogonal axes with the z-axis collinear to the gravity vector. The rotation of the MPC coordinate frame about the z-axis $\alpha$ is not defined, however. We define the rotation to be the allocentric angle of the current position to the target position $\alpha = \atantwo(p_{y_{wayp}} - p_y,p_{x_{wayp}} - p_x)$. In doing so, we project the per-axis velocity constraint to lie in the axis of the dominant motion. Otherwise, the global horizontal velocity constraint would result in being $v_{max} = \sqrt{v_{max_x}^2 + v_{max_y}^2}$ and thus violating the maximum allowed velocity of $\SI{30}{\kilo\meter\per\hour}$ at the MBZIRC.

\subsection{Yaw Control}
Although an arbitrary number of axes can be synchronized by the MPC, we do not consider the yaw-axis to be synchronized with the x, y and z-axis. For simplicity, we use proportional control for the yaw-axis $\Psi$. The proportional yaw rate setpoint $\dot\Psi_{setp} = K_{p} \cdot (\Psi_{setp} - \Psi)$ is sent to the MAV.

\section{Task Control State Machines}
\begin{figure}[t]
  \centering
  \includegraphics[width=0.9\linewidth]{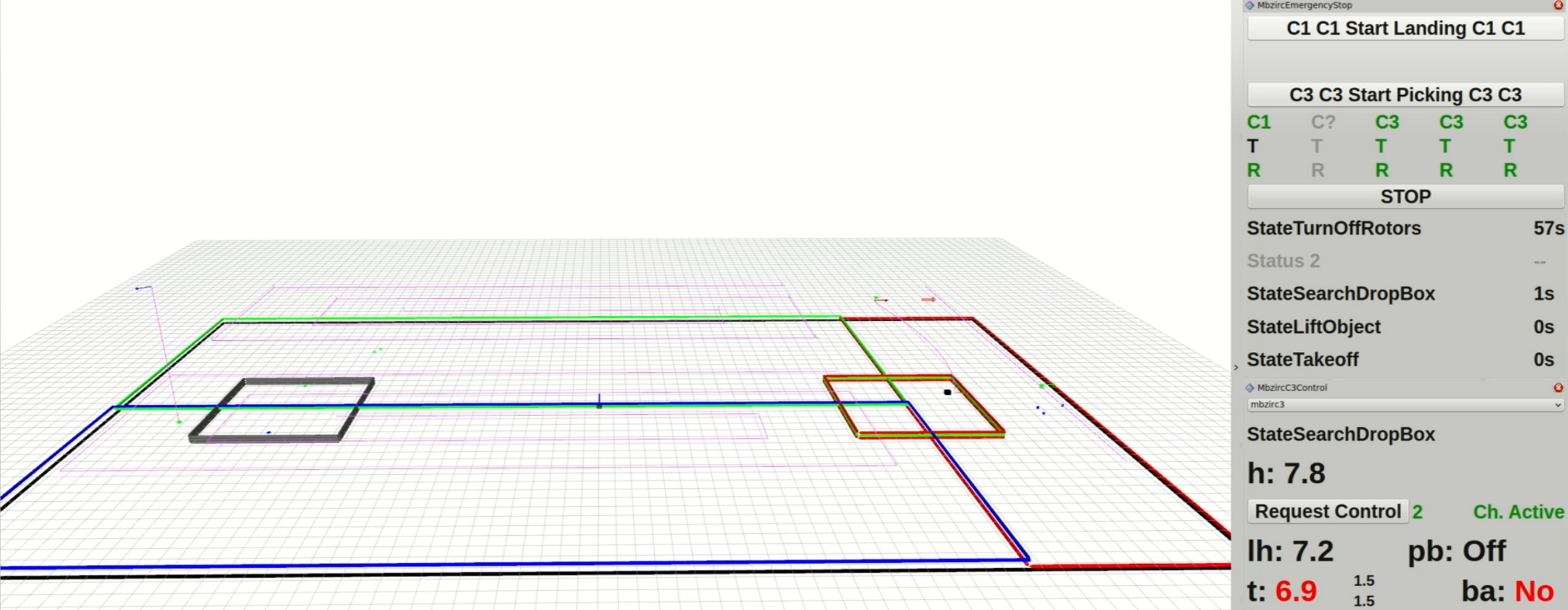}
  \caption{Operator GUI for the Grand Challenge. As no manual interaction with the MAVs is allowed outside of reset phases, we avoid interacting with any operator PCs during the challenge.
  Hence, we visualize as much of the MAVs state in an aggregated passive GUI to the operators. The main view shows MAV positions, navigation targets, perceptions, and assigned operation areas for the Treasure Hunt. The panels on the right show the configured challenge per MAV, communication status and current state machine states. Furthermore, some more detailed telemetry of a selected MAV can be visualized there.}
  \label{fig:gui-c3}
\end{figure}

The top-level coordination of all subcomponents of our MAVs is achieved by a state machine running at \SI{50}{\hertz}. 
The state machine serves as a generator for position, velocity, and yaw setpoints for the lower layers. Furthermore, it configures the perception and navigation modules and the hardware.
Since the state machine is the only subsystem which the operator interfaces with during flight, we built a distinct GUI for situational awareness of the operator.
\Cref{fig:gui-c3} shows the visualization for the Grand Challenge with four active MAVs.

\subsection{Landing}
\begin{figure}[t]
  \centering
  \resizebox{0.7\linewidth}{!}{
    \begin{tikzpicture}[font=\sffamily,>={Stealth[inset=0pt,length=4pt,angle'=45]}]
      \tikzset{terminal_node/.append style={minimum size=1.5em,minimum height=3em,minimum width={width("Search Point")+0.2em},draw,align=center,rounded corners,scale=0.65}}
      \tikzset{content_node/.append style={minimum size=1.5em,minimum height=3em,minimum width={width("Search Point")+0.2em},draw,align=center,scale=0.65,fill=blue!15!white}}
      \tikzset{label_node/.append style={scale=0.5}}
      \tikzset{group_node/.append style={align=center,rounded corners,inner sep=1em,thick}}
      \tikzset{decision_node/.append style={align=center,scale=0.5,shape aspect=1.5,minimum width=7.9em,minimum height=5.4em,diamond,draw,fill=yellow!25!white,font=\sffamily\normalsize}}

      \definecolor{red}{rgb}     {0.5,0.0,0.0}
      \definecolor{green}{rgb}   {0.0,0.5,0.0}
      \definecolor{blue}{rgb}    {0.0,0.0,0.5}
      \definecolor{grey}{rgb}    {0.5,0.5,0.5}

      \node(Takeoff)[terminal_node,fill=red!15!white] at (0.0,0.2) {Takeoff};
      \node(Fly_to_Search_Point)[content_node] at (1.7,0.2) {Fly to\\Search Point};
      \node(Rotate_at_Search_Point)[content_node] at (3.4,0.2) {Rotate at\\Search Point};
      \node(Pattern_Found)[decision_node] at (5.1,0.2) {Found\\Pattern?};
      \node(Rotate_to_Pattern)[content_node] at (6.8,0.2) {Rotate to\\Pattern};
      \node(Yaw_Ok)[decision_node] at (6.8,-1.0) {Angle\\$<$ \SI{20}{\degree}?};
      \node(Follow_Pattern)[content_node] at (5.1,-1.0) {Approach\\Pattern};
      \node(Switches2)[decision_node] at (3.4,-1.0) {Landed?};
      \node(Lost_Pattern)[decision_node] at (1.7,-1.0) {Lost\\Pattern?};
      \node(Landing_Ok)[decision_node] at (0.,-1.0) {Land?};
      \node(Landing)[content_node] at (0.,-2.2) {Landing};
      \node(Switches)[decision_node] at (1.7,-2.2) {Landed?};
      \node(Turn_Off_Rotors)[terminal_node,fill=green!15!white] at (3.4,-2.2) {Turn Off\\Rotors};

      \draw[->, thick] (Takeoff) -- (Fly_to_Search_Point);
      \draw[->, thick] (Fly_to_Search_Point) -- (Rotate_at_Search_Point);
      \draw[->, thick] (Rotate_at_Search_Point) -- (Pattern_Found);
      \draw[->, thick] (Pattern_Found) -- node[label_node,near start,above] {Yes} (Rotate_to_Pattern);
      \draw[->, thick] (Pattern_Found) -- node[label_node,near start,left] {No} (Pattern_Found |- 0.,0.9) --  (Rotate_at_Search_Point |- 0.,0.9) -- (Rotate_at_Search_Point);
      \draw[->, thick] (Rotate_to_Pattern) -- (Yaw_Ok);
      \draw[->, thick] (Yaw_Ok) -- node[label_node,near start,above] {Yes} (Follow_Pattern);
      \draw[->, thick] (Yaw_Ok) -- node[label_node,near start,above] {No} ([xshift=0.6em]Yaw_Ok-|Yaw_Ok.east) -- ([xshift=0.7em]Rotate_to_Pattern.east) -- (Rotate_to_Pattern);
      \draw[->, thick] (Follow_Pattern) -- (Switches2);
      \draw[->, thick] (Switches2) -- node[label_node,near start,above] {No} (Lost_Pattern);
      \draw[->, thick] (Switches2) -- node[label_node,near start,left] {Yes} (Turn_Off_Rotors);
      \draw[->, thick] (Lost_Pattern) -- node[label_node,near start,above] {No} (Landing_Ok);
      \draw[->, thick] (Lost_Pattern) -- node[label_node,near start,left] {Yes} (Lost_Pattern |- 0.,-0.4) -- (Fly_to_Search_Point |- 0.,-0.5) -- (Fly_to_Search_Point);
      \draw[->, thick] (Landing_Ok) -- node[label_node,near start,above] {No} (Landing_Ok -| -0.9,0) -- (-0.9,-2.7) -- (Follow_Pattern |- 0.,-2.7) -- (Follow_Pattern);
      \draw[->, thick] (Landing_Ok) -- node[label_node,near start,left] {Yes} (Landing);
      \draw[->, thick] (Landing) -- (Switches);
      \draw[->, thick] (Switches) -- node[label_node,near start,above] {Yes} (Turn_Off_Rotors);
      \draw[->, thick] (Switches) -- node[label_node,near start,left] {No} (Switches |- 0.,-1.55) -- ([xshift=1em]Landing |- 0.,-1.55) -- ([xshift=1em]Landing.north);
    \end{tikzpicture}
  }
  \caption{The flowchart of our landing state machine. In addition to the basic behavioral control, it features strategies to recover from failed landing approaches.}
  \label{fig:state_machine_c1}
\end{figure}
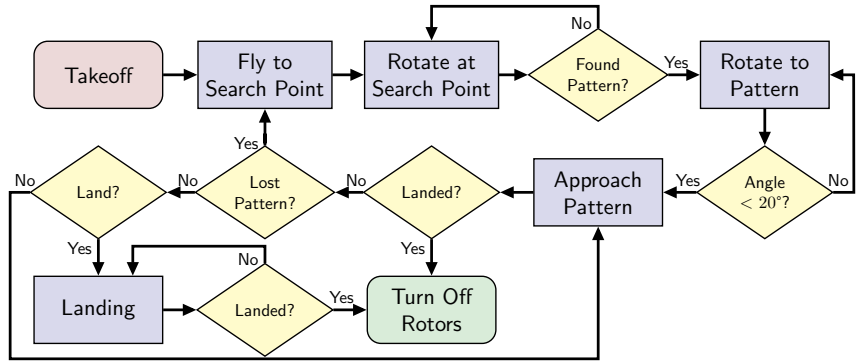

\Cref{fig:state_machine_c1} shows a flowchart of our state machine used in Challenge 1.
Multiple different yaw behaviors can be selected by the high-level state machine that are executed by the MPC, depending on the current situational requirement. The MAV can align its heading direction towards:
\begin{compactenum}
  \item a defined allocentric yaw angle,
  \item the current target,
  \item the optimal interception point,
  \item forward direction (current MAV horizontal velocity vector),
  \item direction of target motion (current target horizontal velocity vector), and
  \item the current yaw feedback, resulting in no yawing motion.\\
\end{compactenum}
After takeoff, the MAV flies with maximum velocity to a search point in the middle of the field with an altitude of \SI{8}{\meter}. Meanwhile it already explores the arena through yaw rotations using Behavior~1. After arriving at the search point, it becomes stationary while constantly yawing.
When the landing pattern is first detected, we rotate the front camera into pattern direction. When the yaw error is smaller than a \SI{20}{\degree} threshold, it switches to Behavior~2. and constantly yaws to follow the target direction while approaching the target.
We restrict the descent rate based on the distance to the target to ensure good perceivability of the pattern in the cameras.

When close to the target ($\leq \SI{5}{\meter}$), the yaw is fixed to the forward direction (Behavior~4) to avoid excessive yawing close to the singularity above the target. Since the MAV is now flying in target direction, this usually means no large yaw setpoint change in comparison to the previous behavior.

The final landing decision is based on relative orientation, distance and relative height to the pattern, and its visibility in the cameras. If the landing decision has been taken, the MAV descends until ground contact is detected by the switches at its feet. This is necessary as the pattern cannot be reliably tracked during the landing due to its proximity to the MAV. The yaw is held constant (Behavior~6) during this maneuver to not disturb the landing process.

To prevent unstable behavior while maneuvering in the vicinity of the fast moving landing pattern or in corner cases for the perception, the descent is completely aborted and the landing procedure restarts from the initial search point above the field when the pattern is lost during following.
For safety, we also detect premature landings in the pattern approaching state and turn off the rotors.

\subsection{Treasure Hunt}
\begin{figure}[t]
  \centering
  \resizebox{0.7\linewidth}{!}{
    \usetikzlibrary{fit,shapes.callouts,shapes.geometric,backgrounds,positioning,arrows.meta}
    \begin{tikzpicture}[auto,font=\sffamily,>={Stealth[inset=0pt,length=4pt,angle'=45]}]]
      \tikzset{terminal_node/.append style={minimum size=1.5em,minimum height=3em,minimum width={width("Search Point")+0.2em},draw,align=center,rounded corners,scale=0.65}}
      \tikzset{content_node/.append style={minimum size=1.5em,minimum height=3em,minimum width={width("Search Point")+0.2em},draw,align=center,scale=0.65,fill=blue!15!white}}
      \tikzset{label_node/.append style={scale=0.5}}
      \tikzset{group_node/.append style={align=center,rounded corners,inner sep=1em,thick}}
      \tikzset{decision_node/.append style={align=center,scale=0.5,minimum height=3em,shape aspect=1.5,diamond,draw,fill=yellow!25!white}}

      \definecolor{red}{rgb}     {0.5,0.0,0.0}
      \definecolor{green}{rgb}   {0.0,0.5,0.0}
      \definecolor{blue}{rgb}    {0.0,0.0,0.5}
      \definecolor{grey}{rgb}    {0.5,0.5,0.5}

      \draw[thick, rounded corners, green!10!white,fill] (0.95,0.95) -- (7.8,0.95) -- (7.8,-.45) -- (0.95,-.45) -- cycle;
      \draw[thick, rounded corners, grey!10!white,fill] (0.95,-0.55) -- (7.8,-0.55) -- (7.8,-2.85) -- (4.55,-2.85) -- (4.55,-1.65) -- ( .95,-1.65)-- cycle;
      \draw[thick, rounded corners, red!10!white,fill] (-0.9,-0.55) -- (-0.9,-4.5) -- (7.8,-4.5) -- (7.8,-2.95) -- (4.45,-2.95) -- (4.45,-1.75) -- ( .85,-1.75) --(0.85,-0.55) -- cycle;
      \node(explore_label)[label_node,scale=1] at (7.95,0.14){\rotatebox{-90}{\textbf{Exploration}}};
      \node(picking_label)[label_node,scale=1] at (7.95,-1.7){\rotatebox{-90}{\textbf{Picking}}};
      \node(dropping_label)[label_node,scale=1] at (7.95,-3.65){\rotatebox{-90}{\textbf{Delivery}}};

      \node(Takeoff)[terminal_node,fill=red!15!white] at (0.0,0.6) {Takeoff};
      \node(Delivery)[content_node] at (0.,-3.6) {Delivery};
      \node(GoToDropZone)[content_node] at (0.,-2.4) {Approach\\drop zone};
      \node(DropObject)[content_node] at (0.,-1.2) {Drop\\object};

      \node(FlyToNextObject)[content_node] at (1.8,-1.2) {Approach\\next object};
      \node(SearchDropBox)[content_node] at (1.8,-3.6) {Search\\drop box};
      \node(ObjectsAvailable)[decision_node] at (1.8,0.) {Objects\\$>$ 0};
      \node(FoundDecision)[decision_node] at (1.8,-2.4) {Found?};

      \node(Sink)[content_node] at (3.6,-1.2) {Sink};
      \node(DeliveryALittleBit)[content_node] at (3.6,-2.4) {Safe\\delivery};
      \node(TransferExplorationDecision)[decision_node,scale=1.0] at (3.4,0.0) {Transfer?};
      \node(DropZoneFreeDecision)[decision_node] at (3.6,-3.6) {Occupied?};

      \node(TransferToExploration)[content_node,scale=0.75,fill=blue!15!white,minimum height=1em] at (4.3,0.5) {Transfer};

      \node(LiftObject)[content_node] at (5.4,-2.4) {Lift\\object};
      \node(ExploreArena)[content_node] at (5.9,0.0) {Explore\\arena};
      \node(WaitInFrontOfDropZone)[content_node] at (5.4,-3.6) {Wait in front\\of drop zone};
      \node(SinkAbortDecision)[decision_node] at (5.4,-1.2) {Contact?};

      \node(TransferToDropZone)[content_node,scale=0.75,fill=blue!15!white,minimum height=1em] at (7.2,-4.2) {Transfer};
      \node(ObjectConfirmDecision)[decision_node,dotted] at (7.2,-2.4) {Confirm?};
      \node(TransferDropZoneDecision)[decision_node,scale=1.0] at (7.2,-3.4) {Transfer?};

      \draw[-, thick] (Takeoff) -- (Takeoff|-ObjectsAvailable);
      \draw[->, thick] (ObjectsAvailable) -- node[label_node,near start] {Yes} (FlyToNextObject);
      \draw[->, thick] (ObjectsAvailable) -- node[label_node,near start] {No} (TransferExplorationDecision);
      \draw[->, thick] (TransferExplorationDecision) -- node[label_node,near start] {Yes} ( TransferExplorationDecision |-TransferToExploration) -- (TransferToExploration);
      \draw[->, thick] (TransferExplorationDecision) -- node[label_node,near start] {No} (ExploreArena);
      \draw[-, thick] (TransferToExploration) -- (TransferToExploration -| 5.0,0.) -- (ExploreArena -| 5.0,0.);
      \draw[->, thick] (ExploreArena) -- (ExploreArena |- 0.,0.8);
      \draw[->, thick] (FlyToNextObject) -- (Sink);
      \draw[->, thick] (Sink) -- (SinkAbortDecision);
      \draw[->, thick] (SinkAbortDecision) -- node[label_node,near start] {Yes} (LiftObject);
      \draw[->, thick] (SinkAbortDecision) -- node[label_node,near start] {No} (SinkAbortDecision |- 0.,-0.6) -- (Sink |- 0.,-0.6) -- (Sink);
      \draw[->, thick] (SinkAbortDecision) -- node[label_node,near start] {Timeout} (SinkAbortDecision -| 7.2,0.) -- (7.2,0.8) -- (ObjectsAvailable |- 0.,0.8) -- (ObjectsAvailable);
      \draw[->, thick] (LiftObject) -- (ObjectConfirmDecision);
      \draw[->, thick] (ObjectConfirmDecision) -- node[label_node,near start] {Yes} (TransferDropZoneDecision);
      \draw[->, thick, dotted] (ObjectConfirmDecision) -- node[label_node,near start] {No} (SinkAbortDecision -| ObjectConfirmDecision);
      \draw[->, thick] (TransferDropZoneDecision) -- node[label_node,near start] {No} (TransferDropZoneDecision -| 6.4,0.) -- (WaitInFrontOfDropZone -| 6.4,0.) -- (WaitInFrontOfDropZone.east);
      \draw[->, thick] (TransferDropZoneDecision) -- node[label_node,near start] {Yes} (TransferToDropZone);
      \draw[->, thick] (TransferToDropZone) -- (TransferToDropZone -| 6.4,0.) -- (WaitInFrontOfDropZone -| 6.4,0.) -- (WaitInFrontOfDropZone);
      \draw[->, thick] (WaitInFrontOfDropZone) -- (DropZoneFreeDecision);
      \draw[->, thick] (DropZoneFreeDecision) -- (DropZoneFreeDecision -| 2.8,0.)-- node[label_node,near start] {No} (2.8,-4.4) -- (Delivery |- (0.,-4.4) -- (Delivery);
      \draw[->, thick] (DropZoneFreeDecision) -- node[label_node,near start] {Yes} (DropZoneFreeDecision |- 0.,-4.4) -- (WaitInFrontOfDropZone|- 0.,-4.4) -- (WaitInFrontOfDropZone);
      \draw[->, thick] (DropZoneFreeDecision) -- node[label_node,near start] {Timeout} (DeliveryALittleBit);
      \draw[->, thick] (DeliveryALittleBit) -- (DeliveryALittleBit |- 0.,-1.8) -- ([xshift=1em]0.,-1.8) -- ([xshift=1em]DropObject.south);
      \draw[->, thick] (DropObject) -- (DropObject |- ObjectsAvailable) -- (ObjectsAvailable);
      \draw[->, thick] (Delivery) -- (SearchDropBox);
      \draw[->, thick] (SearchDropBox) -- (FoundDecision);
      \draw[->, thick] (FoundDecision) -- node[label_node,near start] {Yes} (FoundDecision |- 0.,-1.8);
      \draw[->, thick] (FoundDecision) -- node[label_node,near start] {No} (FoundDecision -| 2.7,0.) -- (SearchDropBox -| 2.7,0. ) --(SearchDropBox);
      \draw[->, thick] (FoundDecision.west) --node[label_node,near start,xshift=-0.3em] {Timeout} (GoToDropZone.east);
      \draw[->, thick] (GoToDropZone) -- (DropObject);
    \end{tikzpicture}
  }
  \caption{Overview of our state machine for the Treasure Hunt task. To avoid any false negatives, the dotted part was shortcut during competition.}
  \label{fig:state_machine_c3}
\end{figure}
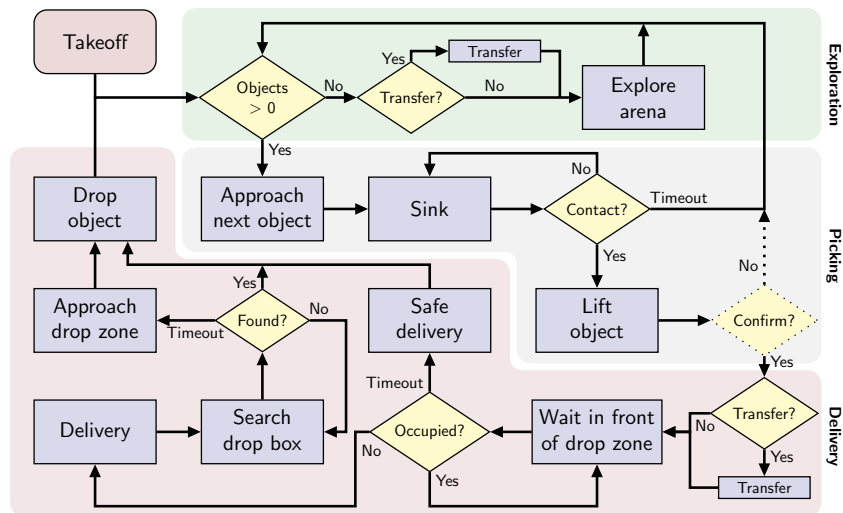
Whereas the competition arena is of rectangular shape without larger obstacles and with good GNSS coverage, picking small objects from the ground and the coordination of a team of multiple collaborating robots pose challenges for navigation and control.

The core of our control system for Challenge 3 is a state machine running at \SI{50}{\hertz}, depicted in \reffig{fig:state_machine_c3}.
The state machine selects navigation targets and configures the perception and navigation modules and the hardware.
After takeoff, and when the list of detected objects is empty, the system starts to explore the arena in a spiral pattern at a height of \SI{4}{\meter}.
The maximum horizontal exploration speed is \SI{6}{\meter\per\second}.
After the first exploration, we randomize the starting positions of the pattern in consecutive flights to avoid repetitive erroneous behaviors.

Immediately after object detection, we approach the closest object for picking.\footnote{Due to the high exploration velocity, it is possible to observe multiple objects before switching to the approach state.}
We transform object detections $p_{C_i}$ from camera frame $C_i$ of MAV $i$ into the common world frame $W$:
\begin{equation}
  p_W = T_{C_i}^W \cdot p_{C_i},
\end{equation}
where $T_{C_i}^W = T_{B_i}^W \cdot T_{C_i}^{B_i}$ consists of a transform from camera frame to ground-aligned MAV body frame $B_i$ and a transform from the MAV pose to the world frame.
The allocentric object detection $p_W$ is broadcasted to all MAVs and aggregated into the individual world models.
Thus, the egocentric object position relative to MAV $j$ is
\begin{equation}
  \underbrace{p_{B_j} = \vphantom{T_{C_j}^{B_j}}T_W^{B_j}}_{\textnormal{Navigation}} \cdot \underbrace{T_{B_i}^W \cdot T_{C_i}^{B_i} \cdot p_{C_i}}_{\textnormal{Perception}}.
\end{equation}
For locally perceived objects, \ie $i=j$, the localization transform $T_B^W$ multiplied with its inverse vanishes and thus the allocentric localization error is mitigated.

When reaching a position above the detected object, we confirm its color and reconfigure the object perception to use its fast tracking mode with only one color.
Using visual servoing, the MAV descends within a cone around the object center until either a) contact of the gripper with the object is detected, b) the measured distance to the ground from the laser falls below a safety height, or c) the object is not perceivable anymore.

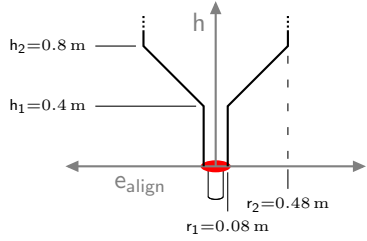
\begin{figure}[t]
  \centering
  \newlength{\scale}
  \setlength{\scale}{2cm}
  \begin{tikzpicture}[font=\sffamily,>={Stealth[inset=0pt,length=5pt,angle'=45]}]
    \tikzset{partial ellipse/.style args={#1:#2:#3}{insert path={+ (#1:#3) arc (#1:#2:#3)}}}

    \draw[-] (-0.05\scale,0\scale) -- (-0.05\scale,0.18\scale);
    \draw[-] (0.05\scale,0\scale) -- (0.05\scale,0.18\scale);
    \fill[red] (0,0.4) ellipse [x radius=0.1\scale, y radius=0.1*2/5*\scale];

    \draw[gray,thick,->] (0\scale,0.2\scale) -- node[very near end,left]{h} ( 0\scale,1.3\scale);
    \draw[gray,thick,<->] (-1\scale,0.2\scale) -- node[below,near start]{$\mathsf{e}_{\mathsf{align}}$} (1\scale,.2\scale);

    \draw[] (0,0) [partial ellipse=180:360:0.1cm and 0.04cm];
    \draw[] (0,0) [partial ellipse=180:360:0.1cm and 0.04cm];

    \draw[thick,-] (-0.08\scale,.2\scale) -- (-0.08\scale,0.6\scale) -- (-0.48\scale,1.\scale) -- (-0.48\scale,1.1\scale);
    \draw[thick,-] ( 0.08\scale,.2\scale) -- ( 0.08\scale,0.6\scale) -- ( 0.48\scale,1.\scale) -- ( 0.48\scale,1.1\scale);

    \draw[thick,densely dotted] (0.48\scale,1.2\scale) -- (0.48\scale,1.1\scale);
    \draw[thick,densely dotted] (-0.48\scale,1.2\scale) -- (-0.48\scale,1.1\scale);

    \draw[-](-0.53\scale,1.\scale) -- (-0.8\scale,1.\scale) node[left,font=\tiny]{$\mathsf{h_2}$=\SI{0.8}{\meter}};
    \draw[-](-0.13\scale,0.6\scale) -- (-0.8\scale,0.6\scale) node[left,font=\tiny]{$\mathsf{h_1}$=\SI{0.4}{\meter}};

    \draw[-](0.08\scale,0.16\scale) -- (0.08\scale,-.1\scale) node[below,font=\tiny]{$\mathsf{r_1}$=\SI{0.08}{\meter}};
    \draw[-](0.48\scale,0.18\scale) -- (0.48\scale,0.05\scale) node[below,font=\tiny]{$\mathsf{r_2}$=\SI{0.48}{\meter}};
    \draw[loosely dashed,-](0.48\scale,0.25\scale) -- (0.48\scale,0.95\scale);
  \end{tikzpicture}
  \caption{Descent for picking. The MAV aligns with the object to pick with visual servoing. We allow deviations up to \SI{1}{\meter} in larger heights. The allowed deviation is reduced to the object radius minus a safety margin between the heights $h_2$ and $h_1$. The MAV descends if the deviation is below the allowed alignment error $e_{align}$. If the error cannot be reduced for several seconds, the picking attempt is aborted.}
  \label{fig:descent_cone}
\end{figure}

For picking, we add the offset of the gripper $G_i$ to the ground-aligned body frame $T_{B_i}^{G_i}$ during descent to determine the navigation goal $p_{\textrm{nav}}$ in the allocentric frame $W$:
\begin{equation}
  p_{\textrm{nav}} = T_{G_i}^W \cdot T_{B_i}^{G_i} \cdot p_{B_i}.
\end{equation}
The transform $T_{B_i}^{G_i}$ is only approximately known as the gripper is passively aligned with the gravity vector without sensors, but the error is usually small for maneuvers with low accelerations.
The MAV descends if the xy-alignment of the gripper with the object with radius $r$ is good enough, \ie
\begin{align}
  e_\textrm{align} & < r_1 + ( r_2 - r_1 ) \max{\left(\frac{\min{\left(h_2,h\right)} - h_1}{h_2 - h_1}, 0\right)}\\
  & \overset{!}{=} 0.8 r + 0.4 \max{\left(\frac{\min{\left(0.8,h\right)} - 0.4}{0.8 - 0.4}, 0\right)},
\end{align}
with decreasing allowed alignment error $e_{\textrm{align}}$ when the height $h$ above the object decreases. \reffig{fig:descent_cone} illustrates the descent: The MAV aligns with the object up to a xy-distance of $r_2 = \SI{0.48}{\meter}$ before the descent starts. Between the heights $h_2$ = \SI{0.8}{\meter} and $h_1$ = \SI{0.4}{\meter} above the object, the allowed deviation from the object decreases conically until the gripper is aligned with the pickable disk minus a safety margin of \SI{20}{\percent} resulting in $r_1 = 0.8 r$. Below $h_1$, the descent is within a cylindric volume.
If the error cannot be reduced over time or the object is no longer visible, the picking attempt is aborted.
In case of abort, the MAV enters the exploration mode again. In the other cases, it ascends and starts visual confirmation whether or not an object is attached.

\begin{figure}
  \centering
  \includegraphics[angle=0,origin=c,width=0.32\linewidth]{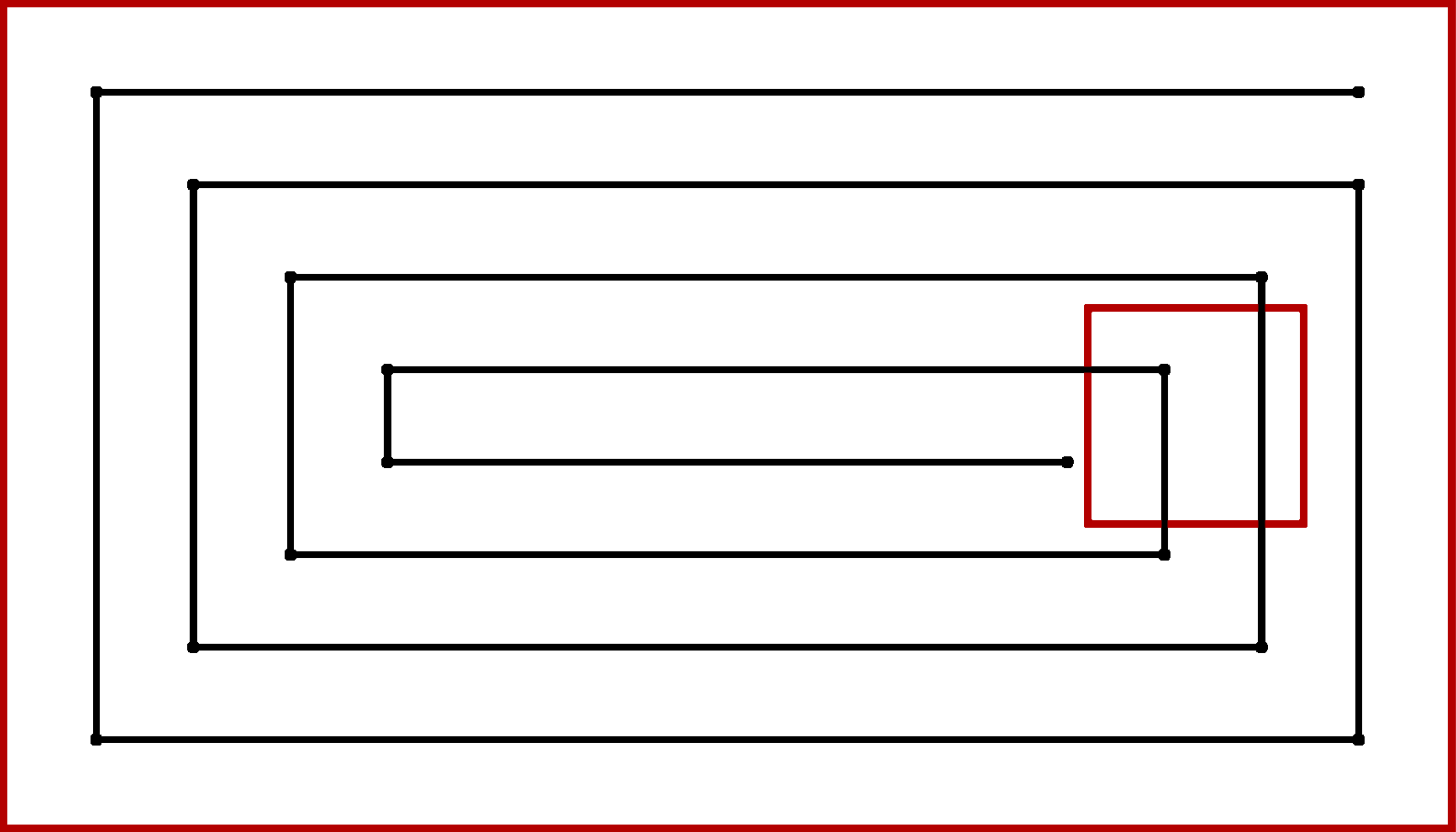}~
  \includegraphics[angle=0,origin=c,width=0.32\linewidth]{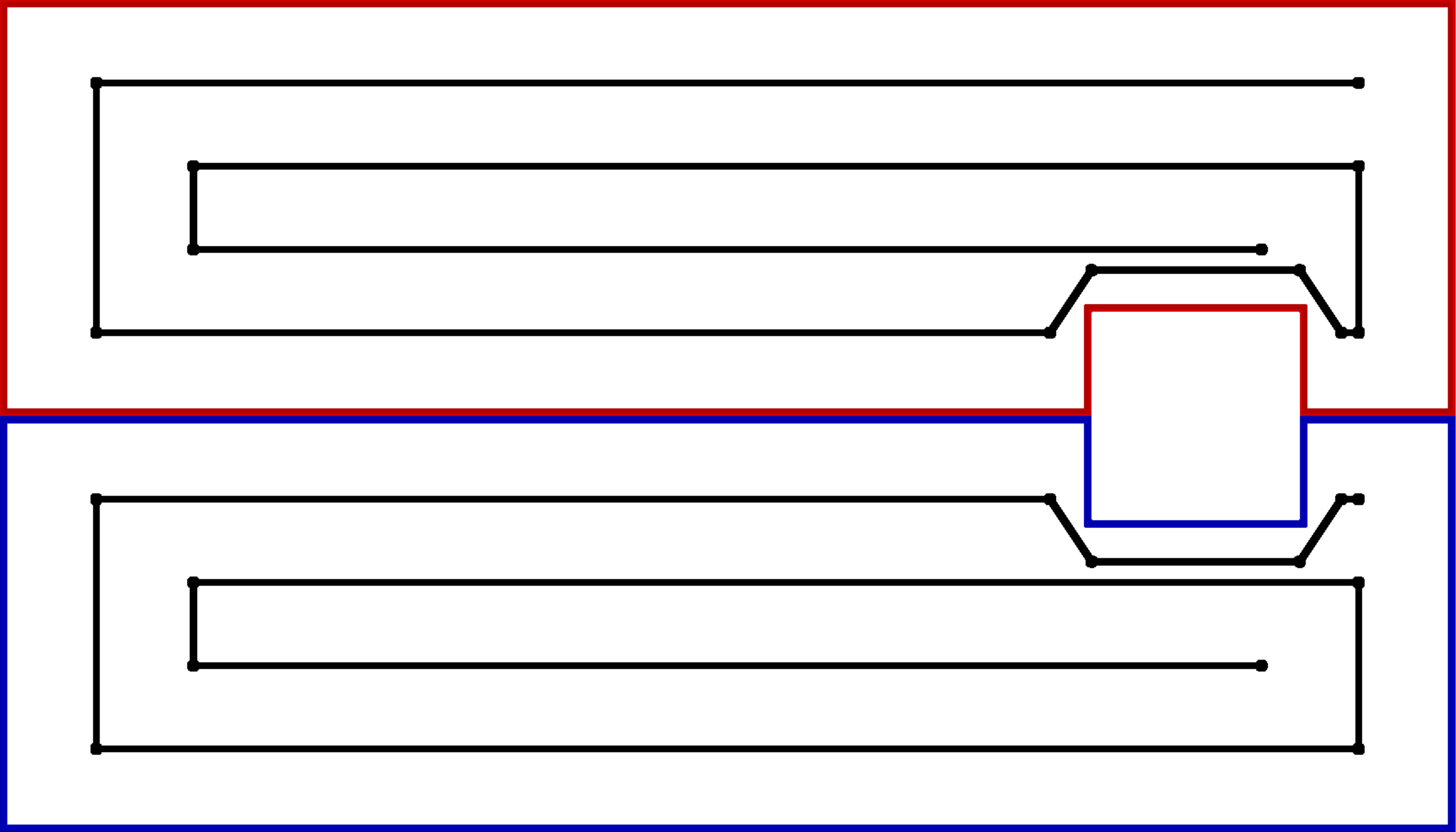}~
  \includegraphics[angle=0,origin=c,width=0.32\linewidth]{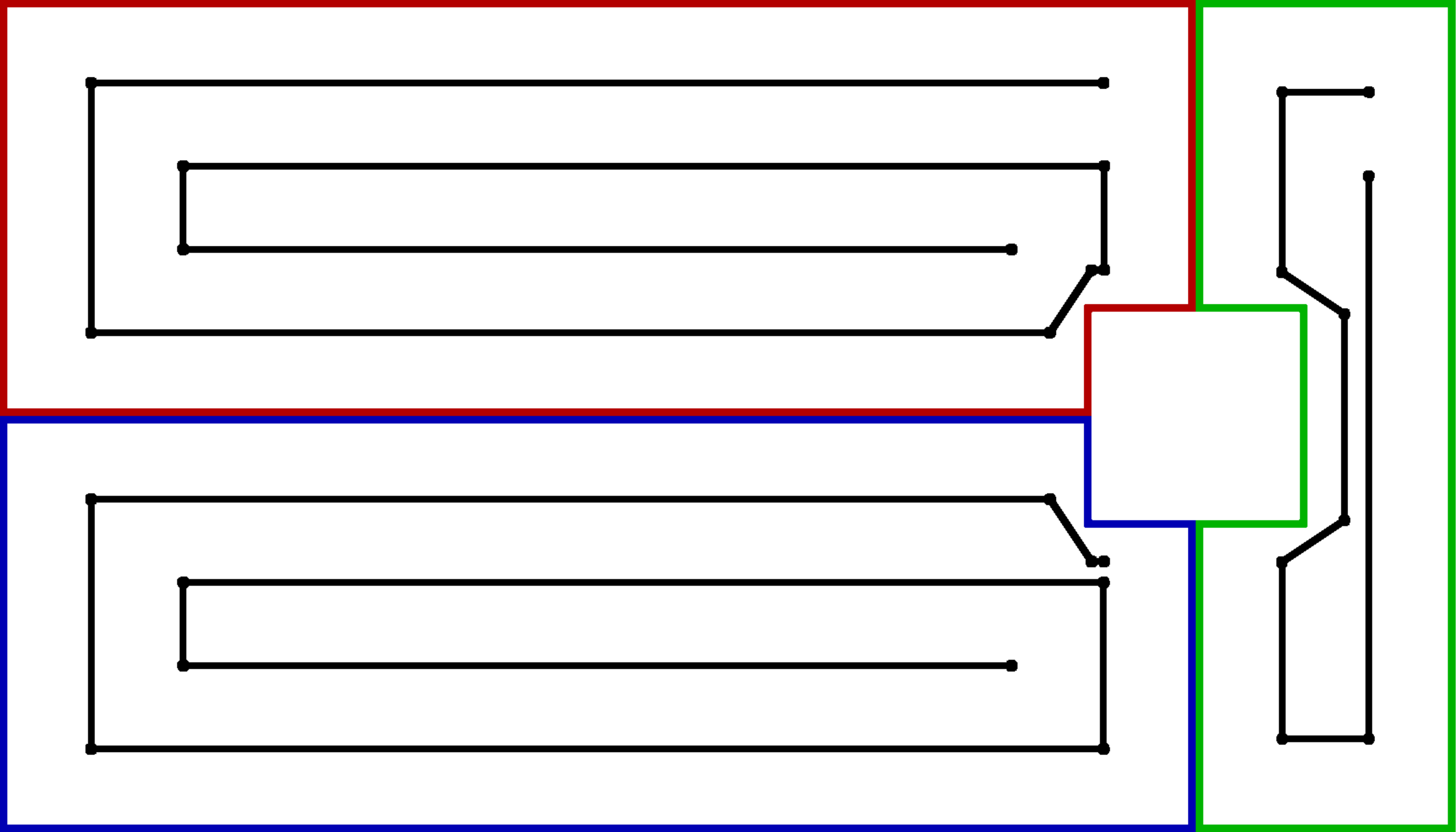}
  \caption{Sectors for safe operation of multiple MAVs. We divide the arena into one to three sectors (outlined in red, blue, green), depending on the number of active MAVs, all with access to the drop zone (small square). The black lines depict the exploration patterns. The drop zone is avoided if necessary.}
  \label{fig:exploration}
\end{figure}

To drop objects, the MAV enters the drop zone at a height of \SI{8}{\meter} and starts a local exploration flight to detect the drop box. If the drop box is detected, the MAV descends to \SI{1}{\meter} height and drops the object.
If the box is not detected in a certain time, the MAV drops the object at the predefined center of the drop zone.
As long as the drop zone is occupied, the MAV waits outside. It enters the drop zone close to its border after a timeout to drop the object safely for partial points.

The allocentric navigation is based on GNSS positions---bias corrected with help of the base station---in a field-centric coordinate system.
Positions in the arena, \eg endpoints of exploration trajectories, starting points of picking attempts, and the drop zone, are directly approached by means of our time-optimal trajectory generator \citep{beul2017icuas} with maximum velocity.
This approach is independent from the accurate weight and other parameters that change when picking or dropping objects and, thus, robust and at the same time efficient.

\subsection{Collaboration}
\label{sec:collaboration}
A particular challenge for flying robots is the payload constraint.
Increased weight significantly reduces the already relatively short flying time.
Thus, our MAVs are not equipped with sensors for 3D obstacle avoidance---in contrast to our previous work \citep{nieuwenhuisen2015jint}.
As a consequence, the MAVs have to avoid each other by coordination,
either explicitly by communication or implicitly by separation.
Wireless communication between robots is not reliable and has---potentially large---latency.
To ensure safe operation of multiple MAVs flying at high speeds of up to \SI{30}{\kilo\meter\per\hour},
we use a combination of separation and communication.
For basic operation, we divide the arena into sectors (see \reffig{fig:exploration}).
The number of sectors and their shape are derived from the number of active robots and their IDs.
Within their assigned sector, the MAVs are allowed to navigate freely below a maximum altitude without communication.
Outside their sectors, the MAVs transfer at assigned higher altitudes---with \SI{2}{\meter} vertical separation between MAVs---on straight lines.
For the MBZIRC, we opted for not equally-sized sectors in favor of straightforward accessibility of the drop box from each sector as it was not possible to assess the reliability of the communication infrastructure before the competition.
Horizontal sectoring has the advantage over vertical separation that the MAVs can explore the arena at an altitude optimal for visual perception of objects that is sufficiently low to reliably detect objects and maximizes the FoV of the camera.

\begin{figure}
  \centering
  \resizebox{0.8\linewidth}{!}{
  \begin{tikzpicture}[font=\sffamily]
    \usetikzlibrary{arrows,calc,positioning}
    \tikzset{zigzag/.style = {to path={ -- ($(\tikztostart)!.55!-5:(\tikztotarget)$) --($(\tikztostart)!.45!+5:(\tikztotarget)$) -- (\tikztotarget)\tikztonodes},sharp corners}}

    \node[inner sep=0pt] (kopter_a) at (-4,0) {\includegraphics[width=.15\textwidth]{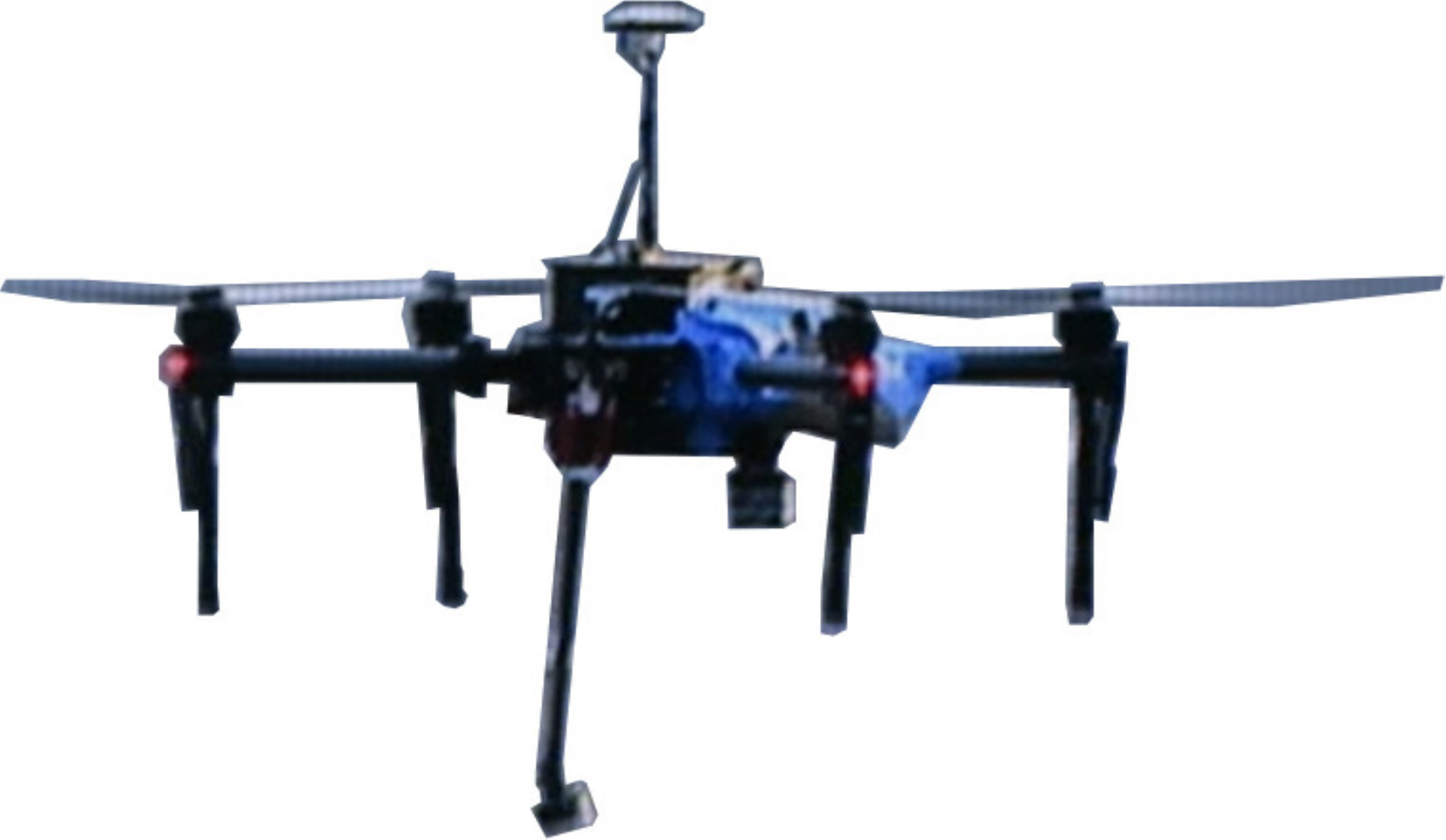}};
    \node[inner sep=0pt] (kopter_b) at (4,0) {\includegraphics[width=.15\textwidth]{ROB-17-0130_fig14a-eps-converted-to.pdf}};
    \node[inner sep=0pt] (kopter_c) at (0,-2) {\includegraphics[width=.15\textwidth]{ROB-17-0130_fig14a-eps-converted-to.pdf}};
    \node[inner sep=0pt] (kopter_l) at (7,-5) {\includegraphics[trim=0 15cm 0 0,clip,width=.45\textwidth]{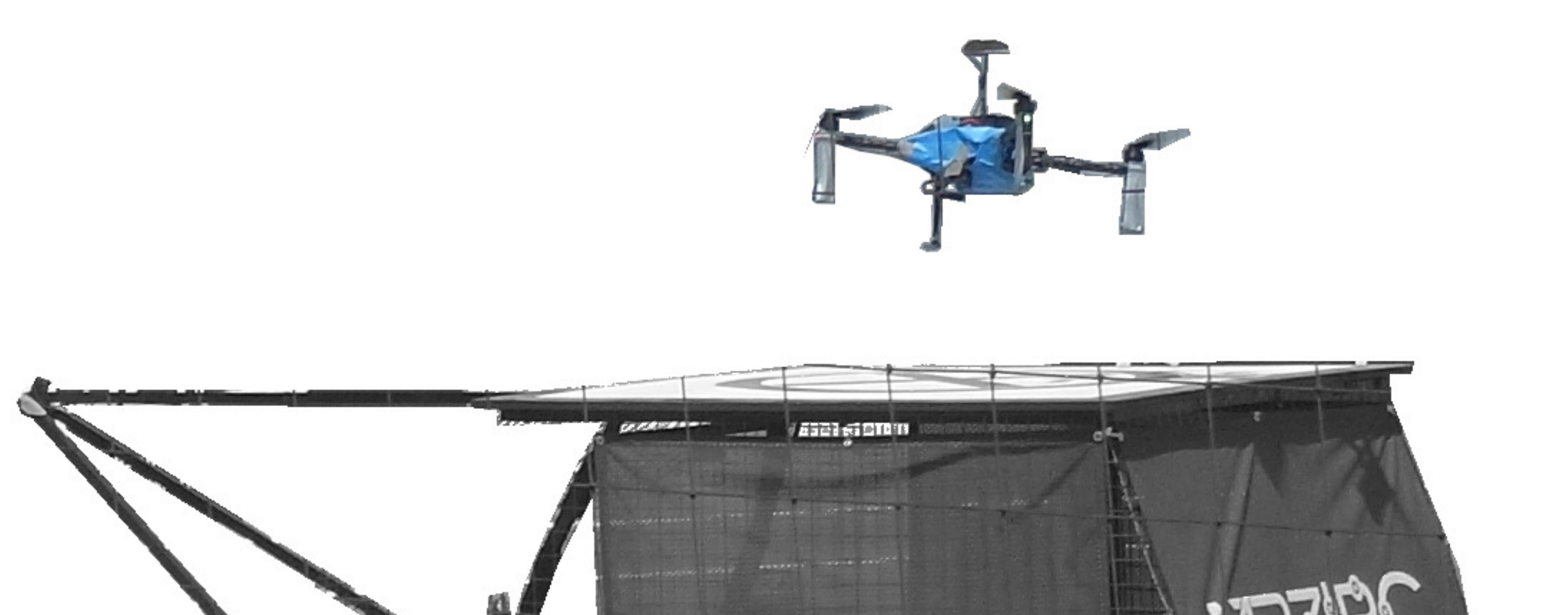}};
    \node[inner sep=0pt] (base_station_img) at (0.3,-7.5) {\includegraphics[width=.06\textwidth]{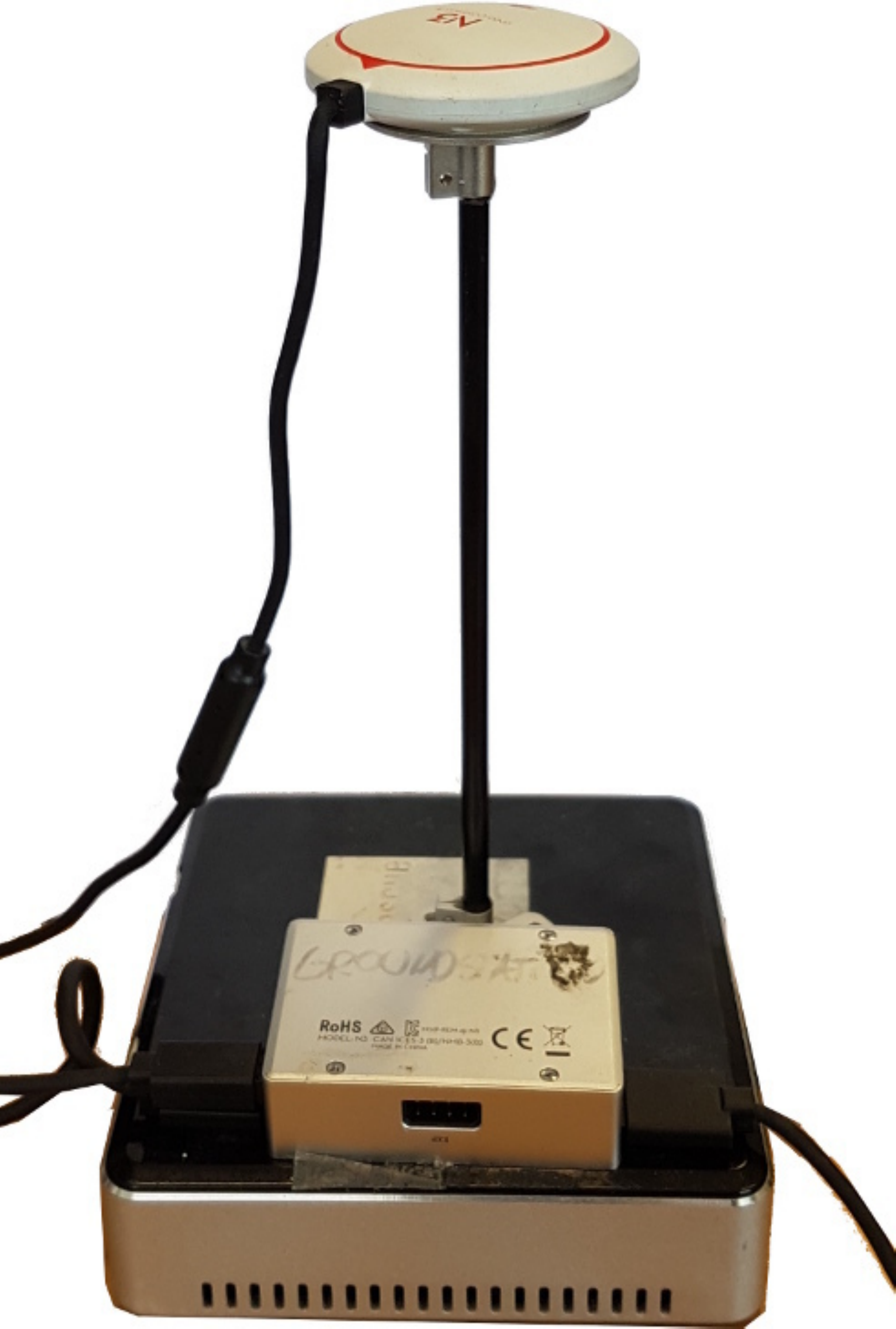}};
    \node[inner sep=0pt] (gcs_img) at (7,-7.2) {\includegraphics[width=.2\textwidth]{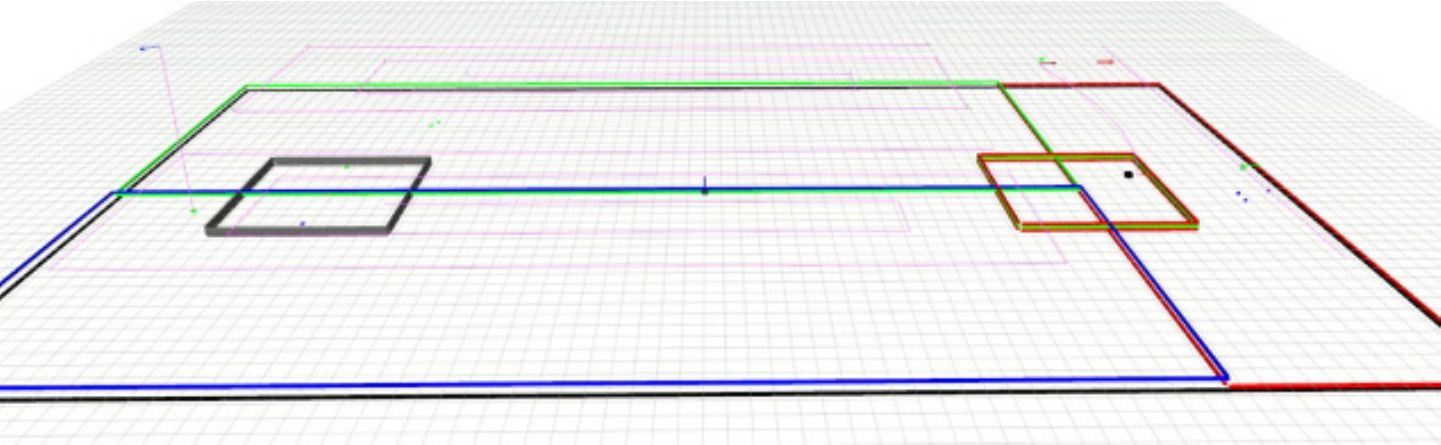}};
    \node[draw,rounded corners,thick] (base_station) at (2,-8) {Base Station};
    \node[draw,rounded corners,thick] (gcs) at (7,-8) {Ground Control Station};

    \node[] (label_a_b_up) at (0,1.4) {Team communication (10\,Hz):};
    \node[] (label_a_b_up) at (0,1) {Pose, flying state, target,};
    \node[] (label_a_b_down) at (0,0.6) {object detections};
    \node[rotate=-58] (label_gcs_up) at (-2,-4) {Uplink (1\,Hz): Corrected field center};
    \node[rotate=-58] (label_gcs_down) at (-2.5,-4) {Downlink (5\,Hz): World state};

    \draw[draw,thick,shorten >=1mm,-,dashed] (kopter_a) to [zigzag] (base_station);
    \draw[draw,thick,shorten >=1mm,-,dashed] (kopter_b) to [zigzag] (base_station);
    \draw[draw,thick,shorten >=1mm,-,dashed] (kopter_c) to [zigzag] (base_station);
    \draw[draw,thick,shorten >=1mm,-,dashed] (kopter_a) to [zigzag] (kopter_b);
    \draw[draw,thick,shorten >=1mm,-,dashed] (kopter_a) to [zigzag] (kopter_c);
    \draw[draw,thick,shorten >=1mm,-,dashed] (kopter_b) to [zigzag] (kopter_c);
    \draw[draw,thick,shorten >=1mm,-,dashed] (7,-4.5) to [zigzag] (base_station);
    \draw[draw,thick,-] (base_station) to (gcs);
  \end{tikzpicture}
  }
  \caption{Team communication. The MAVs receive the configuration of the field coordinate system and GPS corrections from a base station computer and transmit their world state to the base station. The state is visualized at the ground control station for the operator. In addition, the three picking MAVs broadcast a subset of their world state important for team coordination to each other.}
\label{fig:comm}
\end{figure}

We follow a two-fold system design principle aiming to use knowledge about the other agents when communication links are available and to still maintain operative when not.
Our system has no central control instance or explicit negotiation between agents.
The MAVs broadcast selected parts of their knowledge to all other agents, namely a) allocentric 3D position, b) current navigation target, c) detected objects outside of own operation sector, d) if the MAV is flying or landed.
The received information is integrated into the individual world models.
\Cref{fig:comm} shows our communication network topology.
For the task at MBZIRC, coordination is particularly important at the drop zone, \ie no two MAVs can safely drop objects at the same time into a \SI{1x1}{\meter} box, and when following dynamic objects that leave their own sectors while picking.
At the drop zone, we have defined decision positions close to it in the individual sectors.
The MAVs approach these points and decide if it is safe to proceed into the drop zone.
If the team communication is reliable, the agents can enter the drop zone immediately if unoccupied, based on the last position reports of the other MAVs.
If the communication to one or more of the teammates has timed out, we fall back to a time slot procedure.
For each time slot the MAVs decide whether the communication to the corresponding other MAV is reliable or not.
If it is reliable, the MAVs coordinate as in the case with full team communication, if not, the MAV can only enter the drop zone in its assigned slot.
If, by coincidence, two MAVs enter the drop zone at the same time, both have to leave the now mutually occupied zone, return to their decision positions and wait for a random amount of time.
This has been proven necessary after the first challenge trial as the high velocities in combination with communication latencies could lead to oscillating behavior.
We discuss such a case in detail in \refsec{sec:Lessons_Learned_c3}.

To avoid potential deadlock situations, in which \eg a landed MAV blocks the drop zone without reporting its state correctly,
the robots are allowed to enter the drop zone within a safety margin at its boundary after a timeout to drop their carried objects (State "Safe delivery" in \reffig{fig:state_machine_c3}).

MAVs are allowed to fly into neighboring sectors while picking dynamic objects, if the team communication is established.
To avoid collisions with MAVs exploring the sector at high velocity,
the picking MAV is only allowed to transit the exploration altitude when the other MAV is in a safe distance.
Due to the fast picking procedure---compared to the velocity of the dynamic objects at MBZIRC---this coordination procedure was never triggered in the actual competition.

The last received flying state of MAVs is employed to assess if the positions of the teammates are incorporated into the world model.
MAVs that are on the ground in the drop zone are not considered.
Also, MAVs that have been in the air and reported a landing are not considered even if the communication times out---they are assumed to have automatically landed due to low battery power and possibly to have completely shutdown afterwards.

The MAVs send their world model, team communication, state machine data, and planned trajectories to a ground control station at a reduced rate.
The received information is visualized in an aggregated view to a human observer, depicted in \reffig{fig:gui-c3}.

For all wireless communication, we employ a lightweight UDP protocol to encapsulate ROS messages and services without much protocol overhead.
The protocol is designed for high-latency connections, up to several seconds, and robust against connection losses or frequent reconnects.
The transmission of messages is performed on a best-effort base completely transparent to the overlying ROS infrastructure.

\section{Evaluation and Results}
\label{sec:evaluation}
We evaluate our system in simulation as well as with the real MAVs. Videos of our evaluation can be found on our website.\footnote{\url{www.ais.uni-bonn.de/videos/JFR_MBZIRC_2017}}

\subsection{Evaluation in Simulation}
To facilitate fast development and allow for safe testing of the system, we employed simulations on different levels of abstraction.
We simulated individual components in simplified scenarios and the whole system in physics-based simulation environments.

\begin{figure}[t]
  \centering
  \resizebox{1.0\linewidth}{!}{
    \begin{tikzpicture}[auto,>={Stealth[inset=0pt,length=5pt,angle'=45]}]]
      \definecolor{red}{rgb}{0.7,0.0,0.0}
      \node[anchor=south west,inner sep=0pt,scale=1.0] (matlab) at ( 0.0, 0.0) {\includegraphics[trim=00mm 00mm 00mm 00mm,clip,height=0.12\linewidth]{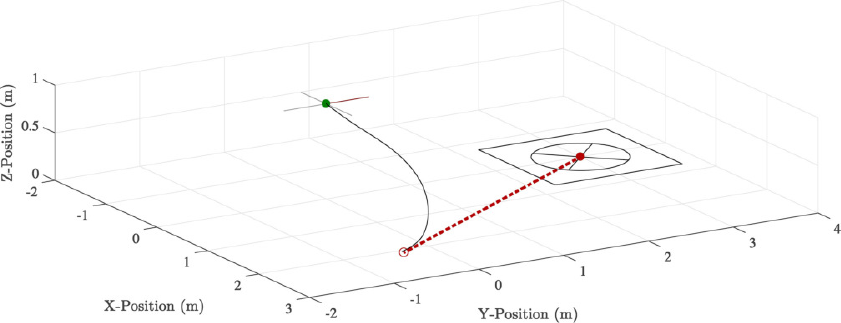}};
      \node[anchor=south west,inner sep=0pt,scale=1.0] (gazebo) at ( 5.67, 0.0) {\includegraphics[trim=00mm 00mm 00mm 00mm,clip,height=0.12\linewidth]{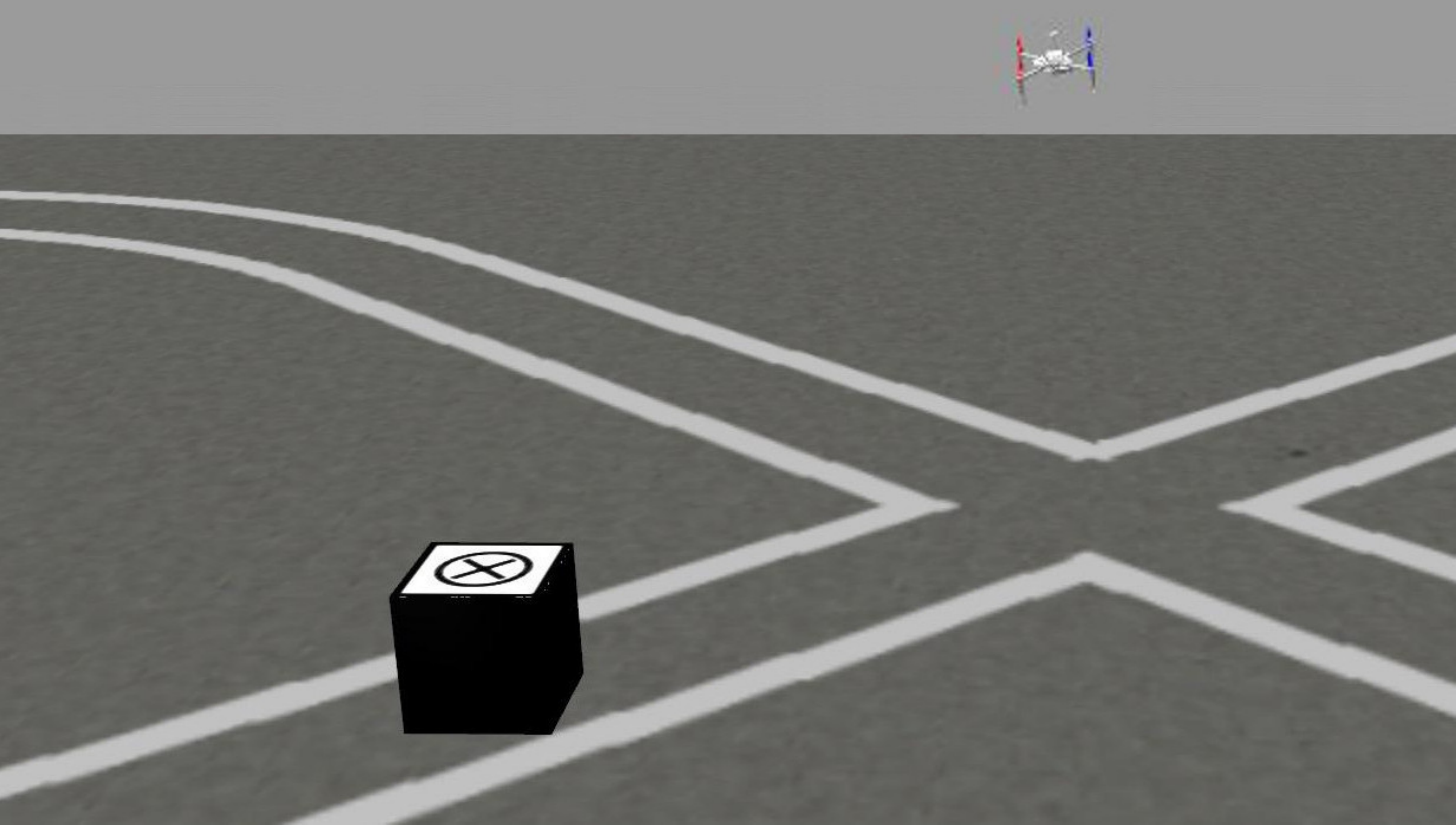}};
      \node[anchor=south west,inner sep=0pt,scale=1.0] (real)   at (9.6, 0.0) {\includegraphics[trim=00mm 00mm 00mm 00mm,clip,height=0.12\linewidth]{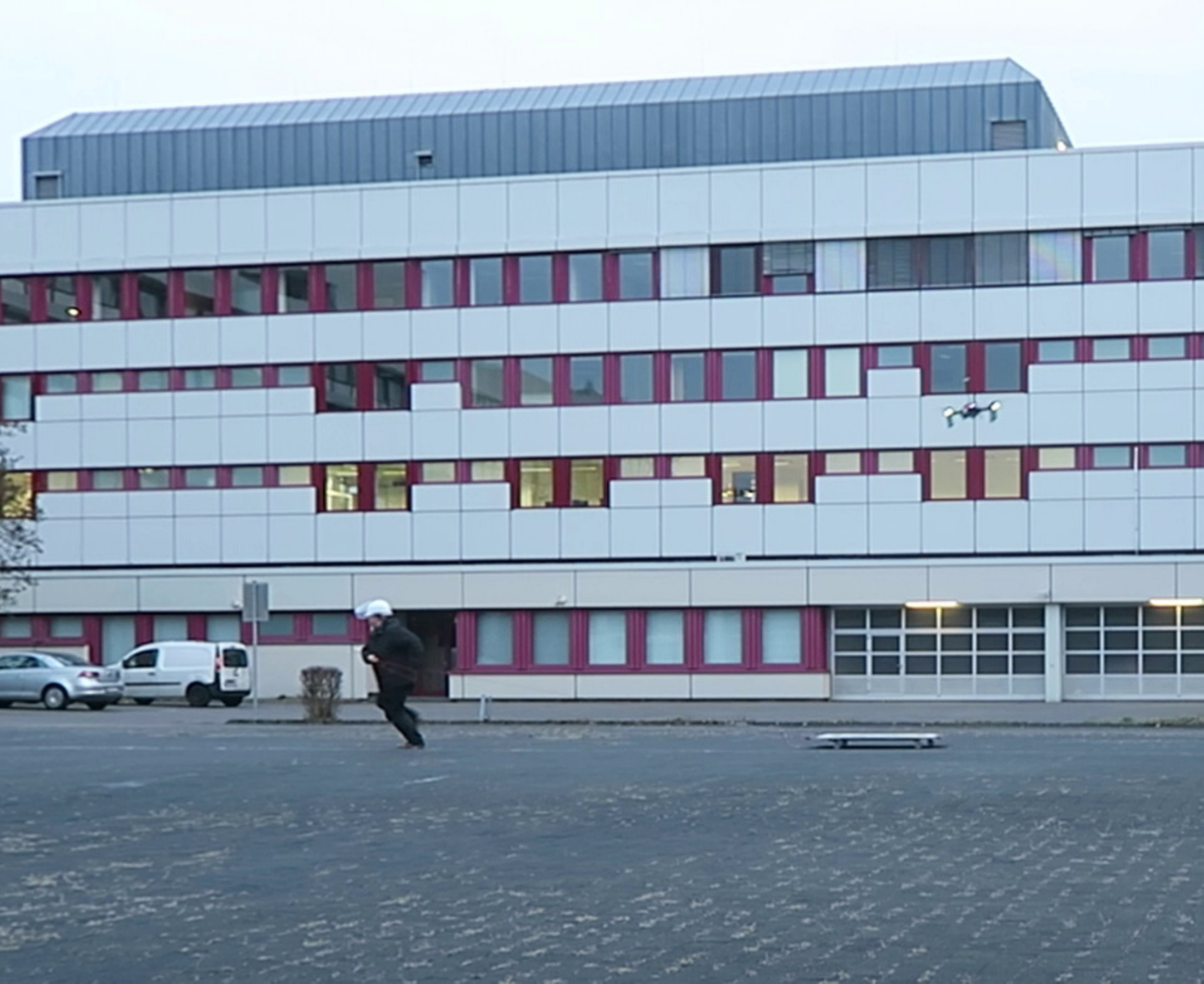}};
      
      \draw[line width=0.3mm, red, opacity=1.0] (11.55,1.15) circle (6pt);

      \draw[->, thick] (matlab) -- (gazebo);
      \draw[->, thick] (gazebo) -- (real);
    \end{tikzpicture}
  }
  \caption{Landing simulation in Matlab, Gazebo, and with a real robot. We first simulate the interception of the target with a simplified linear model. The MAV is marked with a green dot. The target is marked with a solid red dot. The predicted target trajectory is marked with a dashed line, ending in the interception point (red ring). Subsequently, we modeled the MBZIRC arena including the moving target in Gazebo. The MAV can be simulated with hardware in the loop (HIL), employing a complex motion model and challenging environmental conditions. After verifying the behavior in nonlinear simulation, real-robot experiments are conducted.}
  \label{fig:Simulation}
\end{figure}
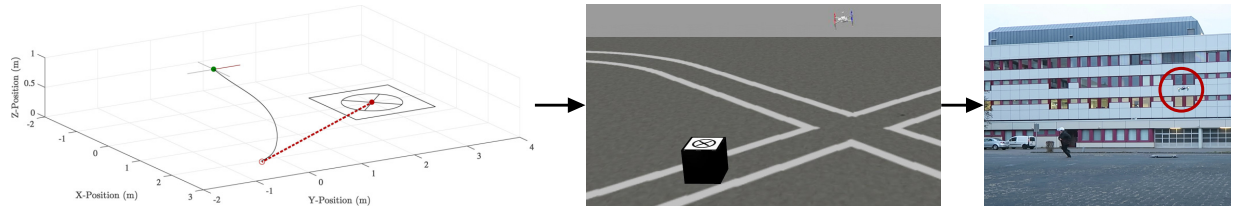

Our time-optimal controller was first simulated in Matlab. This component is in particular crucial for safe flights, \eg the risk of crashing the MAVs is inevitable if the controller works erroneous and, in particular, landing on a target moving at high speed imposes a risk for the MAV and people who move the target.
\Cref{fig:Simulation} shows our Matlab simulation for the optimal interception of a moving landing pattern. We model the MAV as described in \cref{sec:Navigation_and_Control}.
It can be seen that the MAV builds up momentum to eliminate any velocity difference to the target when arriving at the interception point.

To achieve a high level of realism, we also modeled the MBZIRC arena with the moving target for Challenge 1 and the objects and drop box for Challenge 3 in the RotorS simulator \citep{rotors2016}.
We simulate the interfaces of the DJI ROS-SDK---including GPS localization and the MAV cameras---in RotorS.
This simulation is primarily used to develop high-level control components and for integration testing between perception and actuation components.

In addition to the physics-based MAV simulation of RotorS, we implemented a hardware in the loop (HIL) bridge, employing the DJI simulator.
Here, the flight control on the MAV is connected to the DJI Assistant~2 software via USB.
Instead of controlling the motors of the real MAV, the flight control firmware sends control commands to the simulator, at which MAV dynamics and sensors like IMU and GPS are simulated and sent back to the flight control.
Thus, for our ROS middleware it is indeterminable if the real or the simulated MAV is used.
In contrast to the purely software-based simulation, this approach allows for testing with the real responses of the flight control given our inputs but requires access to the MAV hardware, which is a limiting factor for parallel development.
Consequently, this simulation was mainly used for the integration of low-level components and, in the final testing phase, for integration testing of the whole system.

For Challenge 3, we mainly used the RotorS simulation to test the complex interactions between perception, control, and the state machine.
To close the loop, we simulate simplified versions of the Lidar Lite v3 and the gripper.
Even though the visual perception algorithms are mainly developed and tested on recorded data from real flights, the closed-loop simulation revealed many corner cases, not covered by the experiments and helped to address these.
This, and identifying potential integration conflicts between components before testing campaigns involving many team members, helped to save valuable flight and experimentation time with the real MAVs and significantly sped up the development process.

\begin{figure}
  \centering
  \includegraphics[trim=00mm 00mm 00mm 00mm,clip,width=1.0\linewidth]{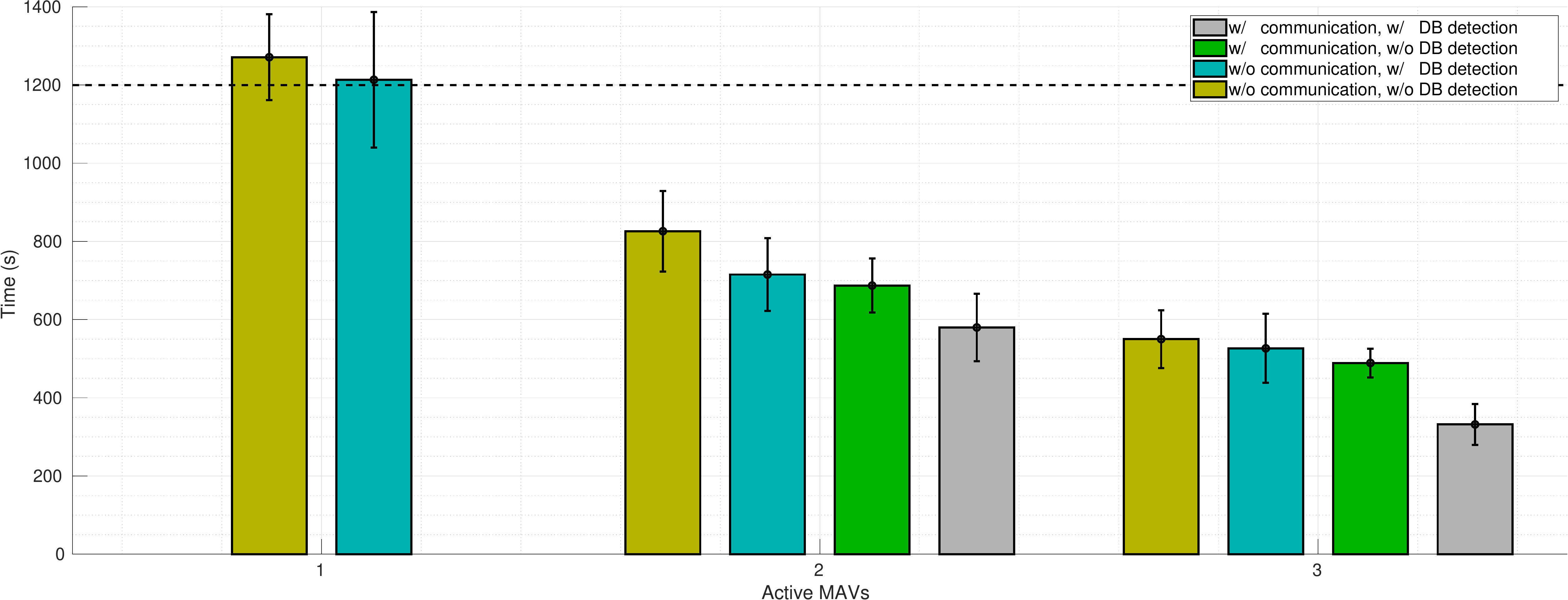}
  \caption{Time to pick all objects in simulation. We evaluate the influence of the number of MAVs in Challenge 3 on the challenge completion time. Gray bars depict the average time to pick and deliver all objects with working team communication and quick drop box detection, green bars with failing drop box detection---the MAVs search for the box until a timeout is reached. Turquoise bars bars depict the times without team communication, \ie the MAVs wait for time slots and ocher bars a combination of failed drop box detection and missing team communication. The dashed line indicates the total Challenge 3 duration. The completion time scales nearly linear with the number of MAVs up to the allowed number of three. It can be seen that team communication can reduce the average completion time substantially, especially in the case with reliable drop box detection. We averaged the times over six runs (five runs for three MAVs, see text) in randomized arena setups.}
  \label{fig:picking_times}
\end{figure}

We evaluate the influence of the number of employed MAVs on the challenge completion time in simulation.
For this evaluation, we used mainly the code base from the Grand Challenge with slight modifications to address erroneous behavior identified during and after the competition without altering the general strategies.
\reffig{fig:picking_times} shows the results based on six different arena setups with 13 static objects---the three yellow dynamic objects were simulated at fixed positions---randomly distributed.
We simulate a worst-case scenario where the drop box cannot be reliably perceived until timeout and a scenario where the drop box is perceived reliably.
In the worst-cast scenario, the delivery process---including entering the drop zone, searching for the drop box for \SI{15}{\second}, dropping the object, and leaving the drop zone again---takes approximately \SI{30}{\second}.
The minimum challenge completion time is, thus, 6:30 minutes if only one MAV is allowed within the drop zone at any point in time.

A single MAV can pick and deliver all objects in 21:12 minutes, which is slightly above the challenge duration of 20 minutes.
This is partially caused by a suboptimal exploration behavior after picking an object combined with the large area to explore.
Furthermore, the drop zone as a shared resource---and consequently one of the major restricting factors to reducing the time---is only occupied for less than \SI{31}{\percent} of the challenge time.

With three MAVS, the delivery process dominates the challenge completion time (8:12 minutes) as most of the time at least one MAV is waiting outside of the occupied drop zone, ready to enter.
The difference of 1:42 minutes to the optimum is mostly caused by the time required to search and deliver the first object and by the uneven distribution of the workload to the MAVs in the end of the challenge due to sectoring the arena.
Furthermore, the difference with and without team communication is mitigated by the fact that the alloted time slots of \SI{30}{\second} are used completely by the delivering MAV and most of the time an MAV is already waiting.
The overhead with and without team communication compared to a single MAV is \SI{5.14}{\percent} and \SI{9.94}{\percent}, respectively.
Please note, that one arena setup was excluded from the time calculation as only 12 out of the 13 objects could be picked due to too restrictive safety margins between two exploration sectors.
This would have had an impact on the final score in the actual challenge.

With two MAVs, the challenge completion time is nearly halved compared to one MAV, but the overhead without team communication is larger than with three MAVs (\SI{14.99}{\percent}).
One possible explanation is that the drop zone is less often occupied than with three MAVs, resulting in idle time slots where an MAV has to wait to enter the unoccupied drop zone.

In the scenario with reliable drop box detection, the duration of the delivery process is reduced by 1/3, resulting in an optimal challenge completion time of 4:17 minutes.
The average completion time with a single MAV was 20:13 minutes, a reduction of \SI{59}{\second}.
As expected, also with two and three MAVs the average challenge completion is reduced.
With two MAVs the reduction is 1:51 minutes without and 1:47 minutes with team communication.
A possible explanation for the large reduction in the case without team communication is that due to the shorter delivery process MAVs can arrive earlier in a consecutive delivery step and use an earlier time slot.
With three MAVs, the average completion time is reduced by \SI{24}{\second} without and 2:47 minutes with team communication.
As most of the time at least one and often two MAVs are already waiting at the drop zone while the third MAV delivers an object, as stated above, the drop zone is occupied most of the time with a \SI{30}{\second} delivery process and, thus, team communication cannot improve the drop zone usage much.
In the case with the quicker delivery process, the drop zone is unoccupied for at least 1/3 of every time slot.
Consequently, communication can greatly improve the usage of this shared resource.

For every number of MAVs, one of the six arena setups leads to the longest completion times regardless of available team communication and delivery duration.
A qualitative analysis of the arena setups indicates that in these cases the object distribution lead to a very different workload for the individual MAVs.

\subsection{Real-robot Evaluation at MBZIRC}
During the MBZIRC, our team came in third in Challenge 1 (Landing) and in Challenge 3 (Treasure Hunt) out of the total of 24, respective 18, competitors.
Together with our ground robot operating a valve, we won the Grand Challenge---with 14 competitors in total---and reached the second highest scores in the subtasks Landing and Treasure Hunt.
We decided to start the Landing task simultaneously with the ground vehicle subtask of turning a valve since for both time was of the essence. We began our Treasure Hunt only after both were finished to reduce the overall challenge complexity.

\subsubsection{Landing}
During the first run in Challenge 1, we first experienced a hardware problem with the USB 3.0 connection of the front camera and were forced to restart.
After fixing this issue, our MAV successfully landed in \SI{32}{\second}---measured from spinning up the rotors to landing on the pattern.
In total, the time from the start of the challenge to landing---including fixing the MAV---was \SI{112}{\second}, resulting in the third place in the final ranking.
When evaluating the flight performance from a control point-of-view by only measuring the time from the first detection of the target to a successful landing and neglecting the idle time, \eg due to waiting for the target, our landing was the fastest in the whole competition (\SI{7.8}{\second}).
Other teams like, \eg~\cite{Cantelli2017} first chased the target in a safe height and then slowly descended onto the target while trying to stay centered above the target. This behavior wasted time while following the target---sometimes multiple laps of the figure eight.
Our control strategy, however, was to directly dive onto the target on a glide path instead of hovering above it. Although this seems more risky, we were very satisfied with the performance and the reliability of the approach.

\begin{figure*}[t]
  \centering
  \begin{tikzpicture}
    \definecolor{yellow}{rgb}{0.7,0.7,0.0}
    \definecolor{pink}{rgb}{0.7,0.0,0.7}
    \definecolor{red}{rgb}{0.7,0.0,0.0}
    \definecolor{green}{rgb}{0.0,0.7,0.0}
    \definecolor{blue}{rgb}{0.0,0.0,0.7}
    \node[anchor=south west,inner sep=0] (image1) at (0,0) {\includegraphics[trim=00mm 00mm 00mm 00mm,clip,width=1.0\linewidth]{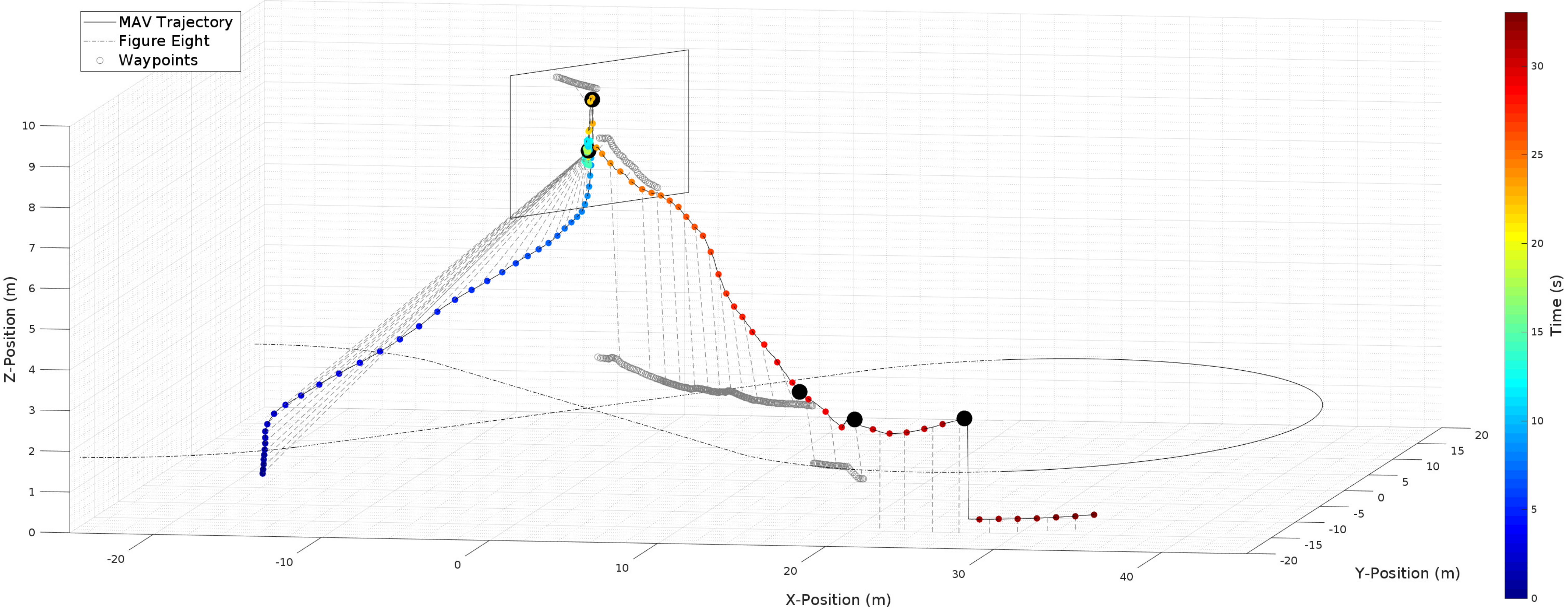}};
    \node[anchor=south west,inner sep=0.1,draw=black] (image2) at (9,2.2) {\includegraphics[trim=40mm 00mm 180mm 30mm,clip,width=0.35\linewidth]{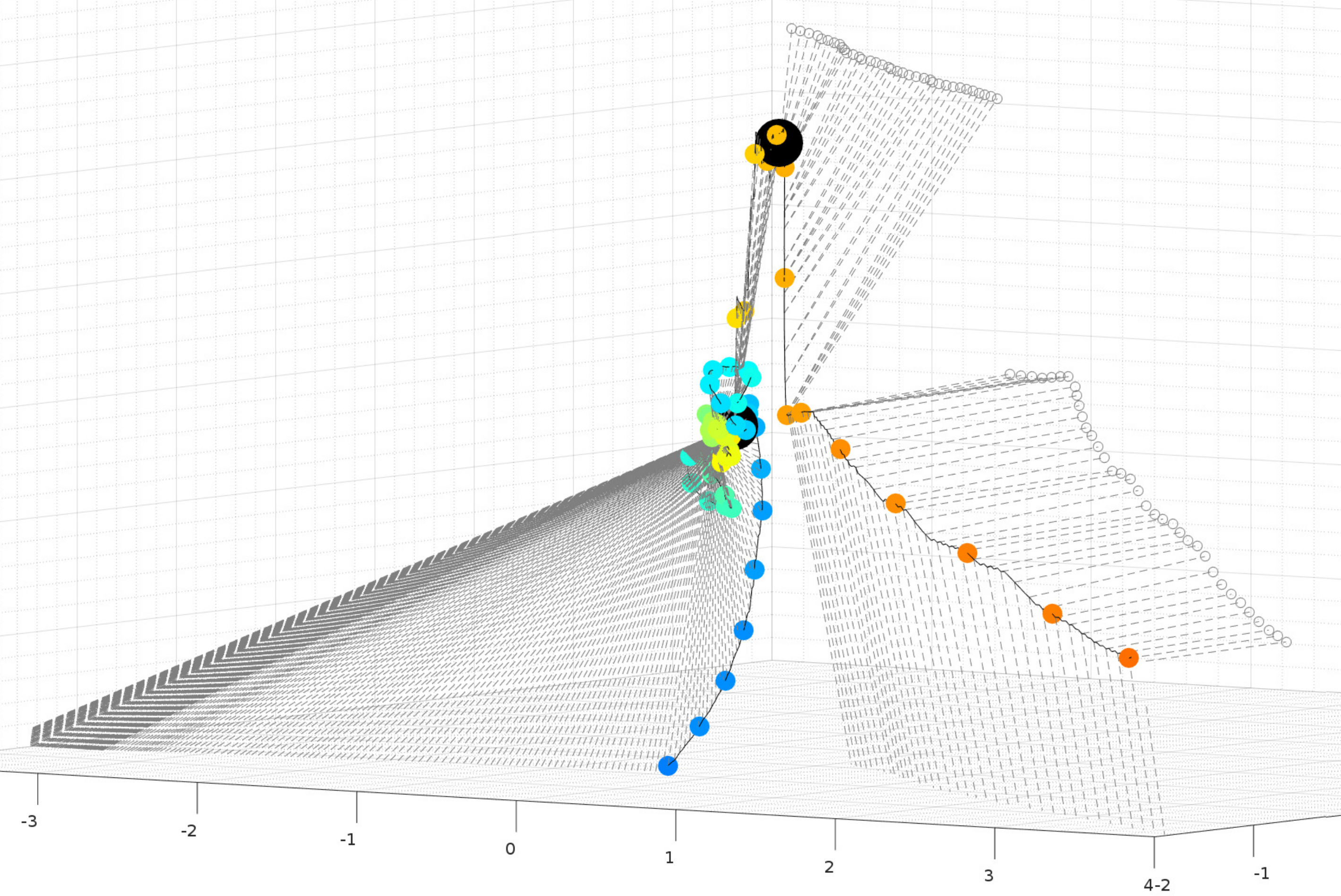}};
    \draw[line width=0.4mm, pink] (6.19,4.85) circle (0.15cm and 0.15cm);
    \draw[line width=0.4mm, yellow] (6.24,5.38) circle (0.15cm and 0.15cm);
    \draw[line width=0.4mm, green] (8.42,2.30) circle (0.15cm and 0.15cm);
    \draw[line width=0.4mm, blue] (9,2.02) circle (0.15cm and 0.15cm);
    \draw[line width=0.4mm, red] (10.16,2.02) circle (0.15cm and 0.15cm);
    \node[font=\sffamily](Start) at (2.3,1.4) {Start};
    \node[font=\sffamily](Landing) at (8.3,1.35) {Landing};
  \end{tikzpicture}
  \caption{Landing in Challenge 1. Colored markers are placed every \SI{250}{\milli\second} on the trajectory.
  Every 20th of all position setpoints (gray) is indicated with the corresponding MAV position.
  Furthermore, we marked the positions of important state machine transitions.
  The violet ring depicts the position at which the MAV leaves the exploration state and starts to rotate towards the pattern.
  Other transitions are the decision to follow the pattern (ocher ring), when the MAV decides to land (green ring), when the switches detect a landing and request motors off (blue ring), and when the motors are finally turned off (red ring).
  Although our method makes no assumption about the path of the target, we depict the approximate path of the landing pattern at a height of \SI{1.7}{\meter}. The pattern is estimated to be at an allocentric height of \SI{2.68}{\meter} at touchdown. Since all navigation is relative, nevertheless the landing succeeds. After touchdown, the position setpoint is set to a large negative value not visible in the image to prevent the MAV from taking off again. After the motors are switched off, the height estimate is reset to \SI{37}{\centi\meter}.}
  \label{fig:challenge_1}
\end{figure*}
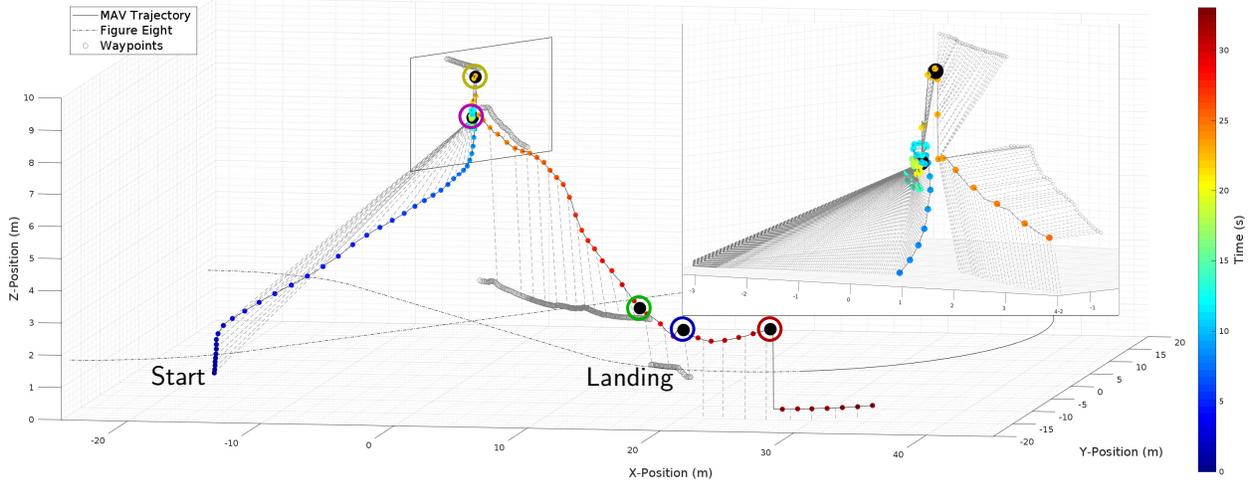

Nevertheless, we have identified several issues after the challenge by thorough analysis of the logged data depicted in \reffig{fig:challenge_1}.
First, the offset between barometric height and true height of the MAV was estimated incorrectly, caused by incorporation of the very noisy first detections of the pattern into our filter.
The offset error was reduced during the remainder of the challenge with more accurate pattern measurements but was still large directly prior to contact with the vehicle due to the fast descent.
Approaching a position \SI{1}{\meter} above the target by navigation relative to the target perceptions before the final landing decision was made, still lead to success.
When the MAV is outside a down-pointing cone with an aperture of \SI{60}{\degree}, it follows the pattern without active descent to avoid risky landings at an acute angle with the landing platform.
This behavior can be seen in the setpoints with a constant height at the very top of the excerpt in \reffig{fig:challenge_1}.
However, a second found issue yielded an alternating behavior between descent and keeping altitude until the planar distance between the MAV and the landing pattern was below \SI{90}{\centi\meter}.
This can be seen by means of the alternating MAV setpoints at the beginning of the descent in \reffig{fig:challenge_1}.
Finally, a spurious measurement very close to the pattern (approx. \SI{10}{\centi\meter} above) before touchdown changed the pattern height estimate, resulting in slightly altered localization and MAV setpoints close to the final landing (blue circle in \reffig{fig:challenge_1}).
Nevertheless, once a final landing decision has been made (green circle), the MAV descends until a successful landing is indicated by the switches on its landing feet.
The final landing position is a point \SI{40}{\centi\meter} below the pattern to ensure a determined landing up to full contact with the landing platform.

In order to fix the USB camera connection, we attached more shielding for the second trial. Unfortunately, this shielding negatively affected the compass of the MAV so that it went into failsafe mode directly after the start.
We canceled the second trial since we could not fix this issue fast enough to improve our time from the first challenge run.

\begin{figure*}[t]
  \centering
  \begin{tikzpicture}
    \node[anchor=south west,inner sep=0] (image1) at (0,0) {\includegraphics[trim=00mm 00mm 00mm 00mm,clip,width=1.0\linewidth]{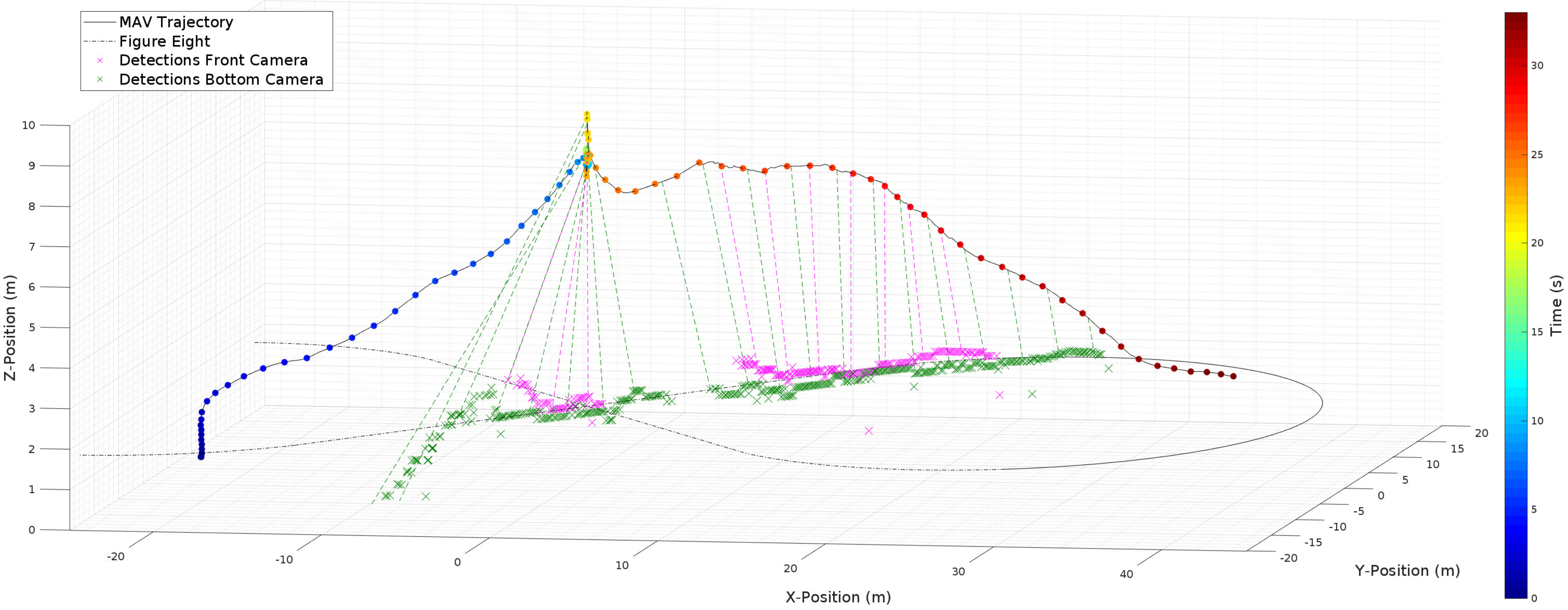}};
    \node[font=\sffamily](Start) at (2.3,1.4) {Start};
    \node[font=\sffamily](Landing) at (12.5,3.0) {Landing};
   \end{tikzpicture}
  \caption{Landing in MBZIRC Grand Challenge. First, the MAV starts in the middle of the left circle and flies straight up to a height of \SI{1.5}{\meter} to not collide with near objects. Next, it flies to the center of the field with a height of \SI{8}{\meter} to search for the landing target. The total ascent takes \SI{8.5}{\second}. After \SI{20.4}{\second}, the landing target is first detected in the bottom camera. Immediately, the MAV begins to yaw into target direction and starts to descend while tracking the target in both cameras. The descent only takes \SI{11.6}{\second}, resulting in a total completion time of \SI{32}{\second}. Due to the fast motion of the target, the MAV cannot descend fast enough to reach the target on the straight segment and has to land in the curved segment of the figure eight.
  The challenge completion time from start signal to landing is \SI{42}{\second}. Colored markers are placed every \SI{250}{\milli\second} on the trajectory. Every 20th of all 585 detections is indicated with the corresponding viewpoint on the trajectory.}
  \label{fig:grand_challenge_c1}
\end{figure*}

In the first trial of the Grand Challenge, our MAV landed in only \SI{42}{\second}. \Cref{fig:grand_challenge_c1} shows the trajectory and detections of this trial.
After reaching the center of the field, the MAV searched the target for \SI{11.9}{\second} because the moving vehicle was in a very disadvantageous position.\footnote{The starting position of the cart was randomly determined for each competitor. The nominal time for the cart to complete one half of the figure eight was \SI{\approx 27}{\second}. Since landing times of the top teams where well under 1 minute, this had significant impact on the results.}

In the second trial, we could not improve our landing time and canceled this trial after \SI{42}{\second}.

\reftab{tab:Evaluation_of_Target_Detections} shows a quantitative evaluation of target detections in the individual cameras in Challenge 1 and Grand Challenge. In both successful landings, the bottom camera was more valuable regarding information gain. In particular in Challenge 1, the front camera was not very valuable, giving less then \SI{1.5}{\second} of valid detections. This was caused by a cable that accidentally came loose during the repair of the USB connection, which was covering a large portion of the camera lens. The importance of the bottom camera is also reflected in the detection percentage of the individual cameras. It is calculated by the number of successful detections per total number of images between the first and the last detection in one of the cameras.
In Challenge 1, after the first detection the tracker continuously tracked the target, while in the Grand Challenge, the tracker lost the target once in each camera. With only \SI{52}{\centi\meter}, also the variance of the first detection distance is much smaller in the bottom camera than in the front camera. Since the front camera is directed, the first detection distance heavily depends on the yaw of the MAV (which is arbitrary) when the target arrives near the MAV. Since the MAV yaws with only \SI{0.1}{\hertz}, the vehicle could travel up to \SI{41.6}{\meter} in a worst case scenario until a first detection in the front camera.
The higher resolution of the cameras accounts for a significant increase in detection range in comparison to other competition participants. In particular, we found the use of attitude readings from the MAV to be highly reliable and beneficial. While all other teams relied on some form of altitude measurements as we did, also correcting (parts of) the camera image for tilt (see \refsec{sec:Visual_Perception}) simplified all detection problems considerably and spared an involved outlier rejection afterwards. In an earlier stage of the preparation, we experimented with more sophisticated detectors relying on ellipses in order to detect a tilted pattern. However, these turned out to be either to slow and yielded an unacceptably high number of false positives.

\begin{table}[t]
\small
\caption{Evaluation of Target Detections}
\label{tab:Evaluation_of_Target_Detections}
\begin{center}
\begin{tabularx}{.85\columnwidth}{p{.3\columnwidth}>{\centering\arraybackslash}
p{.1\columnwidth}>{\centering\arraybackslash}p{.1\columnwidth}>{\centering\arraybackslash}p{.1\columnwidth}>{\centering\arraybackslash}p{.1\columnwidth}}
    \toprule
    & \multicolumn{2}{c}{Challenge 1 Trial 1} & \multicolumn{2}{c}{Grand Challenge Trial 1} \\
    & Front & Bottom & Front & Bottom \\
    \midrule
    \parbox[t]{0.35\columnwidth}{Detections}                            & 57                 & 322                & 199                & 386\\[0.5ex]
    \parbox[t]{0.35\columnwidth}{Detection percentage}                  & \SI{17}{\percent}  & \SI{96}{\percent}  & \SI{47}{\percent}  & \SI{92}{\percent}\\[0.5ex]
    \parbox[t]{0.35\columnwidth}{Average detection rate}                & \SI{30.20}{\hertz} & \SI{38.24}{\hertz} & \SI{27.97}{\hertz} & \SI{36.66}{\hertz}\\[0.5ex]
    \parbox[t]{0.35\columnwidth}{Average tracking rate}                 & \SI{30.20}{\hertz} & \SI{38.24}{\hertz} & \SI{40.02}{\hertz} & \SI{38.96}{\hertz}\\[0.5ex]
    \parbox[t]{0.35\columnwidth}{Estimated distance of first detection} & \SI{11.95}{\meter} & \SI{16.85}{\meter} & \SI{7.19}{\meter}  & \SI{17.37}{\meter}\\[0.5ex]
    \bottomrule
\end{tabularx}
\end{center}
\end{table}

Analysis of the data from the competition after MBZIRC showed, that the MAV only needed \SI{3.12}{\second} in Challenge 1, respective \SI{1.76}{\second} in the Grand Challenge, to accelerate to the horizontal target velocity of \SI{4.17}{\meter\per\second}.

During both landings, the MAV always landed on the outer part of the target. One foot did not even touch the platform in the Grand Challenge. This behavior was anticipated beforehand since our target estimation filter as well as the MPC both assume a constant velocity of the target. Since the MAV landed in the curved segments of the figure eight, the movement of the target as well as the interception point is always projected outwards of the circle. Due to the fast replanning, the accuracy was nevertheless enough to land on the target. 

\subsubsection{Treasure Hunt}
\begin{figure}[t]
  \centering
  \setlength{\figureheight}{0.5\linewidth}
  \includegraphics[height=\figureheight]{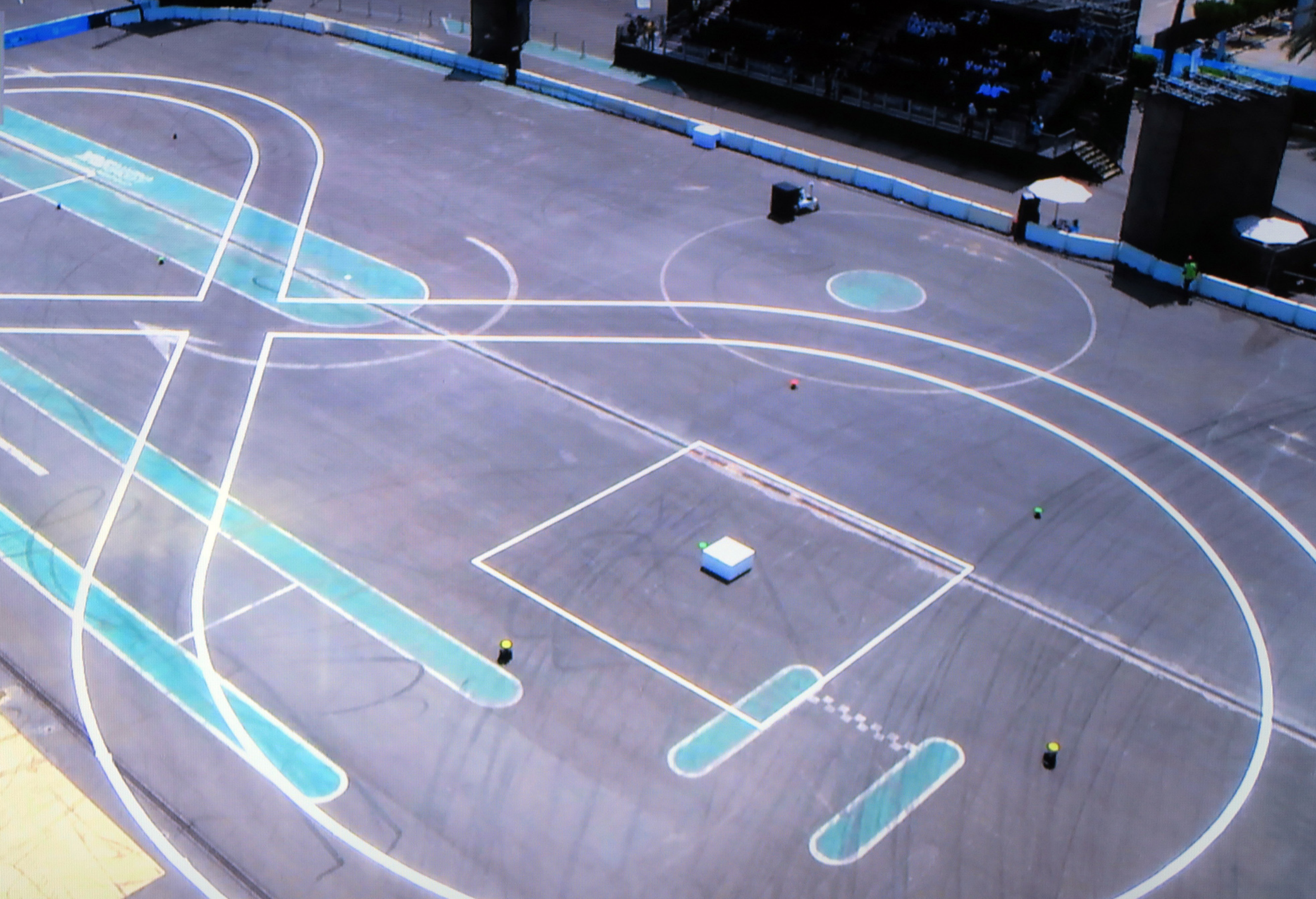}
  \includegraphics[height=\figureheight]{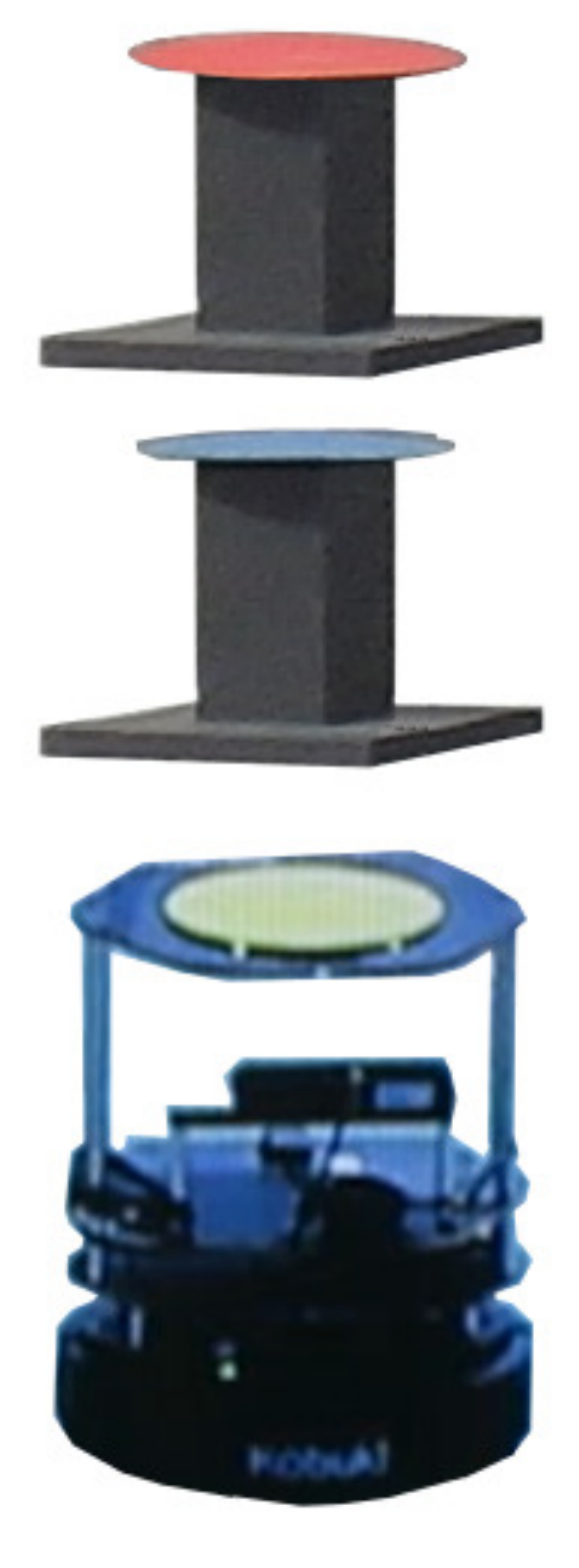}
  \caption{Arena at MBZIRC. The colored disks randomly distributed over the arena had to be detected, picked, and dropped into the white box or its surrounding drop zone. Many white lines and colored markings on the ground posed a challenge for object detection. Right: Closeup of two static objects (red and blue) and a moving object (yellow).\label{fig:arena}}
\end{figure}

In the first attempt of the first trial, we began to explore the arena with three MAVs simultaneously.
After few minutes, the trial was canceled by the organizers because of very strong winds with a speed of up to \SI{9}{\meter\per\second}.
Qualitatively, all MAVs followed their assigned exploration trajectories until then.

\begin{figure}[t]
  \centering
  \includegraphics[width=0.24\linewidth]{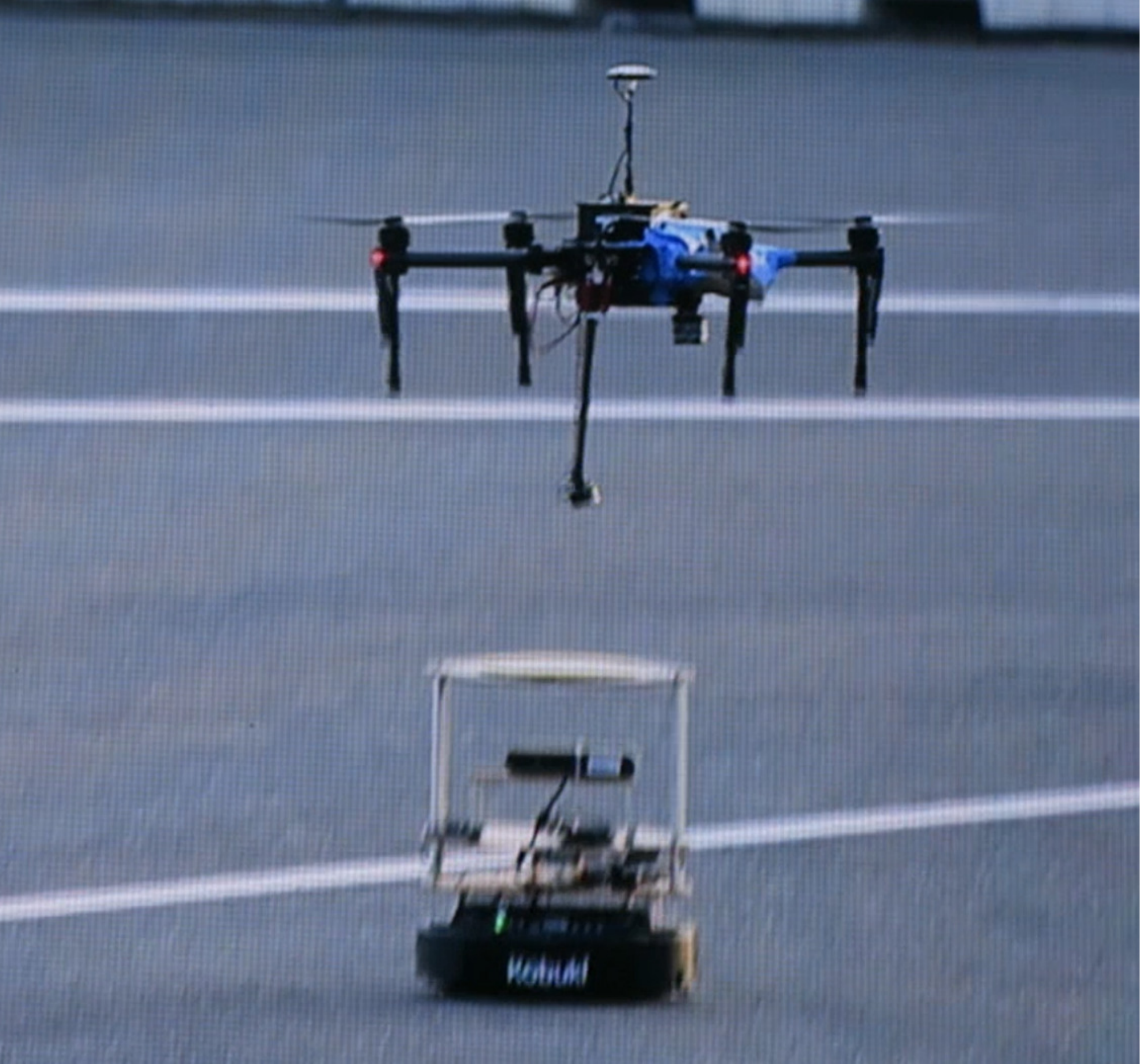}\,
  \includegraphics[width=0.24\linewidth]{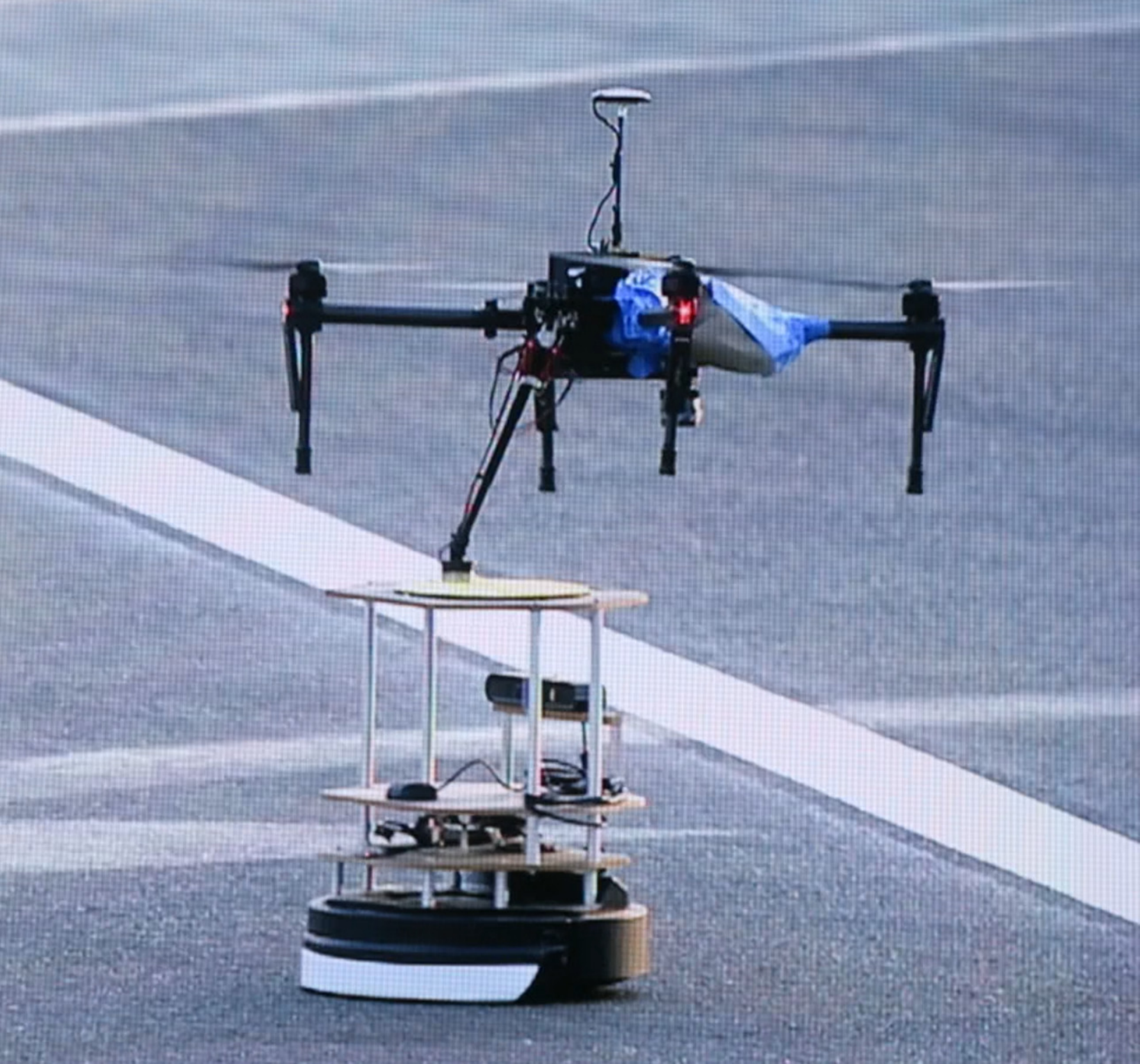}\,
  \includegraphics[width=0.24\linewidth]{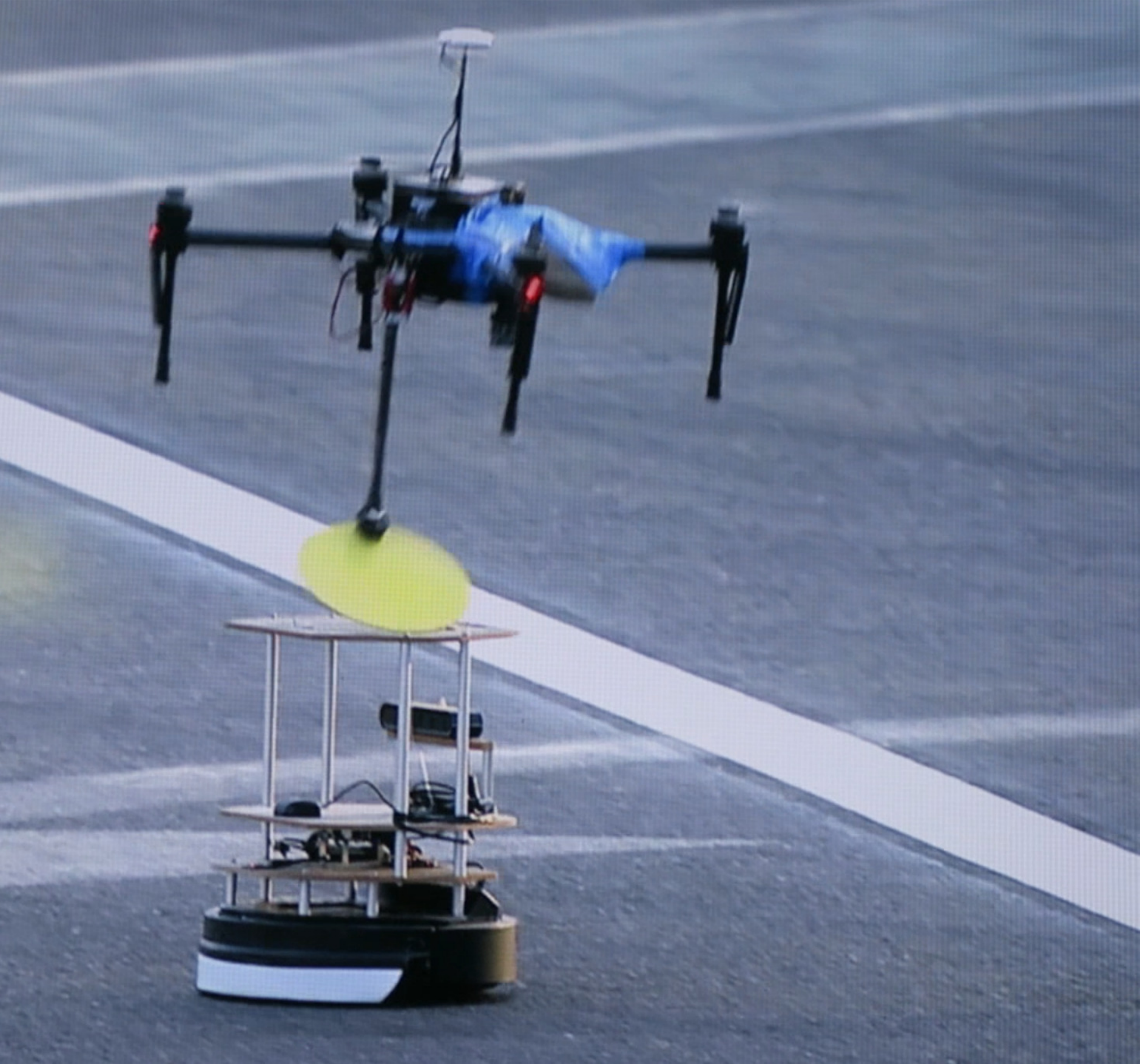}\,
  \includegraphics[width=0.24\linewidth]{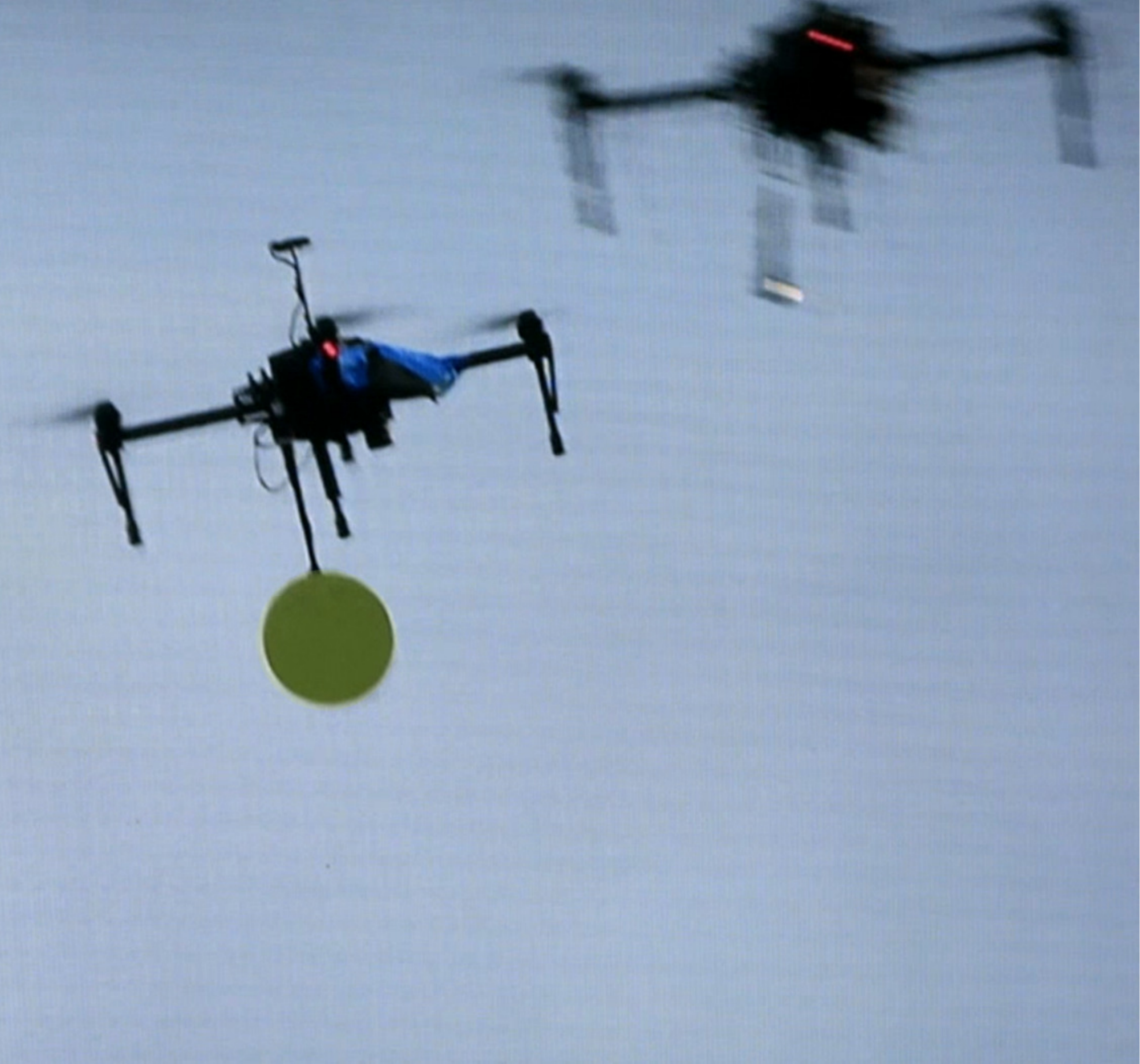}
  \caption{Picking a moving object. Our MAV follows the moving yellow disk with visual servoing.
  The telescopic rod and the ball joint of our electromagnetic gripper allow flexible picking without disturbing attitude control of the MAV.
  The picked objects were delivered to a drop box up to \SI{75}{\meter} away.}
  \label{fig:teaser_c3}
\end{figure}

In the second attempt of the first trial, we explored the arena with three MAVs and successfully picked two disks from moving bases.
One of the disks was delivered into the drop box.
\Cref{fig:teaser_c3} shows an example pick and delivery of a moving object.
Before the second disk could be delivered, the referees called a reset---one of the MAVs seemed to approach the allowed altitude limit---and the MAV landed with the disk still attached.
Due to conservative safety distances to the ground, we could not pick that disk which was no longer elevated on a stand, but on the floor after the reset.
Furthermore, two MAVs arrived at the drop zone at the same time and were kept in a deadlock situation.
Modifications to the system during the competition were not allowed, so we could only address these issues between trials.
This was the fourth-best result of all 36 Challenge 3 trials---18 teams with two trials per team where the better trial counted for the final score---in the Treasure Hunt and worth a third place.

The second trial took place with very strong wind.
Objects were detected reliably and the descent of the MAVs was stable despite the wind, but the MAVs always had an offset of a few centimeters into the wind direction when picking. Due to this issue, we were not able to improve our result from the first trial.

\begin{figure}[t]
  \centering
  \includegraphics[trim=30mm 00mm 10mm 00mm,clip,width=0.49\linewidth]{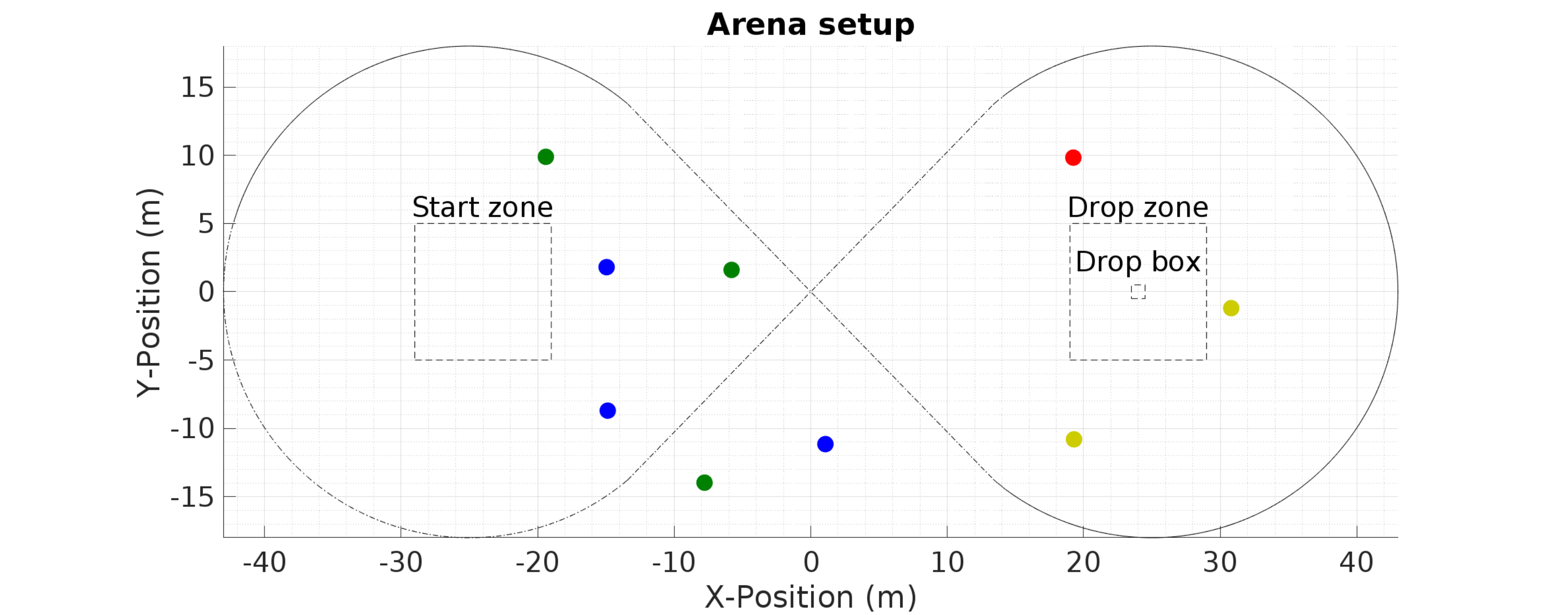}~
  \includegraphics[trim=30mm 00mm 10mm 00mm,clip,width=0.49\linewidth]{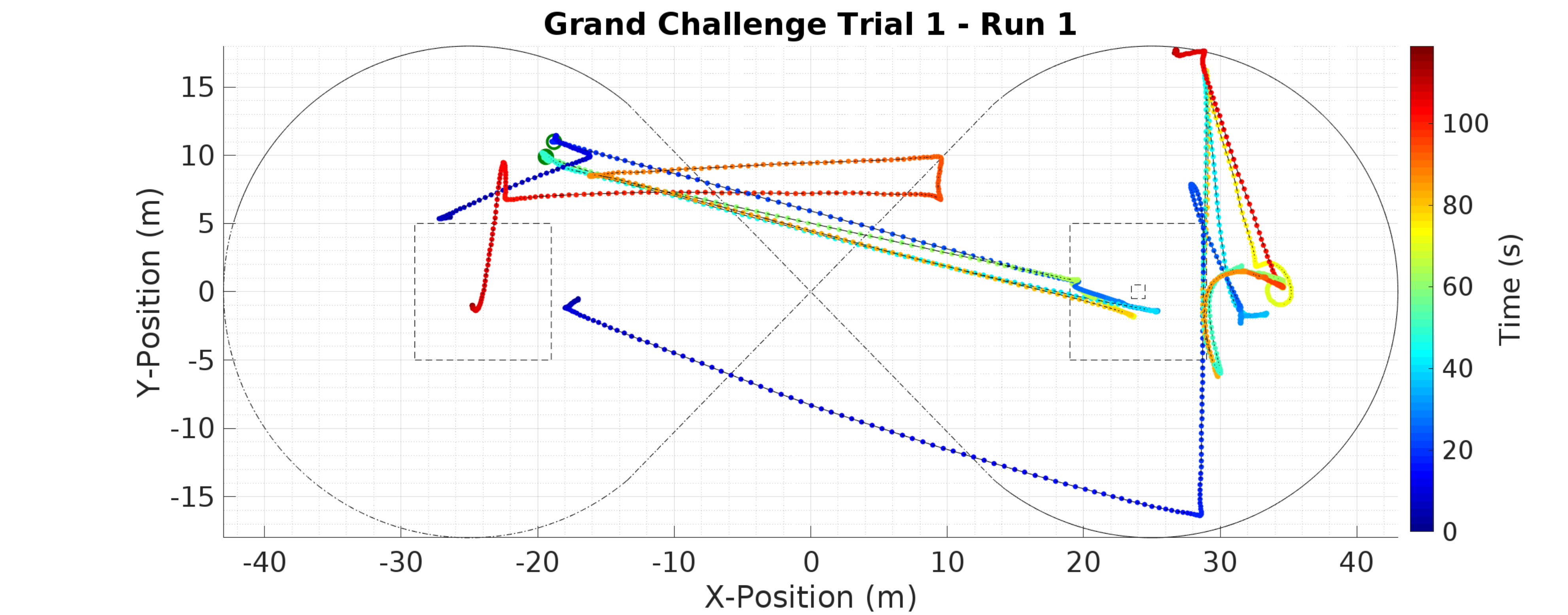}\\
  \includegraphics[trim=30mm 00mm 10mm 00mm,clip,width=0.49\linewidth]{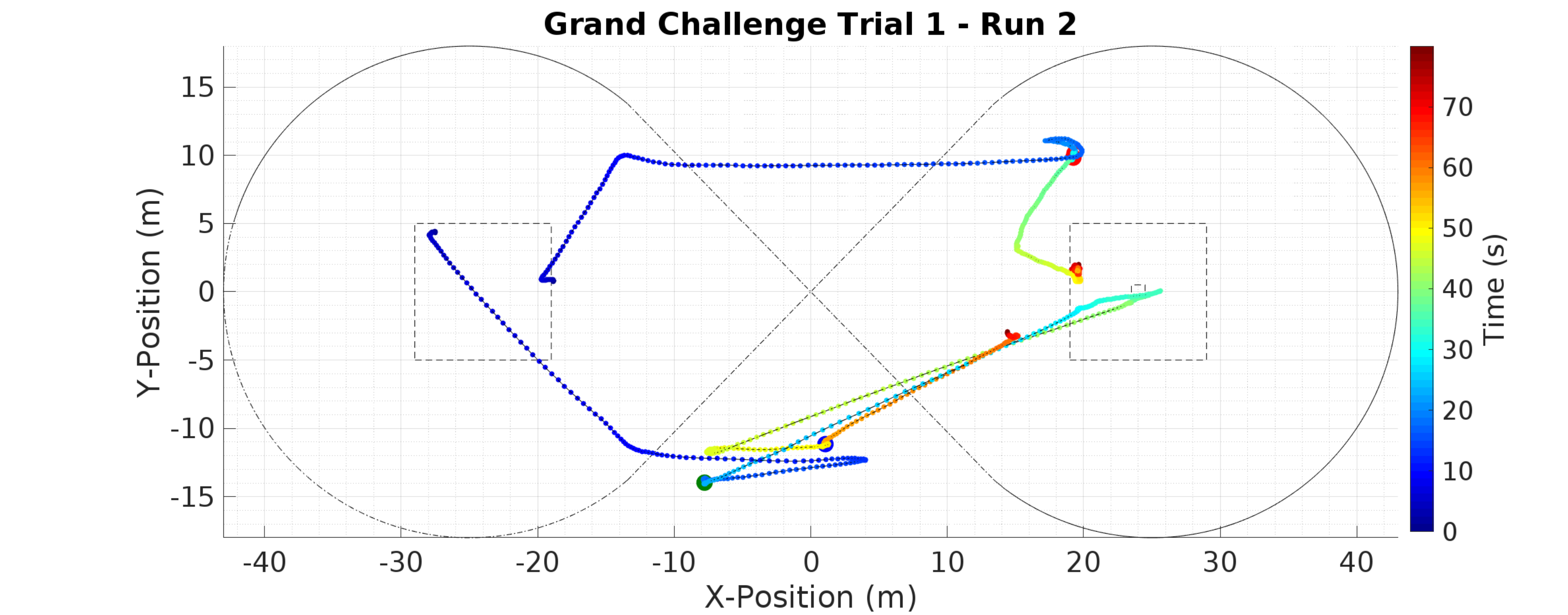}~
  \includegraphics[trim=30mm 00mm 10mm 00mm,clip,width=0.49\linewidth]{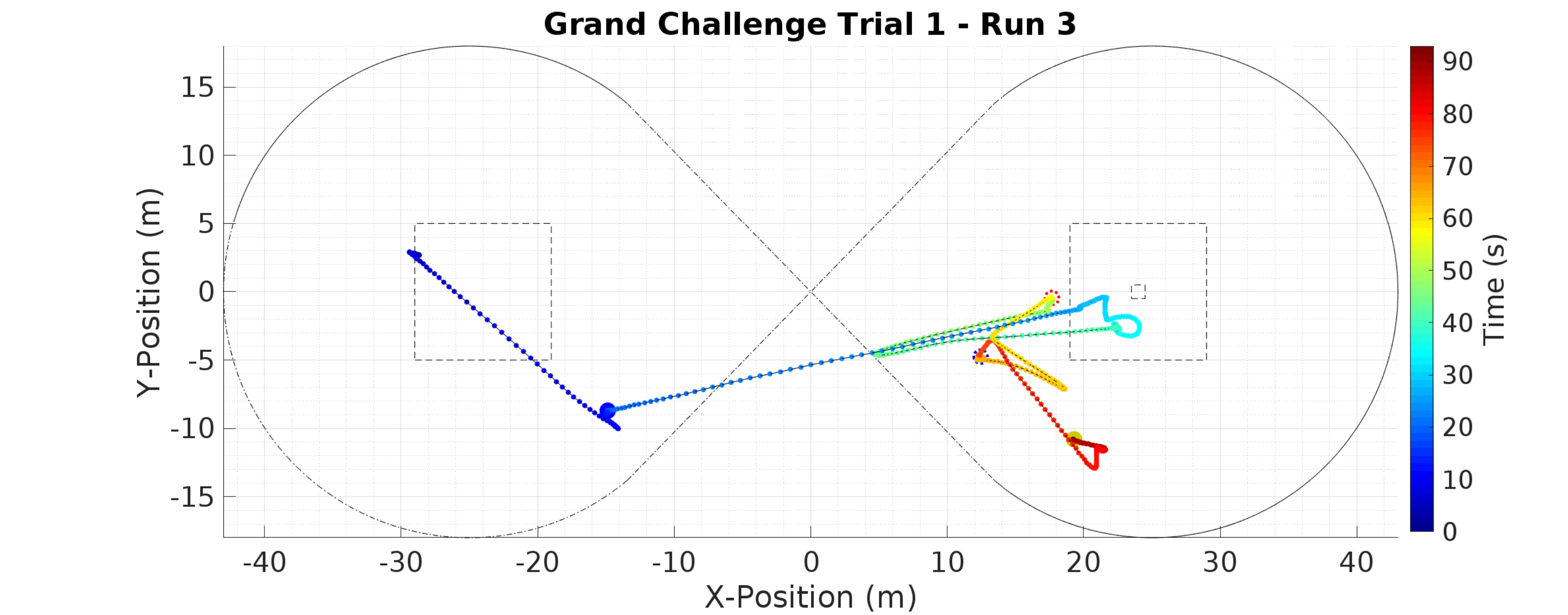}\\
  \includegraphics[trim=30mm 00mm 10mm 00mm,clip,width=0.49\linewidth]{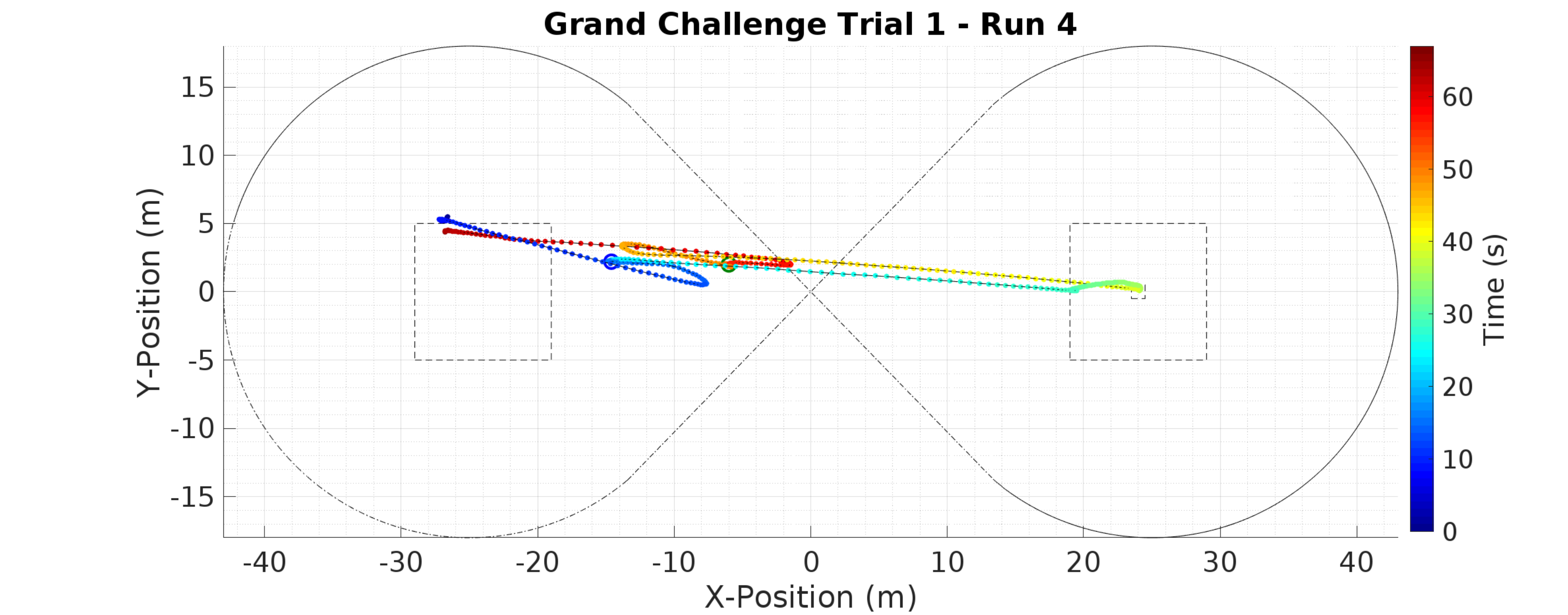}~
  \includegraphics[trim=30mm 00mm 10mm 00mm,clip,width=0.49\linewidth]{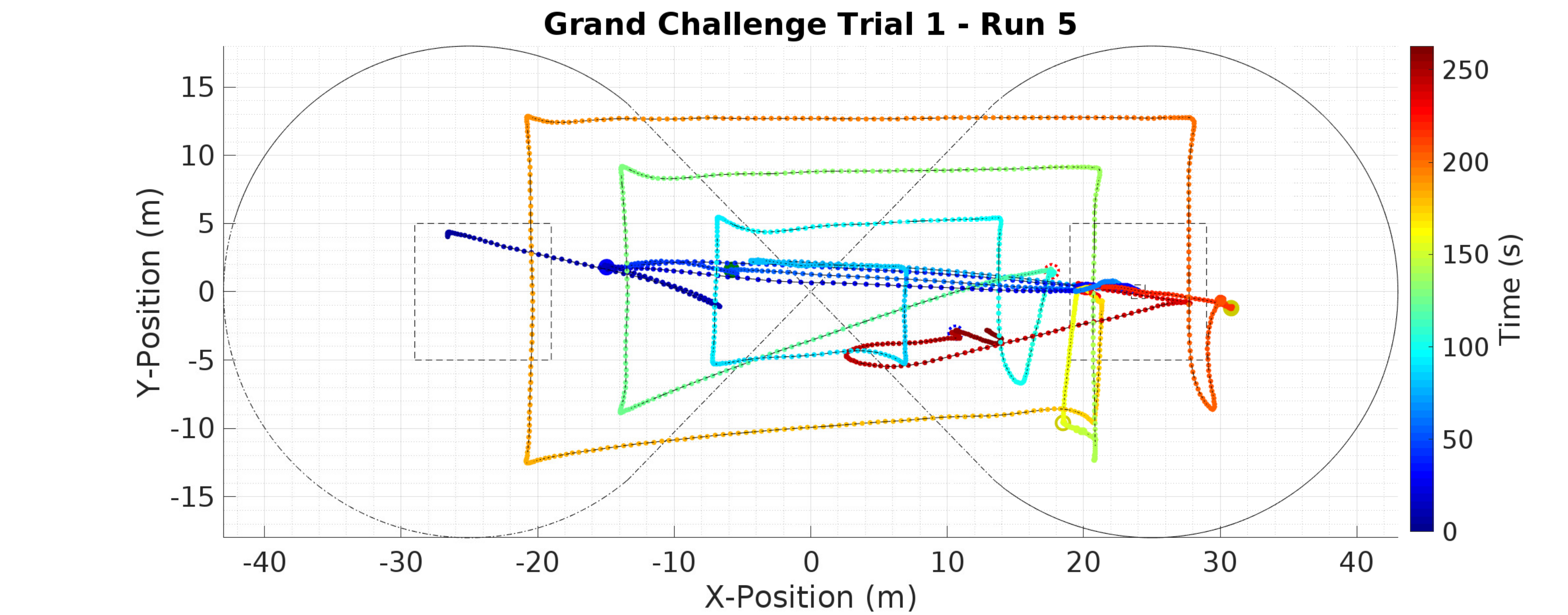}
  \caption{Treasure Hunt in the Grand Challenge Trial 1. Each image shows the trajectories of the active MAVs during the 5 runs (separated by 4 resets). Solid disks represent successful and rings show missed picks. The dotted disks indicate disks lying on the ground. The left rectangle is the starting zone, the right one the drop zone including the drop box.
  In Run~1 and Run~2, two MAVs were active. In Runs~3-5, only one MAV was active since the other one worked erroneously. It flew way to high so we had to call a reset.
  The following colored disks were picked (p) and missed (m) during the Grand Challenge: Run~1: \textcolor{MBZIRCgreen}{\bf m}-\textcolor{MBZIRCgreen}{\bf p}; Run~2: \textcolor{MBZIRCgreen}{\bf p}-\textcolor{MBZIRCred}{\bf m}-\textcolor{MBZIRCred}{\bf p}-\textcolor{MBZIRCblue}{\bf p} (the blue and the red disk had to be put on the ground, because we called a reset. Each disk is attempted to be picked twice later); Run~3: \textcolor{MBZIRCblue}{\bf p}-\textcolor{MBZIRCred}{\bf m}-\textcolor{MBZIRCblue}{\bf m}-\textcolor{MBZIRCyellow}{\bf p} (the yellow disk was picked during a reset and had to be put back on the cart); Run~4: \textcolor{MBZIRCblue}{\bf m}-\textcolor{MBZIRCgreen}{\bf m}; Run~5: \textcolor{MBZIRCblue}{\bf p}-\textcolor{MBZIRCgreen}{\bf p}-\textcolor{MBZIRCred}{\bf m}-\textcolor{MBZIRCyellow}{\bf m}-\textcolor{MBZIRCyellow}{\bf p}-\textcolor{MBZIRCblue}{\bf m}.
  Total time airborne is \SI{624}{\second}.}
  \label{fig:grand_challenge_evaluation_c3}
\end{figure}

In the first trial of the Grand Challenge (\reffig{fig:arena} shows the arena setup in the Grand Challenge), we started with three MAVs. One failed directly at takeoff due to a minor hardware defect (broken motor speed controller) resulting from the preceding challenge (and could not be detected beforehand due to the flying ban between challenges).
The other two explored the arena and started picking and delivering objects.
As the field was not covered in full due to one missing MAV, we reconfigured the system to use only two MAVs in a reset---a modification approved by the judges on site. 
\Cref{fig:grand_challenge_evaluation_c3} reports our trial in detail.
\SI{34}{\second} after the reset, we experienced a problem with the laser-based height correction resulting in one MAV flying too high.
After a second reset---we took out the malfunctioning MAV as the cause of the problem was not apparent---the remaining MAV operated on the whole arena.
We successfully picked nine disks and were able to deliver seven of them---six into the drop zone and one into the drop box.
Two disks were still attached to MAVs during a reset and, thus, were lying on the ground after resuming the trial.
Overall, we scored 10.5 points and reached a second place in this Grand Challenge subtask.

We canceled the second Grand Challenge trial due to severe hardware issues without a score.

\begin{figure}[t]
  \centering
  \begin{tikzpicture}
    \node[anchor=south west,inner sep=0] (image1) at (0,0) {\includegraphics[trim=00mm 00mm 00mm 00mm,clip,width=1.0\linewidth]{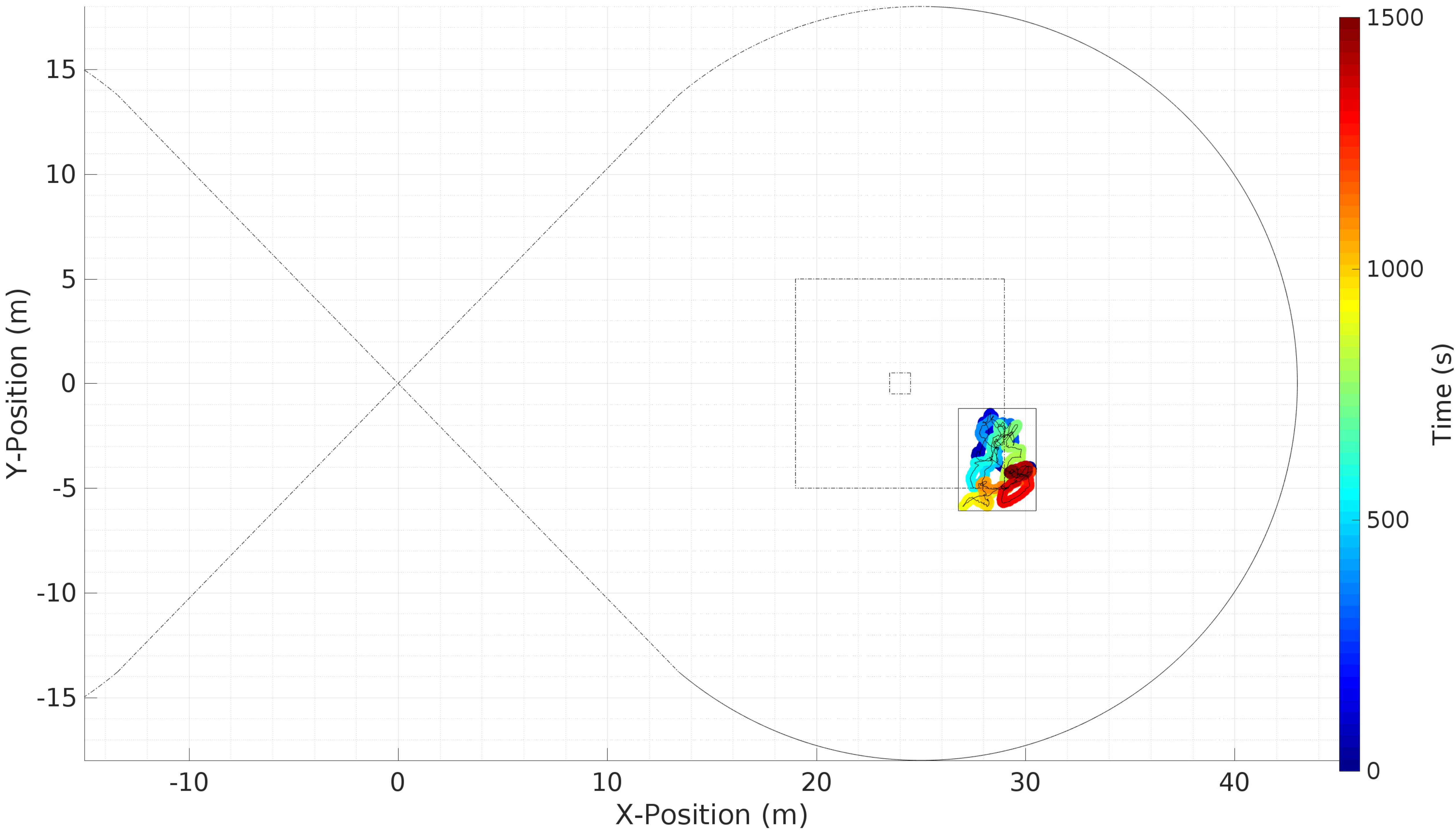}};
    \node[anchor=south west,inner sep=0.1,draw=black] (image2) at (1.8,1.0) {\includegraphics[trim=00mm 00mm 00mm 00mm,clip,width=0.38\linewidth]{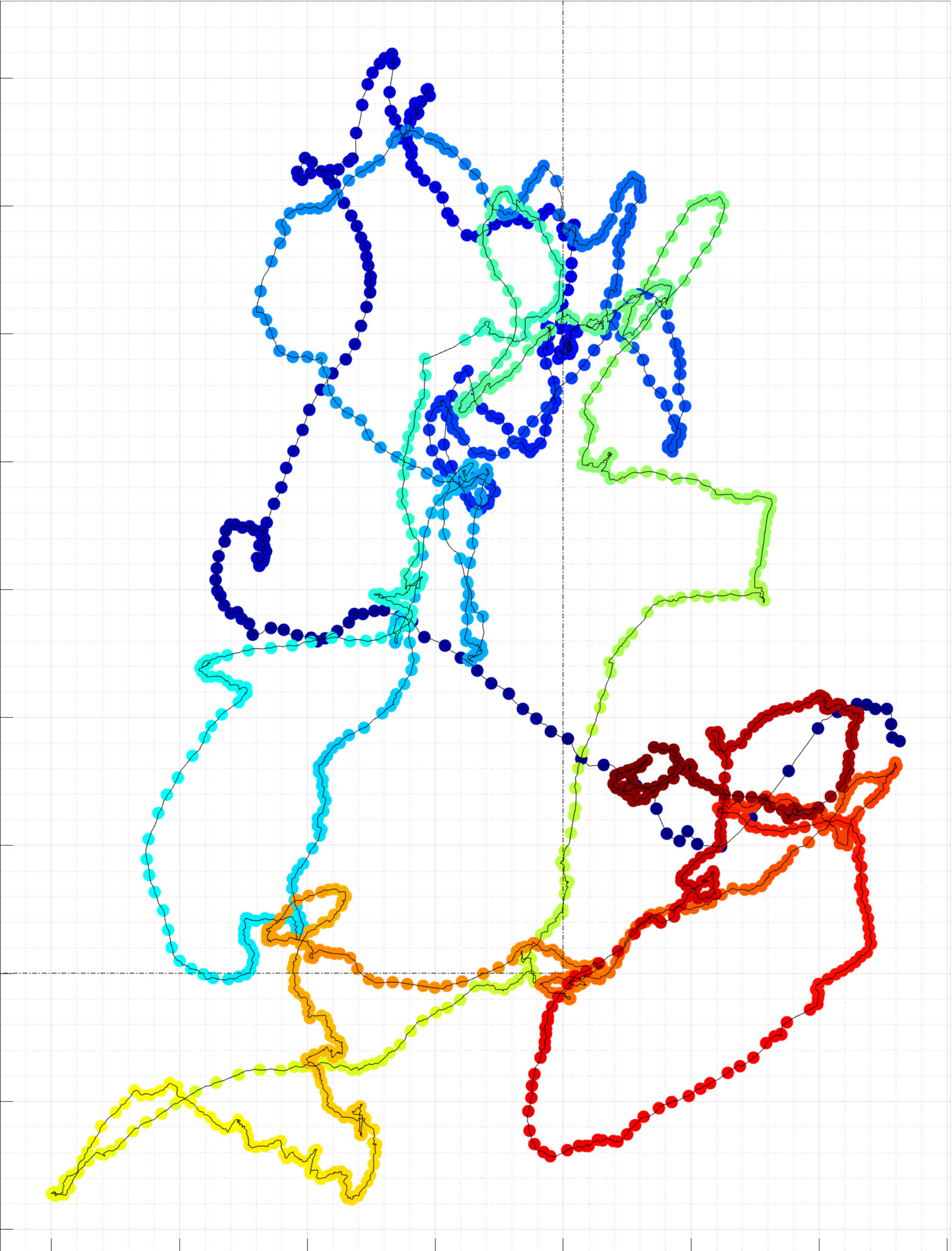}};
  \end{tikzpicture}
  \caption{GNSS drift during Grand Challenge Trial 1 caused by atmospheric effects. The offset of the field center is broadcasted to the individual MAVs and the ground robot to compensate for the drift.
  The colored markers depict the estimated uncorrected position of the drop box center over time. Markers are placed every \SI{1}{\second} on the trajectory.}
  \label{fig:grand_challenge_drift}
\end{figure}

During the competition, we used the base station to send GPS offset corrections to the MAVs and the ground robot. \Cref{fig:grand_challenge_drift} shows the measured offset of the drop box center during Grand Challenge Trial 1. It can be seen that without the correction, the robots would have experienced a static horizontal offset of up to \SI{7.9}{\meter} after \SI{1280}{\second}. This would not only hinder the detection of the drop box, but also lead to collisions with the safety net. When not using the dynamic offset but hard-coding the offset at the beginning of the trial, the maximum deviation would have been only \SI{3.76}{\meter} after \SI{914}{\second}.

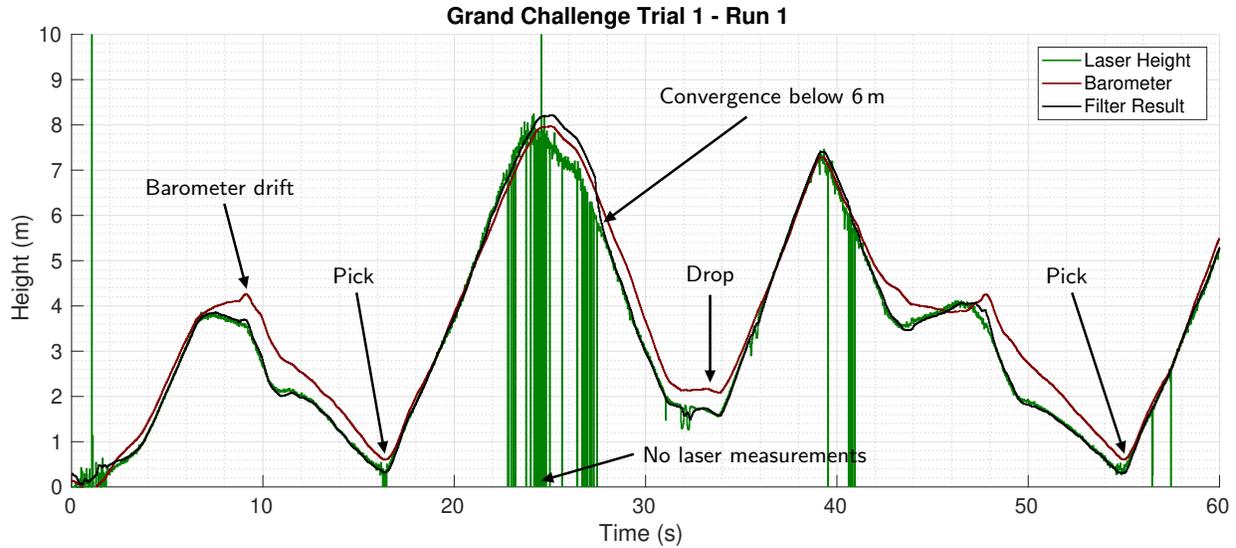
\begin{figure}
  \resizebox{1.0\linewidth}{!}{
    \begin{tikzpicture}[font=\sffamily\small,>={Stealth[inset=0pt,length=4pt,angle'=45]}]]
      \node[inner sep=0pt] (plot) at (0,0) {\includegraphics[width=13.7cm]{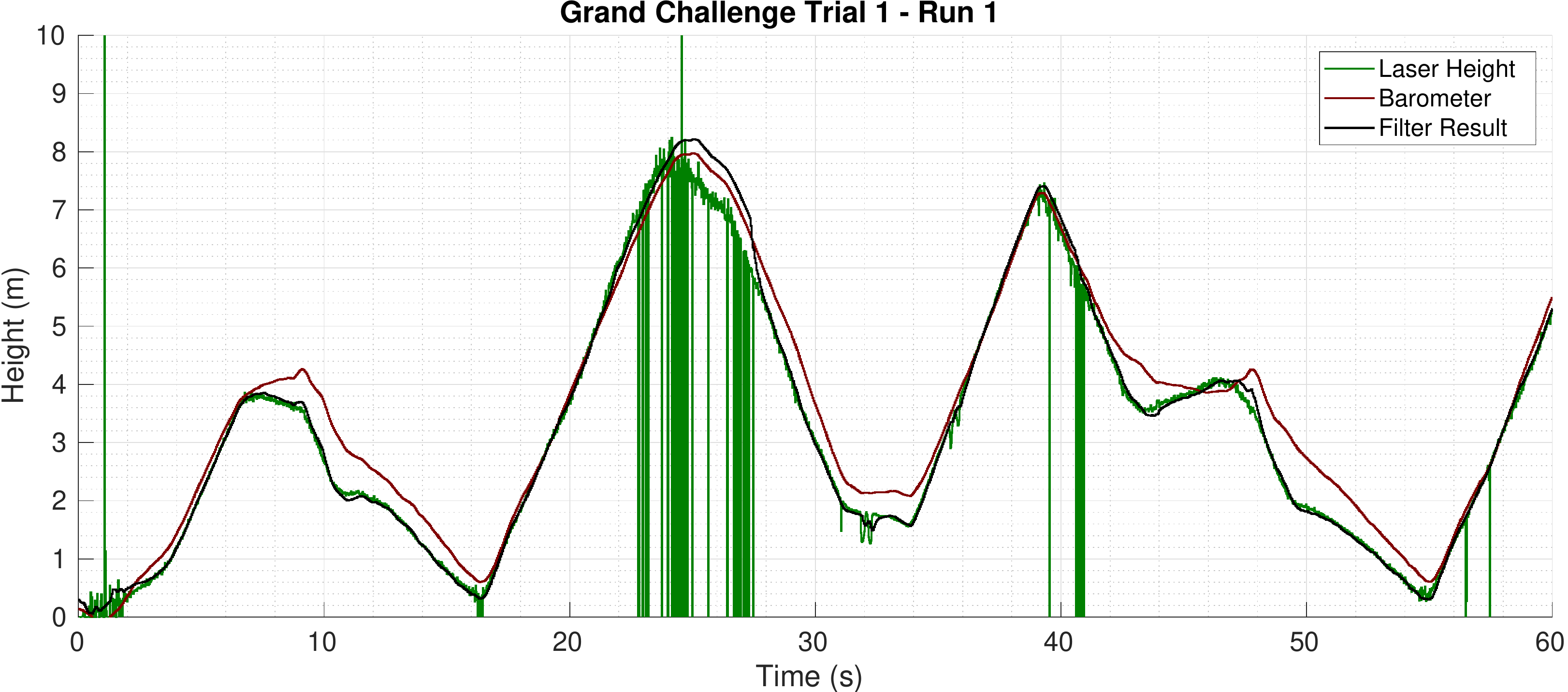}};
      \node[label,scale=0.8] (pick_1) at (-3,0) {Pick};
      \node[label,scale=0.8] (pick_2) at (5,0) {Pick};
      \node[label,scale=0.8] (drop) at (1,0) {Drop};
      \draw[thick,->] (pick_1) -> (-2.65,-2);
      \draw[thick,->] (pick_2) -> (5.65,-2);
      \draw[thick,->] (drop) -> (1,-1.2);

      \node[label,scale=0.8] (drift) at (-4.5,1) {Barometer drift};
      \draw[thick,->] (drift) -> (-4.2,-0.15);

      \node[label,scale=0.8] (laser) at (1.5,-2) {No laser measurements};
      \draw[thick,->] (laser.west) -> (-0.9,-2.3);

      \node[label,scale=0.8] (convergence) at (1.7,2) {Convergence below 6\,m};
      \draw[thick,->] (convergence) -> (-0.20,0.6);
    \end{tikzpicture}
  }
  \caption{Laser height correction. We correct the barometer (red) drift over time with a down-pointing laser (green) when the MAV is below \SI{6}{\meter}. Above \SI{6}{\meter} the laser becomes unreliable in the bright outdoor conditions and the height estimate (black) follows the barometric altitude measurements corrected by an estimated offset. Shown is an excerpt of our first Grand Challenge run with two object picks and one delivery to the drop box.\label{fig:laser_height}}
\end{figure}

During the picking challenges, we corrected the barometric height estimate with laser measurements at lower altitudes.
\Cref{fig:laser_height} depicts an excerpt of the height estimates in the Grand Challenge Trial 1 and shows that both measurements can significantly deviate due to drift---up to \SI{1}{\meter} in the illustrated flight.
Especially in low altitudes during picking or dropping close to the box, an accurate height estimate is crucial.
As stated before, in this run we experienced a problem with the laser-based height correction resulting in one MAV flying too high.
Due to the proximity of time of both Grand Challenge trials on the same day, the problem could be identified only after the competition:
In contrast to tests ahead of the competition, erroneous laser measurements above an altitude of \SI{6}{\meter} could report low but valid distance values and got incorporated into the bias correction---presumably because of the bright sunlight.
These measurements could not be filtered out reliably, as they were indistinguishable from valid measurements.
This resulted in a self-enhancing problem as the MAV climbed even more until it hit a hard safety constraint at \SI{20}{\meter} based on the low-level barometric measurements.
Due to altitude separation, the MAVs with higher IDs were more affected by this problem as they were assigned to higher altitudes when approaching the drop box.
As a consequence, the first MAV operated reliably, while the third MAV was repeatedly affected until we took it out of the competition.

\section{Lessons Learned}
\label{sec:Lessons_Learned}

As described in \refsec{sec:System_Setup}, we experienced major issues with the interference between GPS and the onboard computer during preparation of the competition. During our tests on site, the problems became even more severe. We found that also the USB 3.0 cameras and the cables were strongly interfering with the GPS signal. Consequently, we shielded the cables during the competition with extra layers of Aaronia X-Dream EMC shielding fleece. As this also influenced the magnetometer, a recalibration of the magnetic sensor was necessary. We were not aware of this issue until our landing MAV refused to switch to autonomous mode during Challenge 1 Trial 2.

Furthermore, the achieved frame rate of the cameras dropped and we experienced connectivity errors that required to unplug and reconnect the camera to the bus. The problem was most severe when multiple cameras shared the same USB bus like on our landing MAV. We increased system stability by connecting the second camera to the USB-C connection of our computer, which is internally connected to a different USB 3.0 bus and host controller, avoiding interfering with the other peripheral like landing gear switches, laser and the flight control unit of our MAV.

One major challenge was posed by the non-availability of a testing area beyond the trial runs. Hence, hard- or software changes could not be tested. As reported in the evaluation, this caused Challenge 1 Trial 2 to fail and also triggered the loss of one MAV in Challenge 3 Trial 1. These issues could have been easily fixed within minutes but cause a complete trial to fail when undetected. Time slots for free testing available to all teams, e.g. in the evening hours, could mitigate this issue in future competitions. 

Late changes of the challenge rules and specifications posed another major challenge for all teams. The arena surface, \eg contained many more white and colored lane markings than the expected figure eight, start, and drop zones. Size and shape of the drop box as well as the height of the landing vehicle were different from the draft specifications given to the teams beforehand. Furthermore, the disks in Challenge 3 were much heavier and the moving disks slower. Thus, hardware and software approaches that were not too much tailored to the specifications and easily adaptable were in favor. For example, the higher weight of the disks posed a challenge for some teams' MAV controllers and the mechanical structure of the gripper. Our gripper was designed to be equipped with a variable number of small permanent magnets in addition to the electromagnet to quickly adapt to the weight of the actual objects on site. As a matter of fact, the heavier disks induced a much larger magnetic flux as the beforehand specified objects such that we could omit the permanent magnets completely.

Further difficulties arose from the weather conditions. Strong, steady wind---especially in the afternoon trials---violated the static world assumption and had to be compensated for, which was not done by our MPC. We strongly underestimated the maximum wind speeds---approximately up to \SI{9}{\meter\per\second}---until which the trials will still take place. An appropriate wind compensation mechanism would have been helpful. We were forced to adapt safety margins manually prior to each trial to compensate for the changing weather conditions to avoid contact with the surrounding net.

Although we had a distinct GUI at hand, particularly during the Grand Challenge, controlling multiple MAVs in parallel was very demanding for a single operator. During peak, he was supervising over 30 individual processes on four MAVs, each in a single terminal including the recording of debugging data.
Furthermore, the configuration for individual challenges had to be updated by hand on each system if last minute changes prior to the competition were necessary.
For future competitions, further automation of the configuration distribution and process management are inevitable.
Also, we would like to distribute the tasks on multiple team members and enforcing a command chain like we did in \citet{nimbro_networking} if possible.
At MBZIRC 2017, this was constrained by the number of allowed team members in the control tent from which one per MAV acted as a safety pilot who could not effectively perform other tasks simultaneously.

Post-competition, the thorough analysis of logged data and the extensive simulation of Challenge~3 based on the competition source code base gave valuable insights about problems that occurred during the competition, including problems that were not apparent during on site operation, but still had an impact on the overall performance.
Although post-competition analysis cannot help to perform better at the competition, some of these insights can directly help to improve system parts---hardware and software---that are used in follow-up projects and competitions. Other findings can indirectly help to avoid making comparable mistakes.

\subsection{Landing}
\label{sec:Lessons_Learned_c1}
We report that the MAV always landed on the outer part of the pattern. If the target would have been smaller (or the velocity higher or the radius smaller and thus the acceleration higher), both filter and MPC would have needed to account for this by, \eg using a constant acceleration prediction or by including information about the anticipated movement of the target on the figure eight. The latter method, however, would have led to a loss of generality.

As seen in the evaluation, the total landing time is dominated by the descent phase. This could be decreased by searching for the target at a lower height than \SI{8}{\meter}.

When searching for the target, the MAV yaws with \SI{0.1}{\hertz}. This leads to a worst case detection delay in the front camera of \SI{10}{\second}. Since the vehicle can only arrive from four different directions (from which only two are valid, considering the know direction of movement), a more discontinuous yawing strategy could be advantageous. The same counts for the location at which to search for the target. Searching for the target at one of four (two) start points of the straight segments of the figure eight could prevent landings in the curve.

\subsection{Treasure Hunt}
\label{sec:Lessons_Learned_c3}
During the Treasure Hunt, the arena held more challenges than most teams expected. While color-based detection as such is always prone to failure when the target scenario changes, in particular the dynamic weather and lighting conditions during the competition, as well as specular reflection from grains of sand, impeded vision-based detection in many cases. Prior to each challenge, we hence reconfigured the exposure time and white balance. Our probabilistic color model (see \refsec{sec:Pickable_Object_Detection}) turned out to be a valuable approach as it could quickly be extended with each new trial. Thresholding the pixel values in certain color spaces, as intended by many other teams, did not turn out to be comparably reliable. Especially the yellow-colored objects had a high reflectance and appeared in various shades depending on the current lighting conditions. In combination with the sand that was blown into the arena, this caused false hypotheses on the arena ground. However, these were automatically discarded as their size, shape, and aspect ratio were not fitting. Vice versa, this also caused false negative detections under certain lighting conditions. However, since other objects were detected reliably and no preference for the more valuable yellow objects was given mission-wise, this did not present a strong disadvantage during the challenges. Quite on the contrary, the drop box detection did perform very poorly in both competition trials and missed nearly all detections. A later close inspection of the footage and the source code revealed that the detector worked as intended, but as the objects were heavier than expected, they dangled below the copter and occluded large parts of the camera image. This hindered detection in particular at lower altitudes and did not allow for a reliable tracking during descent. As already reported in the evaluation, the speed of the moving disks was found to be negligible---the noise in our state estimate of their motion was larger than the actual movement---and, thus, we deactivated tracking and relied solely on visual servoing like for the static objects.

To minimize the interference between our MAVs in Challenge 3, we distributed the MAVs throughout the \SI{10 x 10}{\meter} start zone. In advance, we tested the electromagnetic interference between the MAVs by using two MAVs, one with activated camera, the other without computer and camera. Starting at \SI{6}{\metre} distance between both, the MAVs were moved closer while observing the decreasing GPS quality until below \SI{2}{\metre} were the signal could not detected reliably. This showed that the distance between the MAVs in the starting zone was sufficiently large to start all onboard systems simultaneously.

If the drop box detection was unsuccessful over a longer time period or we waited too long to enter the drop zone, we dropped the attached disk on purpose within the drop zone to achieve at least half the points per disk. In the first trial, we restarted the exploration pattern from the beginning after dropping a disk to find previously missed disks. As our disk detector worked very reliable, the probability of finding disks that have not been perceived in the already captured part of the arena turned out to be very low. Consequently, we changed this behavior to proceed with the exploration at the next waypoint and only began to follow a new exploration pattern when an exploration has been finished.

In our first trial, for safety reasons the MAVs were programmed to start one after another with delays between the consecutive takeoffs. The predictable trajectories of the MAVs to their initial positions combined with a reasonable association between starting positions and operation areas of the MAVs made it possible to start all MAVs in parallel in consecutive trials. Thus, to reduce the idle times of MAVs, we deactivated this safety procedure at takeoff as the flight paths could not cross before all MAVs were separated by altitude or sectors.

When the MAVs had to land during the challenge with a disk attached to the gripper because of a reset called by our team or the referees, the magnets were switched off. Thus, disks would remain on the ground. Due to conservative safety constraints, these disks could not be reliably picked afterwards. Due to the reliable detection of these objects combined with a deterministic behavior after system initialization, these objects caused deadlock situations when detected by an MAV. In the later trials, we kept information about object picking attempts and changed retrying to grip an object to only once in order to avoid deadlocks where a disk lay on the ground or when the blowing wind prohibited picking. Objects that could not be picked were not tried again until no other objects could be detected in a complete exploration run.

In addition, we randomized the start of the exploration pattern to avoid that non-pickable disks at the beginning of the pattern deterministically wasted time after every reset during the challenge. To cope with the steady wind, we further reduced the size of the descent cone for picking and added increasing offsets if picking was unsuccessful. As the strategies could not be tested before the trials and modifications of the onboard software during the trials was forbidden in Challenge 3, we implemented three slightly different picking strategies for these failure cases and let the MAV select one of these randomly before each picking attempt to avoid deterministic failure if a strategy fails.

Due to the fast detection and picking of objects during the trials, multiple MAVs arrived at the drop zone more often than anticipated. Thus, conflict resolution at the drop zone was crucial and not only a safety measure for rare conditions. In the first trial, two MAVs arrived at the drop zone at nearly the same time by coincidence. Due to the high transport speed, the second MAV could not stop outside the drop zone after the first MAV reported a position inside of the drop zone. Thus, both MAVs entered the drop zone and blocked it mutually. As a consequence, both MAVs left the drop zone again to a safe waiting position outside. Due to the inertia of the system in combination with communication latencies this led to an oscillating behavior were both MAVs entered and left the drop zone alternatingly resulting in a deadlock in which no MAV could deliver the objects. We addressed this by adding a random wait time in front of the drop zone before reentering it for consecutive trials. This resolves these deadlock situations as the probability for repetitive behavior decreases significantly. Another option would be an explicit semaphore for the drop zone, but this is not compatible with our approach to coordinate solely on broadcasted information.

During the MBZIRC, we disabled the visual object confirmation and assumed that every picking attempt was successful as false positives were far less problematic than false negatives in the scoring scheme. Furthermore, the gripper was much more reliable than expected.

The similar assignment of software MAV IDs to the same hardware in all trials led to misinterpretations of erroneous MAV behavior. As the laser height correction failed repeatedly on the same MAV, we replaced the laser altimeter between trials instead of investigating the software issues.
As discussed in the evaluation section, this problem was caused by the assigned altitude for approaching the drop box which was derived from the MAV ID. Shuffling the IDs between trials or at least a better systematic awareness of this indirect relation between hardware and resulting behavior given the same software when searching for the error could possibly have avoided the repetitive failing.

\section{Conclusion}
Operating complex robotic systems without manual adaptation to the current situation and with virtually no testing time is very challenging. 
Whereas many highly sophisticated state-of-the-art algorithms to all subproblems of the challenge exist, simpler and failsafe solutions are often key to success.
The complexity of the tasks is represented in the final results: From 18 teams participating in the Treasure Hunt only four were able to autonomously achieve partial task fulfillment. Five more teams were able to deliver at least one object with manual control. Also in Challenge 1, only nine teams out of 24 were able to land on the moving platform.
Both tasks were hard to fulfill with manual control due to the required precise and fast movements of the MAVs.

We came in third for both of the individual challenges Landing and Treasure Hunt. 
Together with our ground robot Mario---turning a valve stem with a wrench---our team NimbRo won the Grand Challenge overall---with two second places in the subchallenges Landing and Treasure Hunt and a first place in the ground robot subchallenge out of 14 participating teams.

We have provided detailed insight into our robust MAV setups for quickly landing on a fast moving target and for a collaborative search, pick, and place task. The viability of our approaches has been demonstrated in outdoor scenarios with minimum preparation time during the MBZIRC.

In Challenge 1, in particular the adaptive and fast trajectory replanning combined with a high-frequency pattern detection turned out to reliably match direction and velocity of the moving target. Furthermore, the simultaneous use of two cameras in combination with an adaptive yawing strategy enabled us to track the target pattern under fast maneuvers and in close proximity.
For Challenge 3, the reliable object perception and the robust and flexible gripper yielded a high exploration and picking speed resulting in one of the largest number of successful picks in the challenges.

We addressed many possible problems in advance; still, unforeseen issues occur during actual competitions, \eg the unexpected strong wind and deadlock situations.
The system could be robustified by adding more elements of randomness to the algorithms to prevent repetitive failing.

We believe that our contribution---and in general all experience from the MBZIRC tasks---will facilitate new ideas of how to operate flying robots in dynamic real-world environments.

\subsubsection*{Acknowledgments}
We would like to thank all members of our team NimbRo for their support before and during the competition.

This work has been supported by a grant of the Mohamed Bin Zayed International Robotics Challenge (MBZIRC) and grants BE 2556/7-2 and BE 2556/8-2 of the German Research Foundation (DFG).

\bibliographystyle{apalike}
\bibliography{literature_references}

\end{document}